%% file: egpaper_for_review.tex
\documentclass[10pt,twocolumn,letterpaper]{article}

\usepackage{iccv}
\usepackage{times}
\usepackage{epsfig}
\usepackage{graphicx}
\usepackage{amsmath}
\usepackage{amssymb}
\usepackage{booktabs}
\usepackage{subcaption}
\usepackage{amsthm}
\usepackage{algorithm2e}
\usepackage{sidecap}
\usepackage{wrapfig}

\newtheorem{theorem}{Theorem}

\usepackage[pagebackref=true,breaklinks=true,letterpaper=true,colorlinks,bookmarks=false]{hyperref}
\input{math_commands.tex}

\iccvfinalcopy 

\def\iccvPaperID{10511} 

\newcommand{\citep}[1]{\cite{#1}}
\newcommand{\citet}[1]{\cite{#1}}

\ificcvfinal\fi
\usepackage[capitalize]{cleveref}

\makeatletter
\def\@fnsymbol#1{\ensuremath{\ifcase#1\or \dagger\or *\or \ddagger\or
   \mathsection\or \mathparagraph\or \|\or **\or \dagger\dagger
   \or \ddagger\ddagger \else\@ctrerr\fi}}

\begin{document}

\title{TiDAL: Learning Training Dynamics for Active Learning}

\author{Seong Min Kye$^{1,}$\thanks{Equal contribution.}\quad Kwanghee Choi$^{2,}$\footnotemark[1]\quad Hyeongmin Byun$^{1}$\quad Buru Chang$^{3,}$\thanks{Corresponding author.\\This work was done while all authors were affiliated with Hyperconnect.}\\
$^1$Hyperconnect\quad $^2$Carnegie Mellon University\quad $^3$Sogang University\\
{\tt\small \{harris,hyeongmin.byun\}@hpcnt.com}\quad{\tt\small kwanghec@andrew.cmu.edu}\quad{\tt\small buru@sogang.ac.kr}
} 

\maketitle
\ificcvfinal\thispagestyle{empty}\fi

\input{Sections/0_Abstract}
\input{Sections/1_Introduction}
\input{Sections/2_Method}
\input{Sections/3_Experiments}
\input{Sections/4_Related_Work}
\input{Sections/5_Conclusion}

{\small
\bibliographystyle{ieee_fullname}
\bibliography{egbib}
}

\newpage
\onecolumn
\appendix
\section*{Appendix}
\input{Sections/A_Evidence}
\input{Sections/B_Pilot_Details}
\input{Sections/C_Further_Analysis}

\end{document}

%% file: math_commands.tex
\usepackage{amsmath,amsfonts,bm,amssymb}

\def\eqref#1{equation~\ref{#1}}

\def\1{\bm{1}}

\def\vp{{\bm{p}}}

\def\vs{{\bm{s}}}

\def\vy{{\bm{y}}}

\DeclareMathAlphabet{\mathsfit}{\encodingdefault}{\sfdefault}{m}{sl}
\SetMathAlphabet{\mathsfit}{bold}{\encodingdefault}{\sfdefault}{bx}{n}

\def\gD{{\mathcal{D}}}



%% file: Sections/0_Abstract.tex
\begin{abstract}\label{sec:0_abstract}
Active learning (AL) aims to select the most useful data samples from an unlabeled data pool and annotate them to expand the labeled dataset under a limited budget.
Especially, uncertainty-based methods choose the most uncertain samples, which are known to be effective in improving model performance.
However, previous methods often overlook \textit{training dynamics} (TD), defined as the ever-changing model behavior during optimization via stochastic gradient descent, even though other research areas have empirically shown that TD provides important clues for measuring the data uncertainty.
In this paper, we first provide theoretical and empirical evidence to argue the usefulness of utilizing the ever-changing model behavior rather than the fully trained model snapshot.
We then propose a novel AL method, Training Dynamics for Active Learning (TiDAL), which efficiently predicts the training dynamics of unlabeled data to estimate their uncertainty.
Experimental results show that our TiDAL achieves better or comparable performance on both balanced and imbalanced benchmark datasets compared to state-of-the-art AL methods, which estimate data uncertainty using only static information after model training.
\end{abstract}

%% file: Sections/1_Introduction.tex
\input{Figures/1_Model_Architecture}
\section{Introduction}\label{sec:1_introduction}
\noindent
\textit{``There is a tide in the affairs of men. Which taken at the flood, leads on to fortune." --- William Shakespeare}

Active learning (AL)~\citep{atlas1990training,lewis1994sequential} aims to solve the real-world problem of selecting the most useful data samples from large-scale unlabeled data pools and annotating them to expand labeled data under a limited budget.
Since the current deep neural networks are data-hungry, AL has increasingly gained attention in recent years.
Existing AL methods can be divided into two mainstream categories: diversity- and uncertainty-based methods.
Diversity-based methods~\citep{sener2018active,gissin2019discriminative} focus on constructing a subset that follows the target data distribution. 
Uncertainty-based methods~\citep{gal2017deep,beluch2018power,yoo2019learning} choose the most uncertain samples, which are known to be effective in improving model performance.
Hence, the most critical question for the latter becomes, \textit{``How can we quantify the data uncertainty?''}

In this paper, we leverage \textit{\textbf{training dynamics}} (TD) to quantify data uncertainty.
TD is defined as the ever-changing model behavior on each data sample during optimization via stochastic gradient descent.
Recent studies~\citep{chang2017active,samuli2017temporal,toneva2018empirical,swayamdipta2020dataset} have provided empirical evidence that TD provides important clues for measuring the contribution of each data sample to model performance improvement.
Inspired by these studies, we argue that the data uncertainty of unlabeled data can be estimated with TD.
However, most uncertainty-based methods quantify data uncertainty based on static information (\textit{e.g.}, loss~\citep{yoo2019learning} or predicted probability~\citep{sinha2019variational}) from a fully-trained model ``\textit{snapshot}," neglecting the valuable information generated during training.
We further argue that TD is more effective in separating uncertain and certain data than static information from a model snapshot captured after model training.
In \S\ref{subsec:2_4_effectiveness_of_training_dynamics_for_active_learning}, we provide both theoretical and empirical evidence to support our argument that TD is a valuable tool for quantifying data uncertainty.

Despite its huge potential, TD is not yet actively explored in the domain of AL.
This is because AL assumes a massive unlabeled data pool.
Previous studies track TD only for the training data every epoch as it can be recorded easily during model optimization.
On the other hand, AL targets a large number of unlabeled data, where \textit{\textbf{tracking the TD for each unlabeled sample requires an impractical amount of computation}} (\textit{e.g.}, inference all the unlabeled samples every training epoch).

Therefore, we propose TiDAL (\textit{\textbf{\underline{T}}ra\textbf{\underline{i}}ning \textbf{\underline{D}}ynamics for \textbf{\underline{A}}ctive \textbf{\underline{L}}earning}), a novel AL method that efficiently quantifies the uncertainty of unlabeled data by estimating their TD. 
We avoid tracking the TD of large-scale unlabeled data every epoch by predicting the TD of unlabeled samples with a TD prediction module. 
The module is trained with the TD of labeled data, which is readily available during model optimization.
During the data selection phase, we predict the TD of unlabeled data with the trained module to quantify their uncertainties.
We efficiently obtain TD using the module, which avoids inferring all the unlabeled samples every epoch.
Experimental results demonstrate that our TiDAL achieves better or comparable performance to existing AL methods on both balanced and imbalanced datasets.
Additional analyses show that our prediction module successfully predicts TD, and the predicted TD is useful in estimating uncertainties of unlabeled data.
Our proposed method are illustrated in Figure~\ref{fig:1_model_architecture}.

\textbf{Contributions of our study:} (1) We bridge the concept of training dynamics and active learning with the theoretical and experimental evidence that training dynamics is effective in estimating data uncertainty.
(2) We propose a new method that efficiently predicts the training dynamics of unlabeled data to estimate their uncertainty.
(3) Our proposed method achieves better or comparable performance on both balanced and imbalanced benchmark datasets compared to existing active learning methods. For reproducibility, we release the source code\footnote{\url{https://github.com/hyperconnect/TiDAL}}.

%% file: Figures/1_Model_Architecture.tex
\begin{figure}[t] 
\centering
\includegraphics[width=\columnwidth]{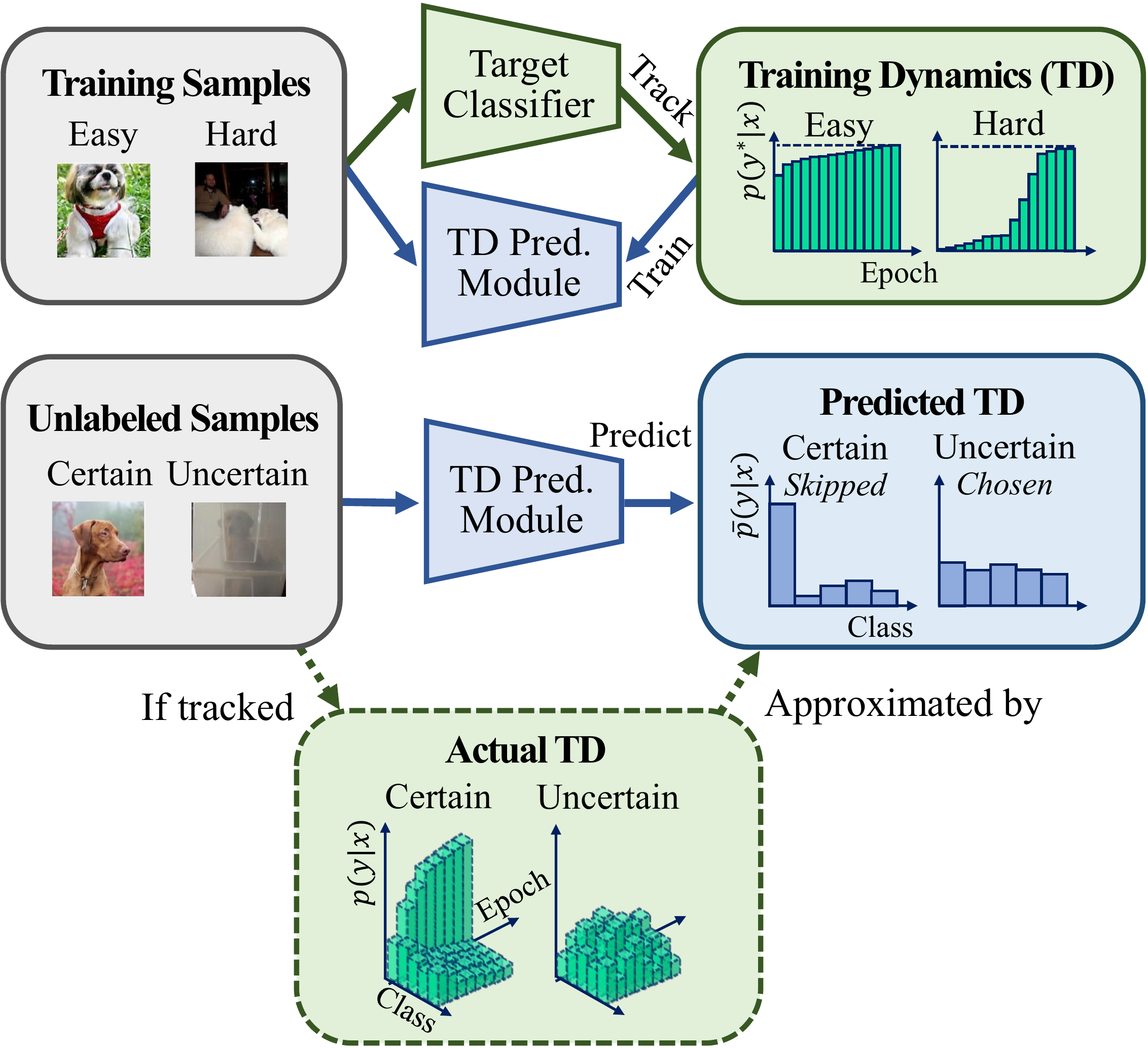}
\caption{
Our proposed TiDAL.
TD of training samples $x$ may differ even if they converge to the same final predicted probability $p(y^*|x)$ (Upper row).
Hence, we are motivated to utilize the readily available rich information generated during training, \textit{i.e.}, leveraging TD.
We estimate TD of large-scale unlabeled data using a prediction module instead of tracking the actual TD of all the unlabeled samples to avoid the computational overhead (Lower row).
}
\label{fig:1_model_architecture}
\end{figure}

%% file: Sections/2_Method.tex
\section{Preliminaries}\label{subsec:2_1_preliminaries}
To better understand our proposed method, we first summarize key concepts, including uncertainty-based active learning, quantification of uncertainty, and training dynamics.
\paragraph{Uncertainty-based active learning.} 
In this work, we focus on uncertainty-based AL for multi-class classification problems.
We define the predicted probabilities of the given sample $x$ for $C$ classes as:
\begin{align}
    \vp = [p(1|x),\enspace p(2|x),\enspace \cdots,\enspace p(C|x)]^T \in [0, 1]^C ,
\end{align}
where we denote the true label of $x$ as $y$ and the classifier as $f$.
$\gD$ and $\gD_u$ denote a labeled dataset and an unlabeled data pool, respectively.
The general cycle of uncertainty-based AL is in two steps: 
(1) train the target classifier $f$ on the labeled dataset $\gD$ and (2) select top-$k$ uncertain data samples from the unlabeled data pool $\gD_u$.
Selected samples are then given to the human annotators to expand the labeled dataset $\gD$, cycling back to the first step.

\paragraph{Quantifying uncertainty.}
The objective of this study is to establish a connection between the concept of TD and the field of AL. 
In order to clearly demonstrate the effectiveness of utilizing TD to quantify data uncertainty, we have employed two of the most prevalent and straightforward estimators, \textit{entropy}~\citep{shannon1948mathematical} and \textit{margin}~\citep{roth2006margin}, to measure data uncertainty in this paper.
Entropy $H$ is defined as follows:
\begin{align}
    H(\vp) &= - \sum\nolimits_{c=1}^C p(c|x) \log p(c|x),
\label{eq:entropy}
\end{align}
where the sample $x$ is from the unlabeled data pool $\gD_u$.
Entropy concentrates on the level of the model's confidence on the given sample $x$ and gets bigger when the prediction across the classes becomes uniform (\textit{i.e.}, uncertain).
Margin $M$ measures the difference between the probability of the true label and the maximum of the others:
\begin{align}
    M(\vp) = p(y|x) - \max_{c\ne\hat{y}}p(c|x),
\label{eq:margin}
\end{align}
where $y$ denotes the true label.
The smaller the margin, the lower the model's confidence in the sample, so it can be considered uncertain.
Both entropy and margin are computed with the predicted probabilities $\vp$ of the fully trained classifier $f$, only taking the snapshot of $f$ into account.

\paragraph{Defining training dynamics.}\label{subsec:2_2_definition_of_training_dynamics}
Our TiDAL targets to leverage TD of unlabeled data to estimate their uncertainties.
TD can be defined as any model behavior during optimization, such as the area under the margin between logit values of the target class and the other largest class \citet{pleiss2020identifying} or the variance of the predicted probabilities generated at each epoch \cite{swayamdipta2020dataset}.
In this work, we define the TD $\bar{\vp}^{(t)}$ as the area under the predicted probabilities of each data sample $x$ obtained during the $t$ time steps of optimizing the target classifier $f$:
\begin{align}
    \vp^{(i)} &= [p^{(i)}(1|x),\enspace p^{(i)}(2|x),\enspace \cdots,\enspace p^{(i)}(C|x)]^T, \\
    \bar{\vp}^{(t)} &= [\bar{p}^{(t)}(1|x),\enspace \bar{p}^{(t)}(2|x),\enspace \cdots,\enspace \bar{p}^{(t)}(C|x)]^T \nonumber \\
    &= \sum\nolimits_\tau \vp^{(\tau)} \Delta \tau \simeq \sum\nolimits_{i=1}^{t}\vp^{(i)}/t,
\label{eq:td_vector_def}
\end{align}
where $\vp^{(i)}$ is the predicted probabilities of a target classifier $f$ at the $i$-th time step.
$\Delta \tau$ is the unit time step to normalize the predicted probabilities.
For simplicity, we record $\vp^{(i)}$ every epoch and choose $\Delta \tau = 1/t$, namely, averaging the predicted probabilities during $t$ epochs \citep{swayamdipta2020dataset, song2019selfie}.
The TD $\bar{\vp}^{(t)}$ takes all the predicted probabilities during model optimization into account.
Hence, it encapsulates the overall tendency of the model during $t$ epochs of optimization, avoiding being solely biased towards the snapshot of $\vp^{(t)}$ in the final epoch $t$.

\input{Figures/2_Pilot_Study}
\section{Is TD Useful for Quantifying Uncertainty?}\label{subsec:2_4_effectiveness_of_training_dynamics_for_active_learning}
In this section, we provide empirical and theoretical evidence to support our argument: \textit{TD is more effective in separating uncertain data from certain data than the model snapshot}, where the latter is often utilized to quantify data uncertainty in previous works \cite{yoo2019learning,sinha2019variational}.

\subsection{Motivating Observation}\label{subsubsec:motivation_observation}
\paragraph{Settings.}
We aim to observe and compare the behavior of TD and the model snapshot for different sample difficulties.
However, it is nontrivial to directly measure sample-wise difficulty, inhibiting the quantitative analysis of data uncertainty.
To avoid this, we borrow the theoretical and empirical results of long-tailed visual recognition~\citep{liu2019large,cao2019learning,hong2021disentangling}: it is hard for the deep neural network-based model to train with fewer samples.
Hence, we regard major and minor class samples to contain many certain and uncertain samples for the model, respectively.
We train the target classifier $f$ on the long-tailed dataset during $T$ epochs to obtain the TD and the model snapshot.
We apply both approaches to the common estimators, entropy and margin.
We denote entropy and margin scores from the model snapshot as $H$ and $M$.
In opposition, we denote the TD-driven scores as $\bar{H}$ and $\bar{M}$.
More details and discussions are described in Appendix~\ref{appendix:b_pilot_details}.

\paragraph{Results.}
Figure~\ref{fig:2_pilot_study} shows the distribution the scores calculated with TD (\textit{x}-axis) and model snapshot (\textit{y}-axis).
We can observe that scores from TD ($\bar{H}, \bar{M}$) successfully separate the major and the minor class samples, whereas scores from the model snapshot ($H, M$) fail to do so.
We conclude that compared to model snapshots, TD is more helpful in separating uncertain samples from certain samples.

\subsection{Theoretical Evidence}
\begin{theorem}\label{theorem:1_elastic}
(Informal) Under the LE-SDE framework \citep{zhang2021imitating}, with the assumption of local elasticity \citep{he2019local}, certain samples and uncertain samples reveal different TD; especially, certain samples converge quickly than uncertain samples.
\end{theorem}
The above theorem discusses different model behaviors depending on the difficulty of the sample.
Compared to the uncertain sample, the certain sample has the same class samples nearby, which is the fundamental idea of level set estimation \citep{jiang2018trust} and nearest neighbor \citep{papernot2018deep} literature.
We suspect that, due to the local elasticity of deep nets, samples close by have a bigger impact on the certain sample, hence changing its predicted probability more rapidly.
As the certain sample is quicker to converge, its TD is larger than that of the uncertain sample.
Intuitively, slower to train, struggling the classifier is to learn, hence TD capturing the uncertainty in the classifier's perspective.

\begin{theorem}\label{theorem:2_entropy_td}
(Informal) Estimators such as Entropy (Equation \ref{eq:td_aware_entropy}) and Margin (Equation \ref{eq:td_aware_margin}) successfully capture the difference of TD between easy and hard samples even for the case where it cannot be distinguished via the predicted probabilities of the model snapshot.
\end{theorem}
The above theorem discusses the validity of entropy and margin on whether they can successfully differentiate between two samples of different TD but with the same final prediction.
With Theorem~\ref{theorem:1_elastic}, one can conclude that the common estimators' scores calculated with TD are effective in capturing the data uncertainty.
Due to the space constraints, we provide the details of the above results in Appendix~\ref{appendix:a_1_theoretical_guarantee}.

\section{Utilizing TD for Active Learning}\label{sec:2_method}
As tracking the TD of all the unlabeled data is computationally infeasible, we devise an efficient method to estimate the TD of unlabeled samples.
We train the module that directly predicts the TD of each sample by feeding the training samples, where its TD are freely available during training.
Then, based on the predicted TD of each unlabeled sample, we use the common estimators, entropy or margin, to determine which sample is the most uncertain so that human annotators can label it.
Hence, in this section, we describe the details of the module that estimates TD ($\S$\ref{subsec:2_5_training_dynamics_prediction_module}) and how to train the module ($\S$\ref{subsec:2_6_objective}).
Finally, calculating the uncertainties using the module predictions for active learning is illustrated ($\S$\ref{subsec:2_7_td_aware_uncertainty}).

\subsection{Training Dynamics Prediction Module}\label{subsec:2_5_training_dynamics_prediction_module}
As mentioned, it is not computationally feasible to track TD for the large-scale unlabeled data as it requires model inference on all the unlabeled data every training epoch.
Thus, we propose the TD prediction module $m$ to efficiently predict the TD of unlabeled data at the $t$-th epoch.
Being influenced by the previous studies \cite{corbiere2019addressing,yoo2019learning,sinha2019variational,kim2021task} that use additional modules to predict useful values such as loss or confidence by the target model outputs, multi-scale feature maps are aggregated and passed into our TD prediction module.
The module produces the $C$-dimensional predictions:
\begin{align}
\Tilde{\vp}^{(t)}_m = [\Tilde{p}^{(t)}_m(1|x),\enspace \cdots,\enspace \Tilde{p}^{(t)}_m(C|x)]^T \in [0,1]^C    
\end{align}
estimating the actual TD $\bar{\vp}^{(t)}$ of the given sample $x$ in Equation~\ref{eq:td_vector_def}.
TD prediction module is jointly trained with the target classifier using a handful of parameters, having a negligible computational cost during training.
The detailed architecture of the module is described in Appendix \ref{appendix:module_detail}.

Even though the architecture is similar to previous works \cite{yoo2019learning,sinha2019variational,kim2021task}, we observed that ours were much more stable during optimization and easier to train.
We suspect that it is due to the target task difference; previous works trained the module that outputs only a single value via regression, whereas our module outputs $C$-dimensional probability distribution, which is similar to the main task of classifying images.

\subsection{Training Objectives}\label{subsec:2_6_objective}
To train the target classifier $f$ at the $t$-th epoch, we use the cross-entropy loss function $\mathcal{L}_{\mathrm{target}}$ on the predicted probability $\vp^{(t)}$ and a one-hot encoded vector $\vy \in\{0, 1\}^{C}$ of the true label $y$:
\begin{equation}
    \mathcal{L}_{\mathrm{target}} = \mathcal{L}_{\mathrm{CE}}(\vp^{(t)}, \vy) = -\log p^{(t)}(y|x). \label{eq:loss_ce}
\end{equation}
Meanwhile, the prediction module $m$ learns the TD of a sample $x$ by minimizing the Kullback–Leibler (KL) divergence between the predicted TD $\Tilde{\vp}^{(t)}_m$ and the actual TD $\bar{\vp}^{(t)}$:
\begin{align}
    \mathcal{L}_{\mathrm{module}} &= \mathcal{L}_{\mathrm{KL}}(\bar{\vp}^{(t)}||\Tilde{\vp}^{(t)}_m)  \nonumber \\
    &= \sum\nolimits_{c=1}^C \bar{p}^{(t)}(c|x) \log \left(\frac{\bar{p}^{(t)}(c|x)}{\Tilde{p}_m^{(t)}(c|x)} \right). 
\label{eq:loss_kl}
\end{align}
The final objective function of our proposed method is defined as follows:
\begin{equation}
    \mathcal{L} =\mathcal{L}_\mathrm{target}+ \lambda  \mathcal{L}_\mathrm{module}
\end{equation}
where $\lambda$ is a balancing factor to control the effect of $\mathcal{L}_\mathrm{module}$ during model training.

\subsection{Quantifying Uncertainty with TD}\label{subsec:2_7_td_aware_uncertainty}
We argue that uncertain samples can be effectively distinguished from unlabeled data using the predicted TD.
To verify the effectiveness of leveraging TD, we feed the predicted TD to entropy and margin ($\S$\ref{subsec:2_1_preliminaries}) by replacing snapshot probability $\vp$ with the predicted TD $\bar{\vp}$.
We choose these estimators as they are widely used for quantifying uncertainty.
We feed $\bar{\vp}$, replacing $\vp$, to the entropy $\bar{H}$:
\begin{equation}
    \bar{H}(\bar{\vp}) = -\sum\nolimits_{c=1}^C \bar{p}(c|x) \log \bar{p}(c|x).
\label{eq:td_aware_entropy}
\end{equation}
Entropy $\bar{H}$ is maximized when $\bar{\vp}$ is uniform, \ie, the sample is uncertain for the target classifier.
Margin $\bar{M}$ is also similarly employed:
\begin{equation}
    \bar{M}(\bar{\vp}) = \bar{p}(\hat{y}|x)-\max_{c\ne \hat{y}} \bar{p}(c|x).
\label{eq:td_aware_margin}
\end{equation}
Since we do not have true labels of unlabeled samples, we use the predicted labels $\hat{y}$ of the target classifier instead of the true labels.
There are several possible variants of $\bar{M}$ depending on the definition of $\hat{y}$.
We conduct experiments to compare $\bar{M}$ with its variants.
The experimental details and results are in Appendix~\ref{appendix:c_2_variants_margins}.

At the data selection phase, we use the predicted TD $\Tilde{\vp}_m^{(T)}$ instead of the actual TD $\bar{\vp}^{(T)}$ as in Equation~\ref{eq:td_aware_entropy} \&~\ref{eq:td_aware_margin} to estimate the TD-driven uncertainties of the unlabeled sample $x$ at the final epoch $T$.
By using the estimated uncertainty with the predicted TD, we select the most informative samples for model training.

%% file: Figures/2_Pilot_Study.tex
\begin{figure}[t]
\centering
\subfloat[Entropy Distribution]{\includegraphics[width=0.24\textwidth]{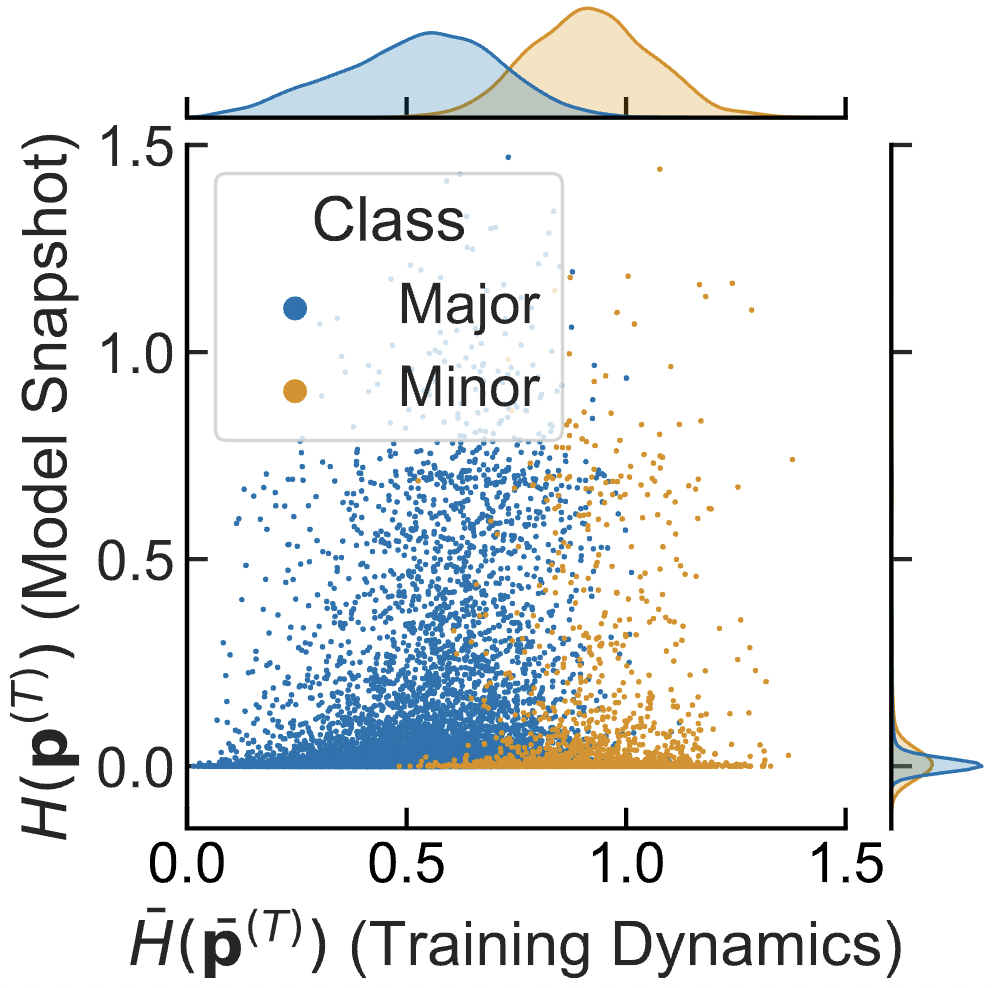}}
\subfloat[Margin Distribution]{\includegraphics[width=0.24\textwidth]{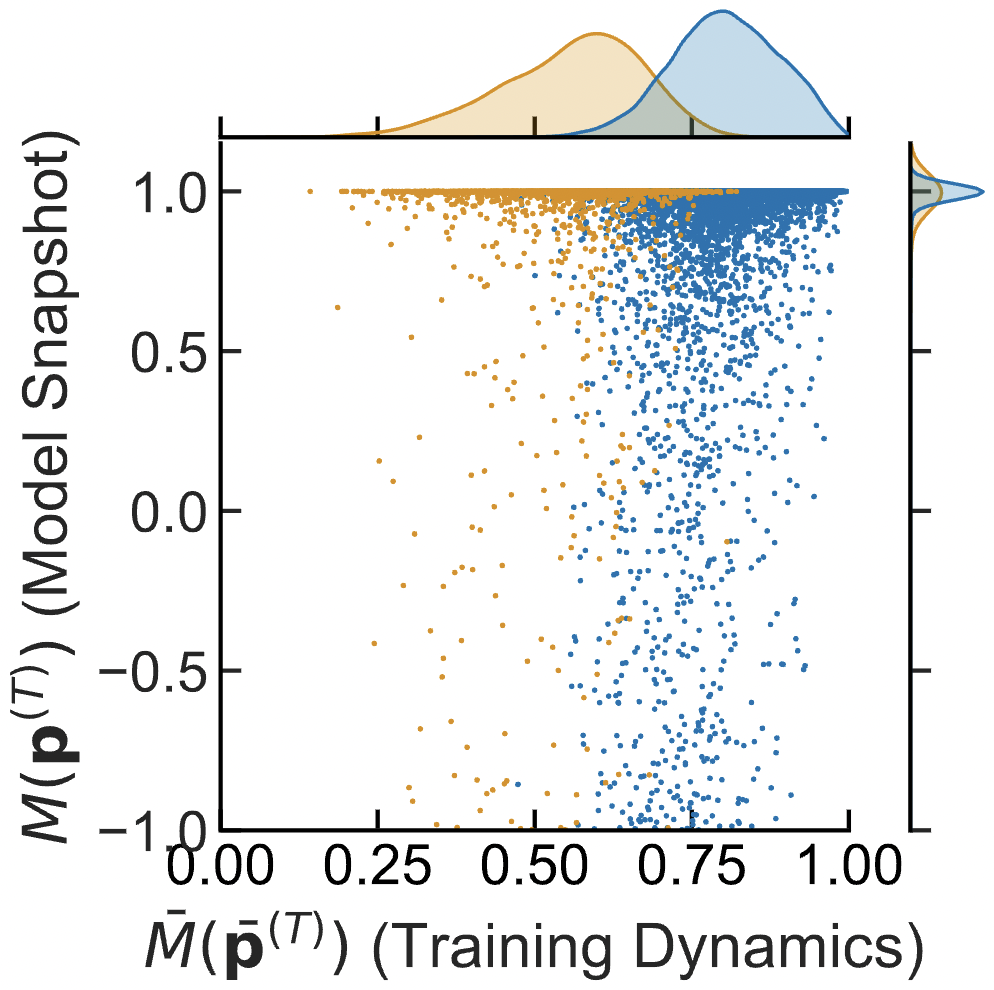}}
\vspace*{-2mm}
\caption{Score distribution after long-tailed training.
We plot the marginal distributions using kernel density estimation (KDE).
It is difficult to separate major (certain) and minor (uncertain) samples by the model snapshot-based scores (horizontal), unlike the TD-driven scores (vertical) that enable clearly separating the certain and uncertain samples.
}
\label{fig:2_pilot_study}
\end{figure}

%% file: Sections/3_Experiments.tex
\section{Experiments}\label{sec:3_experiments}
\input{Figures/2_Balanced_SOTA}
In this section, we experimentally verify the effectiveness of our method, TiDAL, which utilizes the estimated training dynamics from the prediction module to discern uncertain samples from unlabeled data.
We describe the detailed settings and the baseline methods for our experiments ($\S$\ref{subsec:3_1_experimental_setup}) and show the results on both balanced ($\S$\ref{subsec:3_2_balanced_results}) and imbalanced datasets ($\S$\ref{subsec:3_3_imbalanced_results}).
We further analyze whether the TD prediction module is effective for AL performance and can successfully estimate the TD ($\S$\ref{subsec:3_4_analysis}).
We end the section by discussing the potential limitations of our method ($\S$\ref{subsec:3_5_limitation}).

\subsection{Experimental Setup}\label{subsec:3_1_experimental_setup}
\paragraph{Datasets.}
To assess the performance of our proposed method and baseline methods, we conduct experiments on the following five datasets: CIFAR10/100~\citep{krizhevsky2009learning}, FashionMNIST~\citep{xiao2017fashion}, SVHN~\citep{netzer2011reading}, and iNaturalist2018~\citep{van2018inaturalist}.
Since CIFAR and FashionMNIST are both balanced, we further modify them to simulate the data imbalance in the real world, following the previous long-tail visual recognition studies~\citep{cao2019learning,liu2019large,zhou2020bbn,hong2021disentangling}. 
The imbalance ratio is defined as $N_{\text{max}}/N_{\text{min}}$ where $N$ is the number of samples in each class.
We make two variants with data imbalance ratios 10 and 100 for each dataset.
Unlike the above, SVHN and iNaturalist18 are already imbalanced.
Especially, iNaturalist2018 is commonly chosen to demonstrate how methods work in imbalanced real-world settings.
The dataset statistics are summarized in Appendix~\ref{appendix:b_expeirments_details}.

\paragraph{Baselines.}
For a fair comparison, we compare our TiDAL with the following baselines which train a target classifier with only labeled data.
\textbf{Random sampling}: a simple baseline that randomly selects data samples from the unlabeled dataset.
\textbf{Entropy sampling}~\citep{shannon1948mathematical}: an uncertainty-based method that selects data samples based on the maximum entropy.
\textbf{BALD}~\citep{gal2017deep}: an uncertainty-based method that selects data samples based on the mutual information between the model prediction and the posterior.
\textbf{CoreSet}~\citep{sener2018active}: a diversity-based method that selects representative data samples covering all data through a minimum radius.
\textbf{LLoss}~\citep{yoo2019learning}: an uncertainty-based method that learns to estimate the errors of the predictions (loss) made by the learner and select data samples based on the predicted loss.
\textbf{CAL}~\citep{zhang2021cartography}: recent work on using TD, gathering samplewise TD information on whether the classifier was consistently correct or not during training.
CAL splits the samples into two classes by applying a heuristic threshold to the TD information to train a binary classifier that outputs uncertainty score.
To verify the effectiveness of TiDAL, we further compare it with the two semi-supervised AL methods, \textbf{VAAL}~\citep{sinha2019variational} and \textbf{TA-VAAL}~\citep{kim2021task} in Appendix \ref{appendix:vaal}.
Note that these methods further utilize unlabeled data for training the selection module, thus it is unfair for our TiDAL.


\paragraph{Active learning setting.}
We follow the same setting from \citet{beluch2018power,yoo2019learning} for the detailed AL settings.
For the initial step, we randomly select initial samples to be annotated from the unlabeled dataset, where we use them to train the initial target classifier.
Then, we obtain a random subset from the unlabeled data pool $\gD_u$ to choose the top-$k$ samples based on the criterion of each method, where those samples will be annotated.
We repeat the above cycle, training a classifier from scratch from the continuously expanding labeled set.

\input{Figures/3_Imbalanced_Synth_SOTA}
\paragraph{Implementation details.}
For a fair comparison, we use the same backbone network ResNet-18 \citep{he2016deep} except for iNaturalist2018, where we use ResNet-50 \citep{he2016deep} pretrained on ImageNet \citep{deng2009imagenet}.
All models are trained with SGD optimizer with momentum $0.9$, weight decay $5\cdot10^{-4}$, and learning rate (LR) decay of $0.1$.
For CIFAR10/100 and SVHN, we train the model for 200 epochs with an initial LR of $0.1$ and decay at epoch $160$.
For FashionMNIST, $100$ epochs with an initial LR of $0.1$ and decay at epoch $80$.
For iNaturalist2018, $50$ epochs with an initial LR of $0.01$ and decay at epoch $40$.
For CIFAR10/100, SVHN and FashionMNIST, we set the batch size and the unlabeled subset size to be $128$ and $10^4$, respectively.
For iNaturalist2018, which is much larger than other datasets, we set the batch size and the unlabeled subset size to 256 and $10^6$, respectively. We set the balancing factor to 1.0.

\paragraph{Evaluation details.}
To compare with other state-of-the-art baselines, we show the average accuracy and 95\% confidence interval with three trials.
We mainly compare the model performances with relative accuracy improvement to random sampling, demonstrating how much it improves upon the naive approach on each cycle.
Additionally, absolute accuracy is also plotted in Appendix~\ref{appendix:b_expeirments_details}.

\subsection{Results on Balanced Datasets}\label{subsec:3_2_balanced_results}
Figure~\ref{fig:2_balanced_sota} and \ref{fig:13_appendix_absolute_real_sota} compare our TiDAL against the state-of-the-art methods on various balanced datasets: CIFAR10, CIFAR100, and FashionMNIST.
For all the datsets, the two variants of TiDAL outperform all the baselines at all AL cycles except for LLoss, which shows better improvement than TiDAL ($\bar{M}$) on CIFAR10 with an imbalance ratio of 100.
Nonetheless, our TiDAL achieves the best final performance compared to all the baselines.
CAL, which uses training dynamics, generally underperforms compared to others.
We suspect that CAL is sensitive to its threshold hyperparameter.

\subsection{Results on Imbalanced Datasets}\label{subsec:3_3_imbalanced_results}
\noindent
\textbf{Synthetically imbalanced datasets.}
Similar to the above, Figure~\ref{fig:3_imbalanced_synth_sota}, \ref{fig:10_additional_imbalanced_synth_sota}, and \ref{fig:14_appendix_absolute_imbalanced_synth_sota} shows the performance improvements on the synthetically imbalanced datasets with the two imbalance ratios, 10 and 100.
Except for the CIFAR10 with an imbalance ratio of 100, our methods show superb performance across all the imbalanced settings.
TiDAL performs especially well with a small variance in imbalanced CIFAR100, where the number of classes is the largest.
In imbalanced FashionMNIST, the performance quickly rises to 2.5k labeled images and then saturates. 
This implies that FashionMNIST is easier than other datasets, and needs to focus more on the early training steps to compare with other models.
TiDAL also shows overall better performance on FashionMNIST, especially in the early steps.

\paragraph{Real-world imbalanced datasets.}
Figure~\ref{fig:4_imbalanced_real_sota} and \ref{fig:13_appendix_absolute_real_sota} shows evaluation results on real-world imbalanced datasets.
For iNaturalist2018, which is the large-scale long-tailed classification dataset, TiDAL shows outstanding performance compared to other methods.
For SVHN, TiDAL shows the best improvements with low variance as the number of labeled images increases except for the initial stage.
LLoss shows outstanding performance only in the initial stage, where we presume that the loss prediction module of LLoss acts as a regularizer during model optimization.

\input{Figures/4_Imbalanced_Real_SOTA}
\input{Figures/5_Ablation}

\subsection{Analysis on the TD Prediction Module}\label{subsec:3_4_analysis}
\paragraph{Effectiveness of the TD prediction module.}
In order to verify the efficacy of using the predicted TD $\Tilde{\vp}_{m}$, we conduct an ablation test that compares the performance between when using and not using the TD prediction module $m$.
Figure~\ref{fig:5_ablation} shows the results on balanced CIFAR10/100.
We observe that $\bar{H}(\Tilde{\vp}_m)$ and $\bar{M}(\Tilde{\vp}_m)$ using the predicted TD $\Tilde{\vp}_m$ to estimate the data uncertainty significantly outperform the methods $H(\vp)$ and $M(\vp)$ that use only the final predicted probabilities $\vp$ of the target classifier $f$, showing better performance in the whole training cycle.
Even $M(\vp)$ shows temporary improvement in earlier steps on CIFAR100, $\bar{H}(\Tilde{\vp}_m)$ and $\bar{M}(\Tilde{\vp}_m)$ maintain stable improvement, eventually winning over $M(\vp)$.
This indicates that the predicted TD $\Tilde{\vp}_m$ of the TD prediction module $m$ produces better data uncertainty estimation than the predicted probability $\vp$ of the target classifier $f$.


\paragraph{Predictive performance of the TD prediction module.}
We verify whether the TD prediction module $m$ accurately predicts the actual TD $\bar{\vp}$.
Its prediction performance is crucial as we use the predicted TD $\Tilde{\vp}_m$ of the module $m$ to quantify uncertainties of unlabeled data.
Using the KL divergence $\mathcal{L}_{\mathrm{KL}}$, we analyze that the predicted TD $\Tilde{\vp}_m$ converges to the actual TD $\bar{\vp}$ at the data selection phase.
We calculate $\mathcal{L}_{\mathrm{KL}}(\bar{\vp}^{(T)}||\tilde{\vp}^{(t)}_m)$ and compare it with $\mathcal{L}_{\mathrm{KL}}(\bar{\vp}^{(T)}||\vp^{(t)})$ which is set as a baseline computed with the actual TD $\bar{\vp}$ and the predicted probabilities $\vp$ (snapshot) of the target classifier $f$.
In this analysis, we use the balanced CIFAR10 where the sample-wise averaged KL divergence scores are computed on the test set.
Figure~\ref{fig:6_kl_divergence} shows that the final predicted TD successfully approximates the actual TD, while the predicted probability is highly different from the actual TD.
We conclude that the TD prediction module $m$ can produce the TD efficiently, leading to performance improvement, and the predicted TD acts as a better approximation of the actual TD than the predicted probability of a model snapshot captured at each epoch.

\subsection{Limitations}\label{subsec:3_5_limitation}
We found two potential limitations of our TiDAL derived from the fact that it relies on the outputs of the target classifier to compute the TD.
First, TiDAL is designed only for classification tasks, and thus it cannot be applied to AL targeting other tasks, such as regression~\citep{cohn1994improving,gong2022meta}.
Second, TiDAL is highly influenced by the performance of the target classifier, especially when the target classifier wrongly classifies the hard negative samples with a high confidence during model optimization.
These samples can be treated as certain samples (i.e. will not be selected for annotation) because they have low estimated uncertainties from the predicted TD, even though the target classifier fails to predict the true label of the samples correctly.
As a future work, we will study extending our TiDAL in the task-agnostic ways with a safeguard combating the wrongly classified samples. 

%% file: Figures/2_Balanced_SOTA.tex
\begin{figure*}[t] 
\centering
\subfloat{\includegraphics[width=0.31\textwidth]{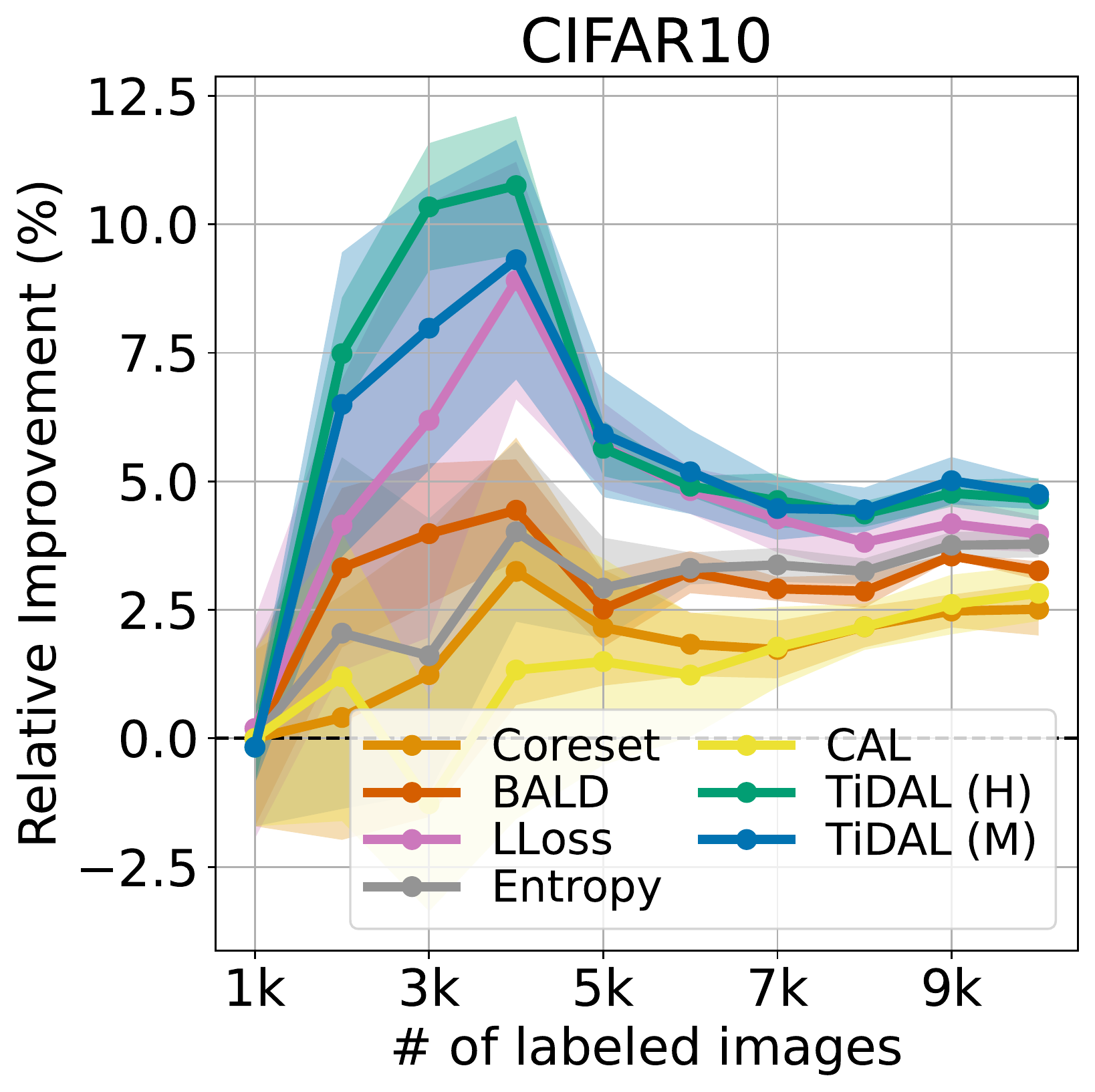}}
\subfloat{\includegraphics[width=0.31\textwidth]{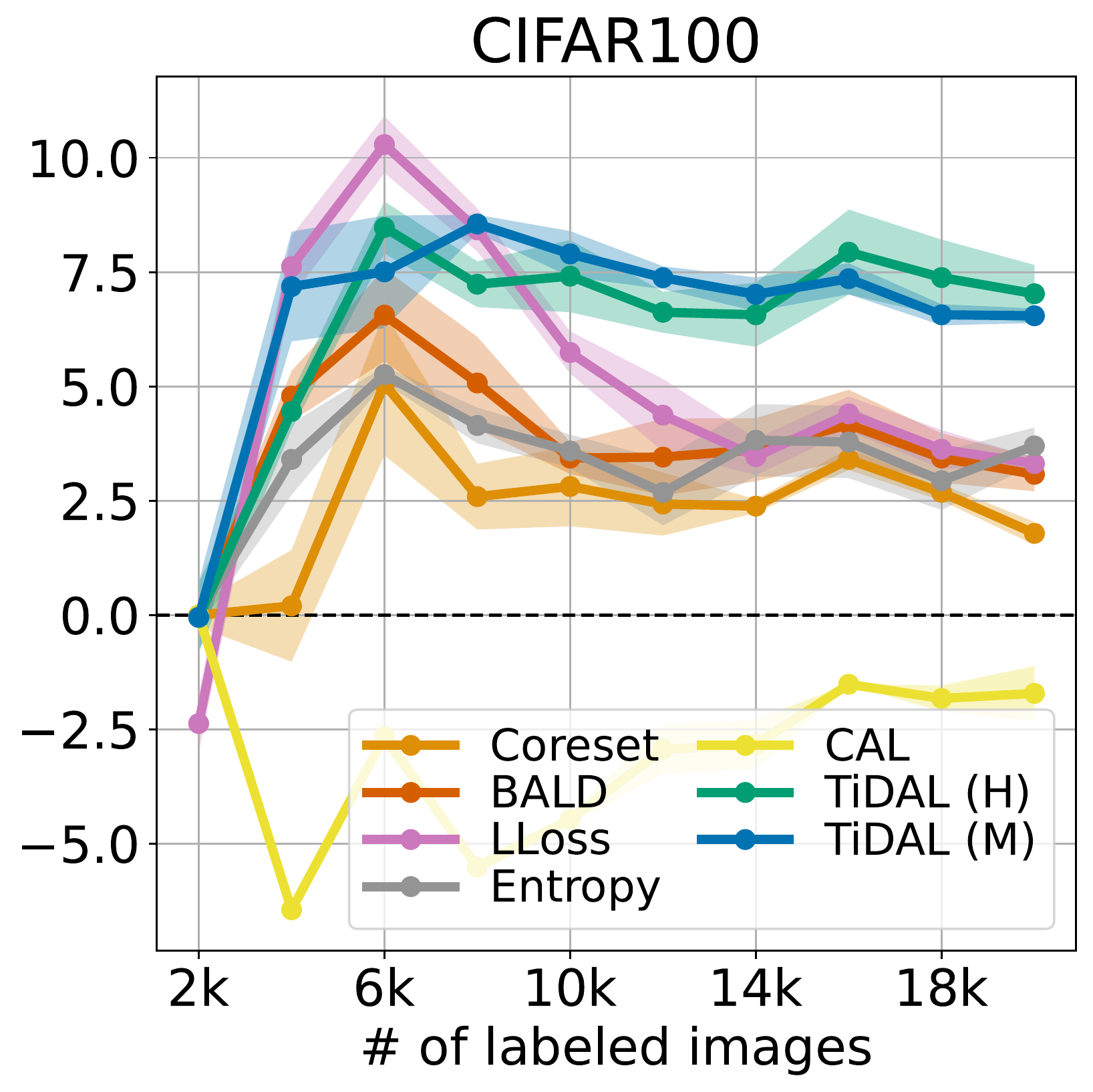}}
\subfloat{\includegraphics[width=0.31\textwidth]{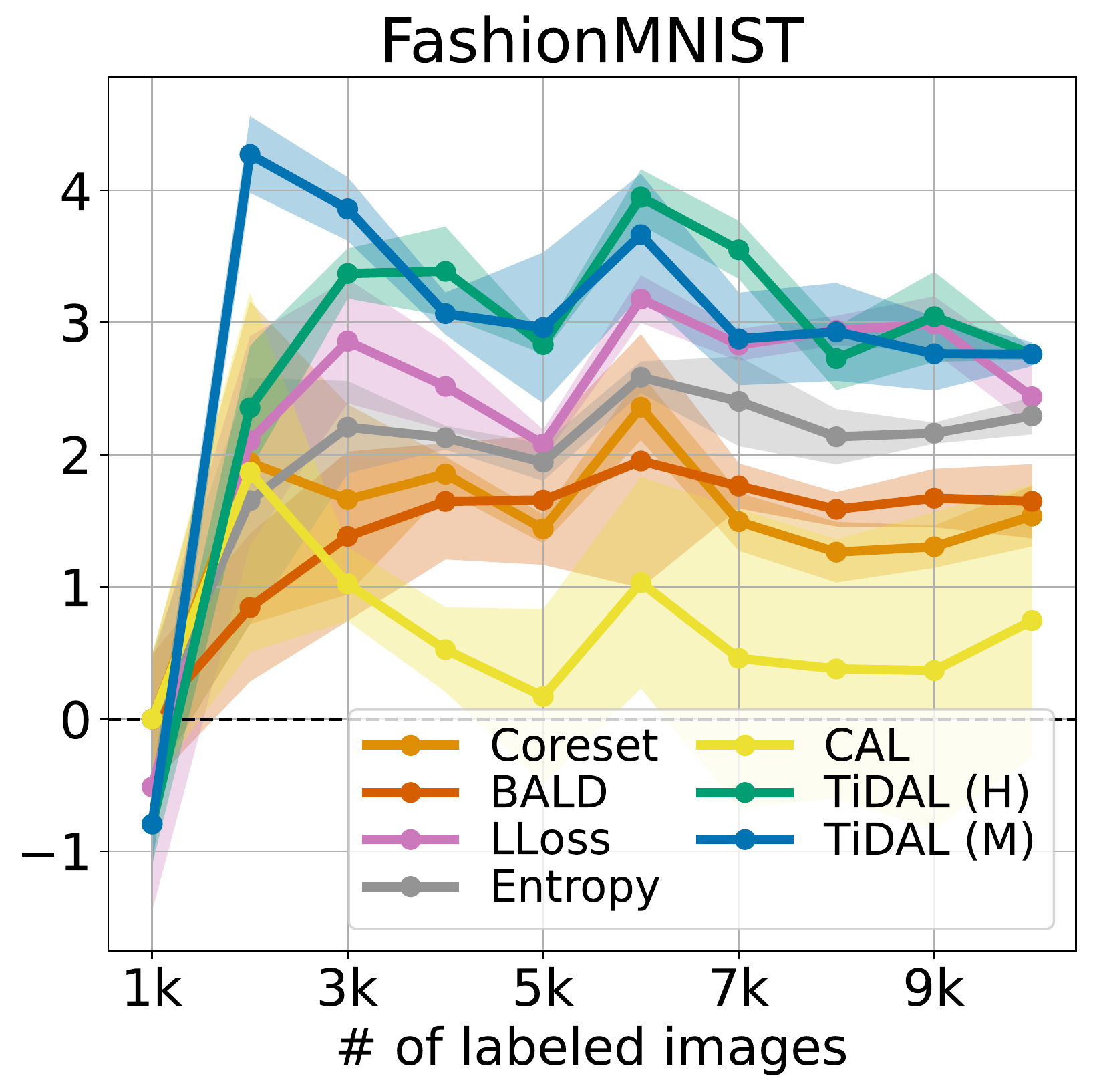}}
\vspace*{-2mm}
\caption{
    Averaged relative accuracy improvement curves and their 95\% confidence interval (shaded) of AL methods over the number of labeled samples on balanced datasets.
    TiDAL ($\bar{H}$) and TiDAL ($\bar{M}$) denote the performance of TiDAL when with entropy $\bar{H}$ and margin $\bar{M}$ as the data uncertainty estimation strategy, respectively.
}
\vspace*{-2mm}
\label{fig:2_balanced_sota}
\end{figure*}

%% file: Figures/3_Imbalanced_Synth_SOTA.tex
\begin{figure*}[t] 
\centering
\subfloat{\includegraphics[width=0.31\textwidth]{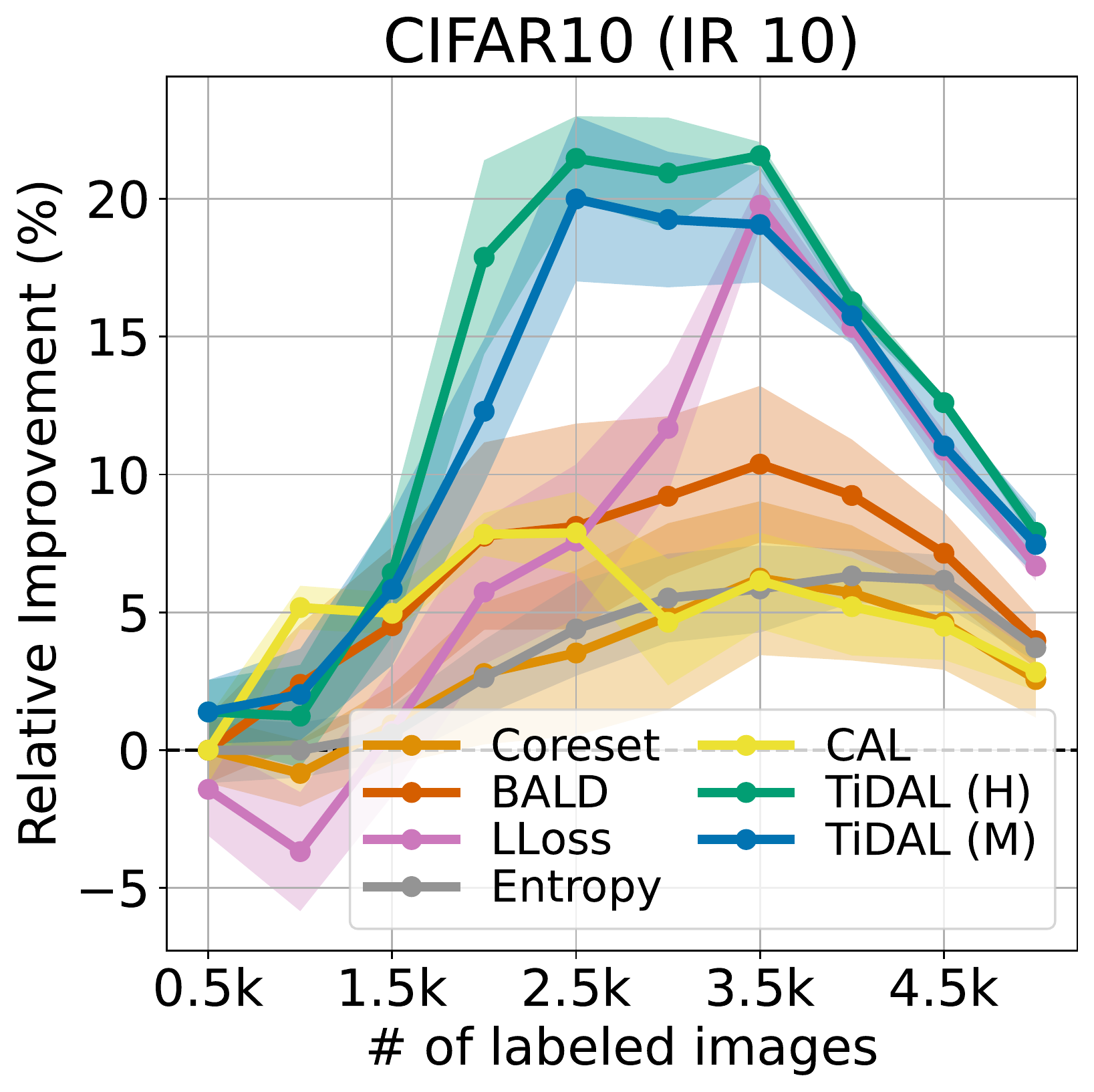}}
\subfloat{\includegraphics[width=0.31\textwidth]{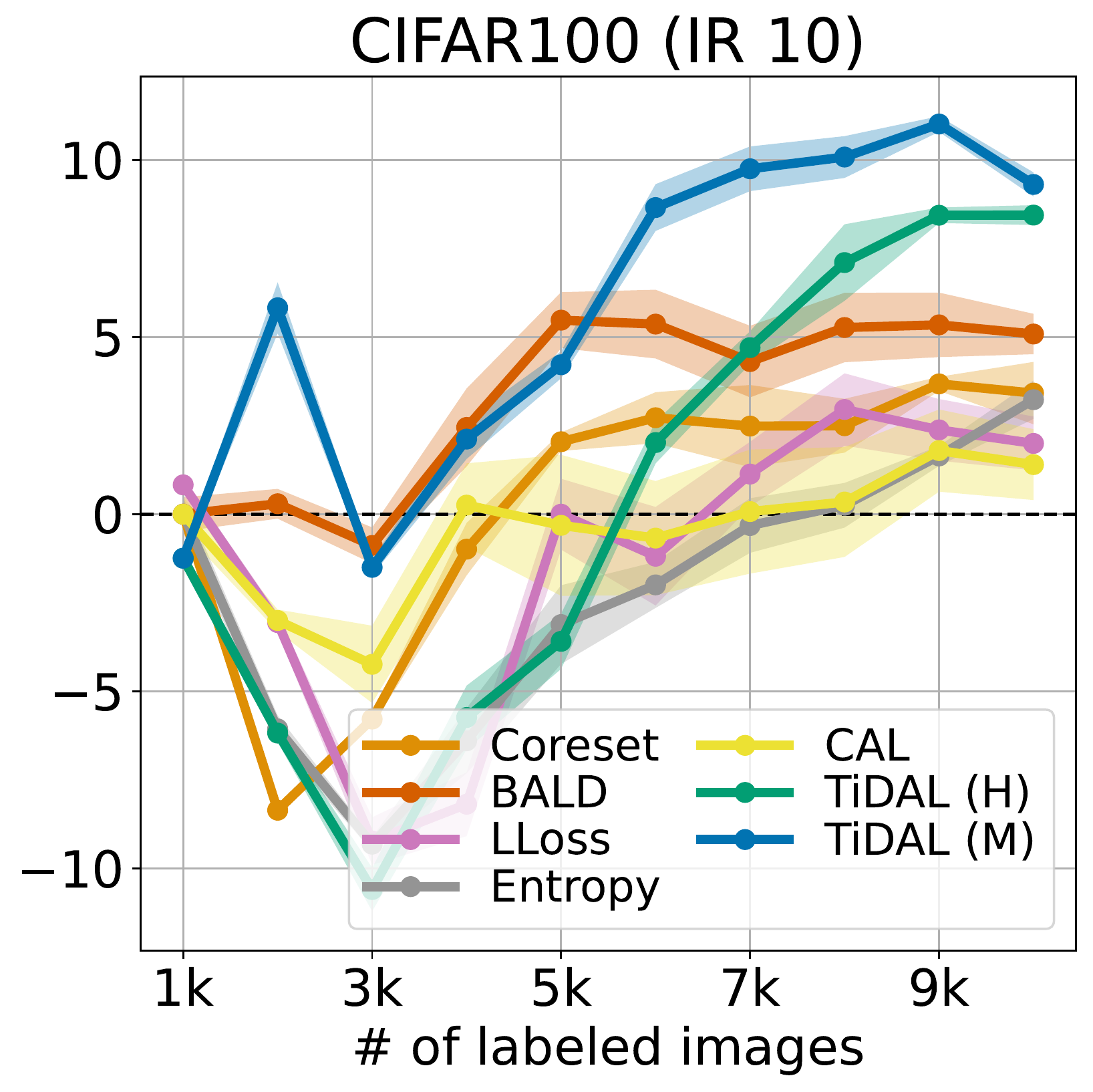}}
\subfloat{\includegraphics[width=0.31\textwidth]{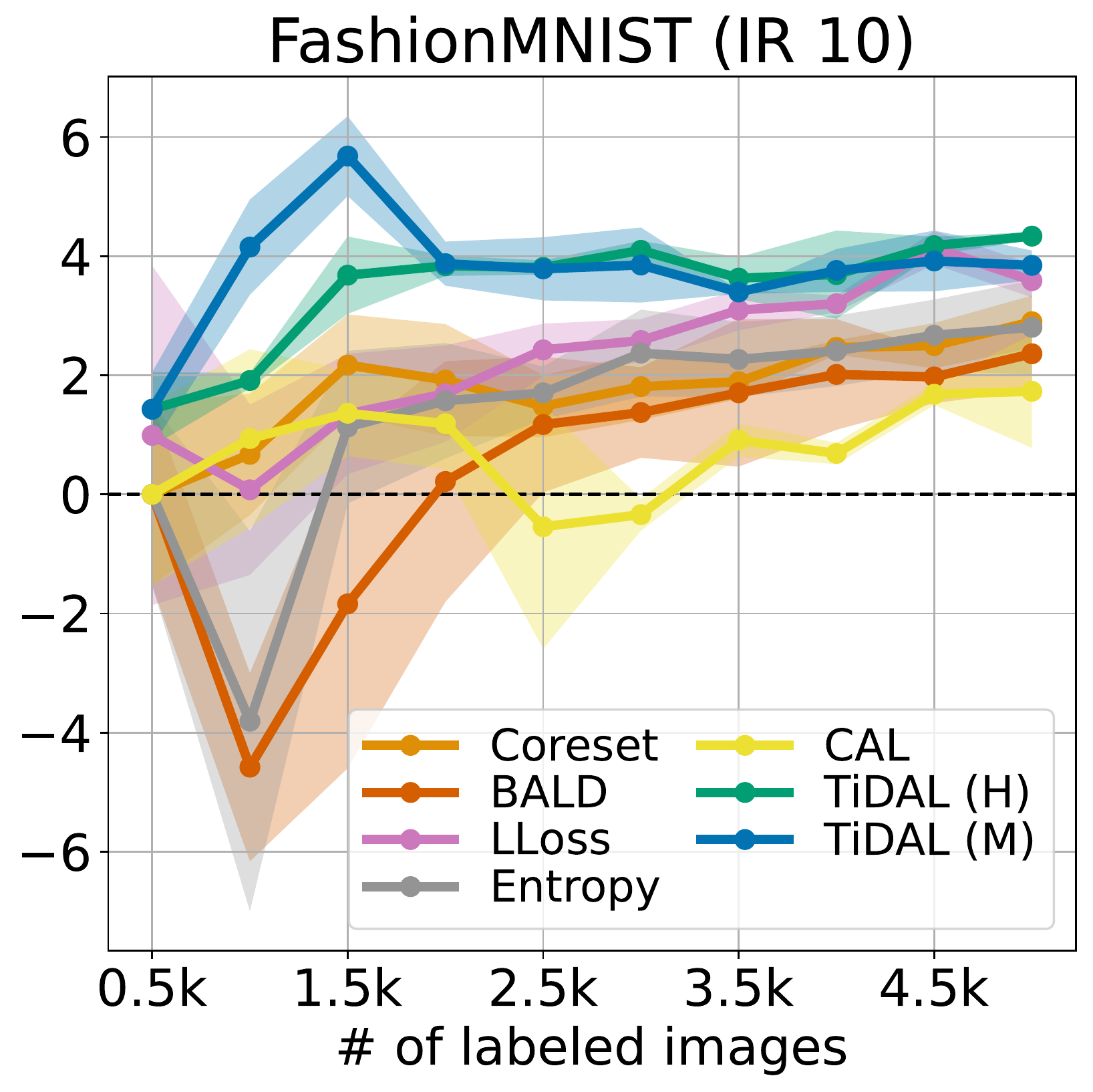}}\\
\subfloat{\includegraphics[width=0.31\textwidth]{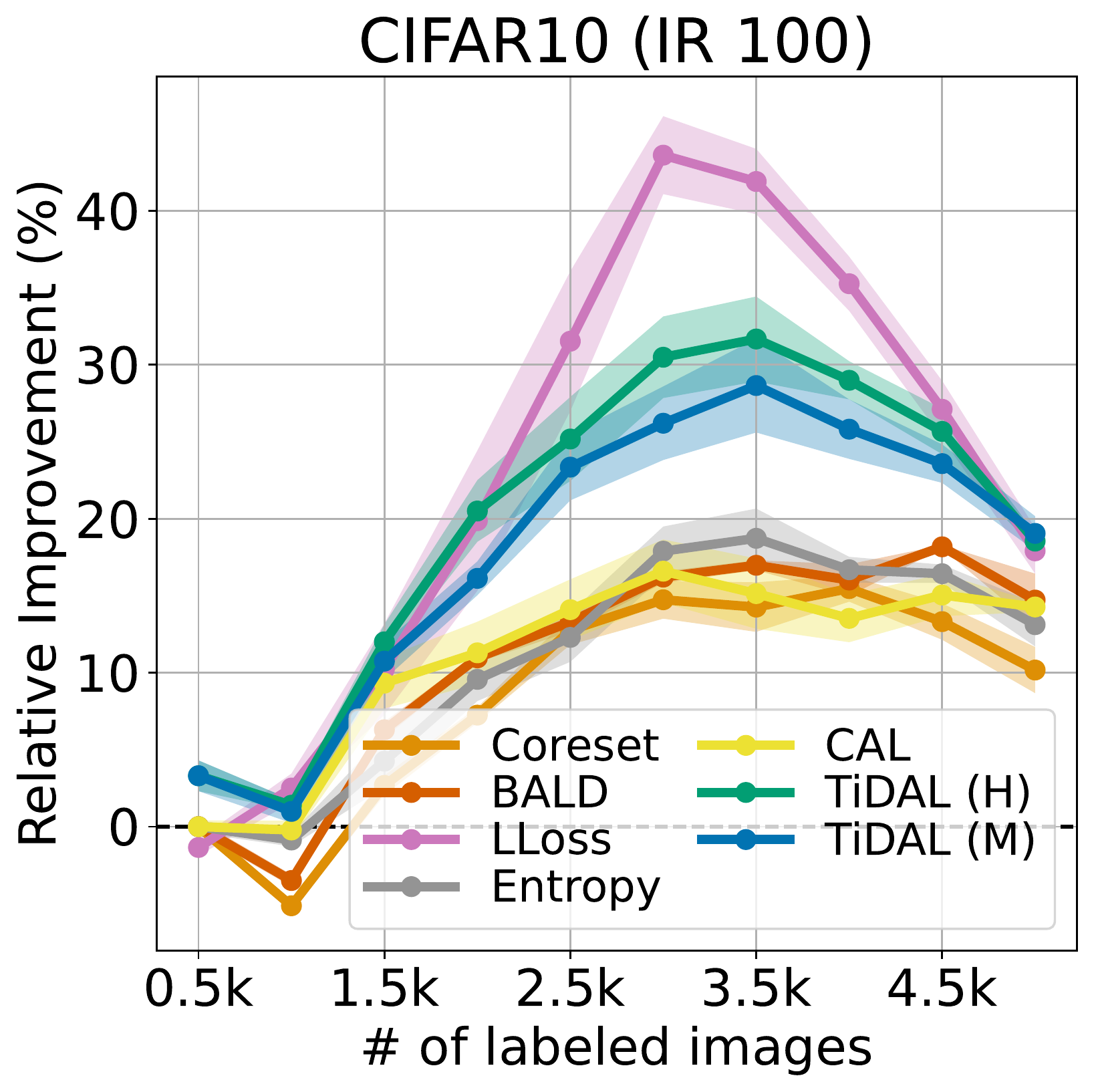}}
\subfloat{\includegraphics[width=0.31\textwidth]{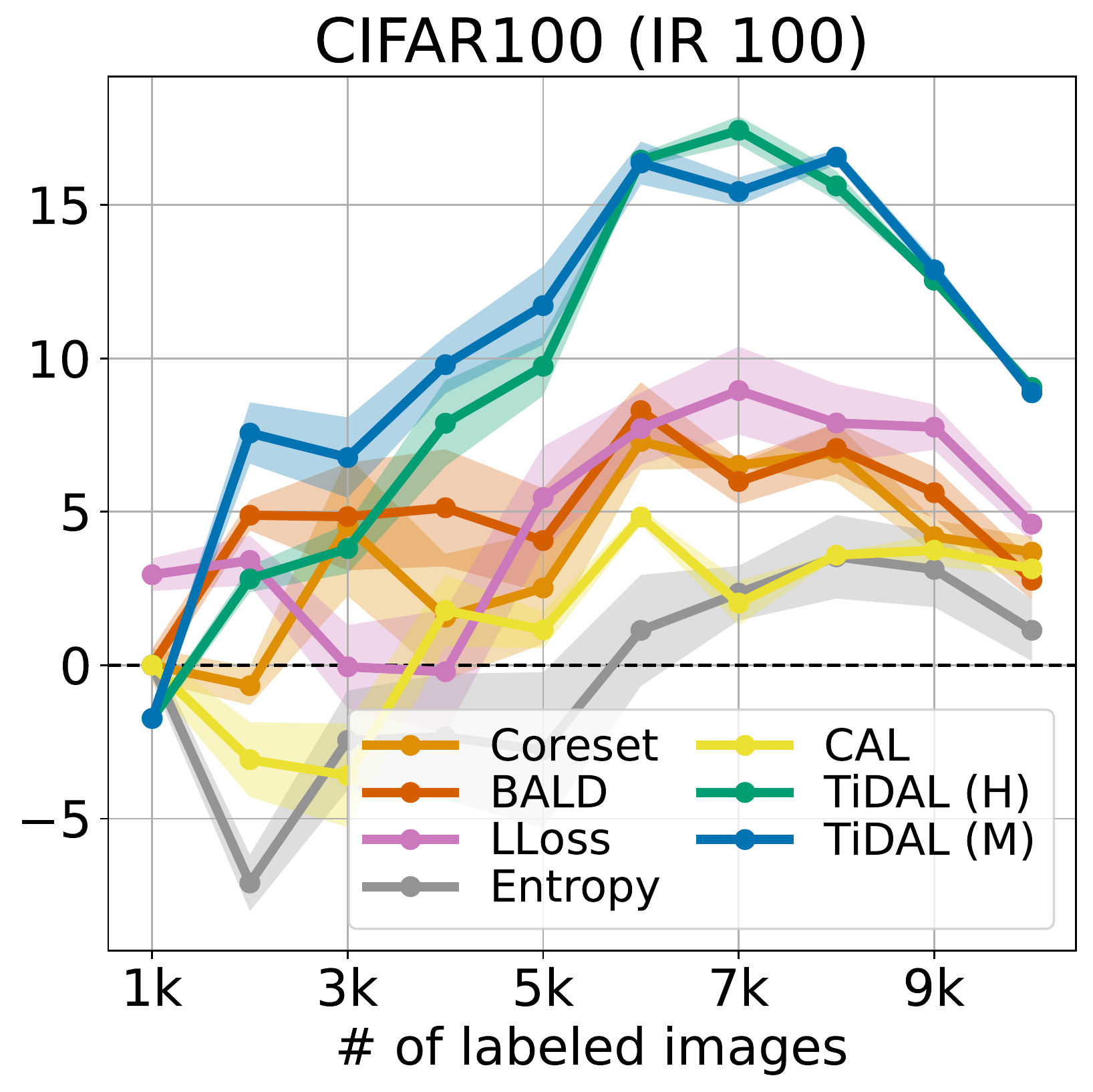}}
\subfloat{\includegraphics[width=0.31\textwidth]{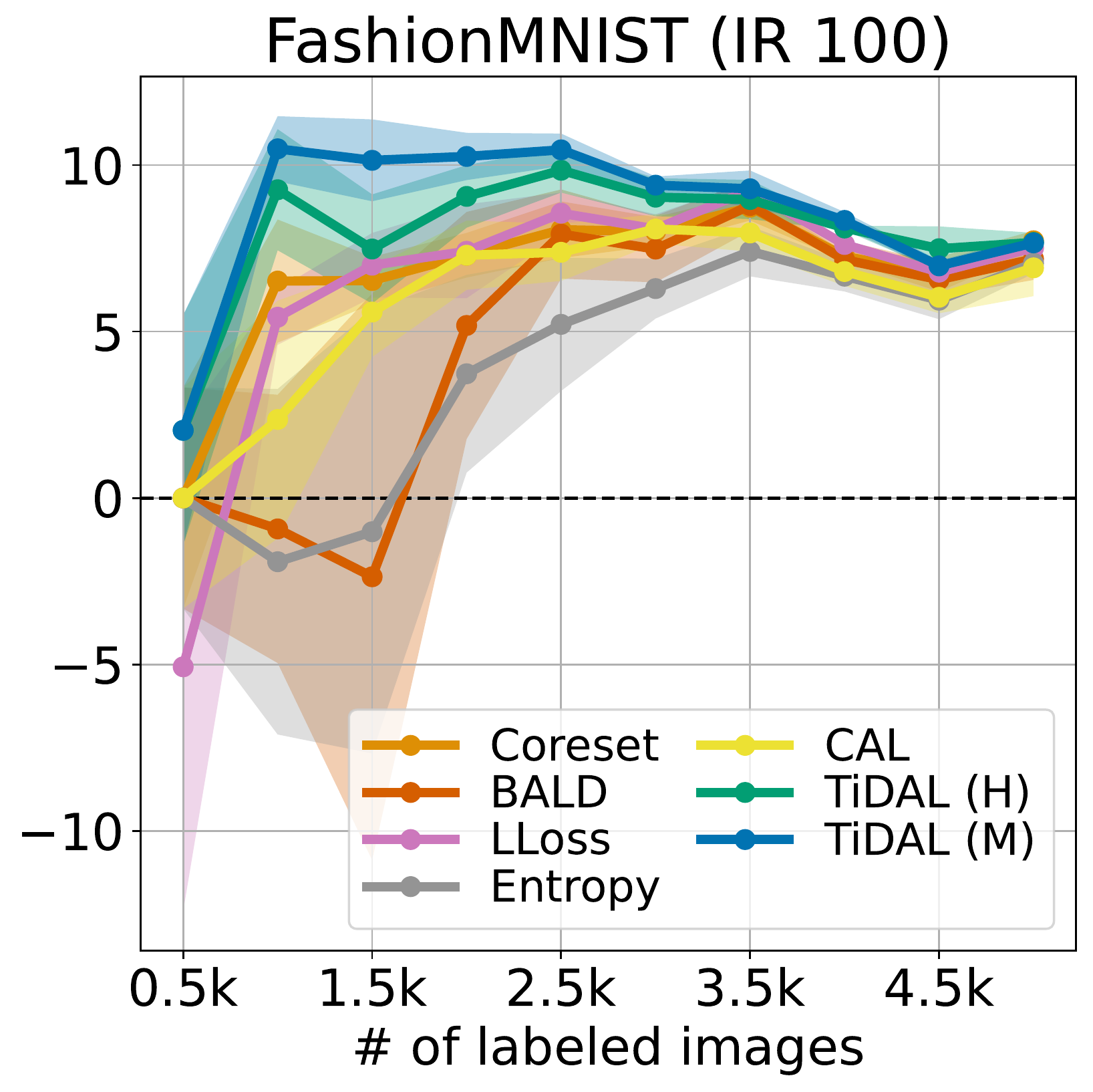}}
\vspace*{-2mm}
\caption{
    Averaged relative accuracy improvement curves and their 95\% confidence interval (shaded) of AL methods over the number of labeled samples on synthetically imbalanced datasets.
    We use the imbalance ratio (IR) of 10 and 100 on CIFAR10, CIFAR100, and FashionMNIST.
}
\vspace*{-2mm}
\label{fig:3_imbalanced_synth_sota}
\end{figure*}

%% file: Figures/4_Imbalanced_Real_SOTA.tex
\begin{figure*}[t] 
\centering
\subfloat{\includegraphics[width=0.33\textwidth]{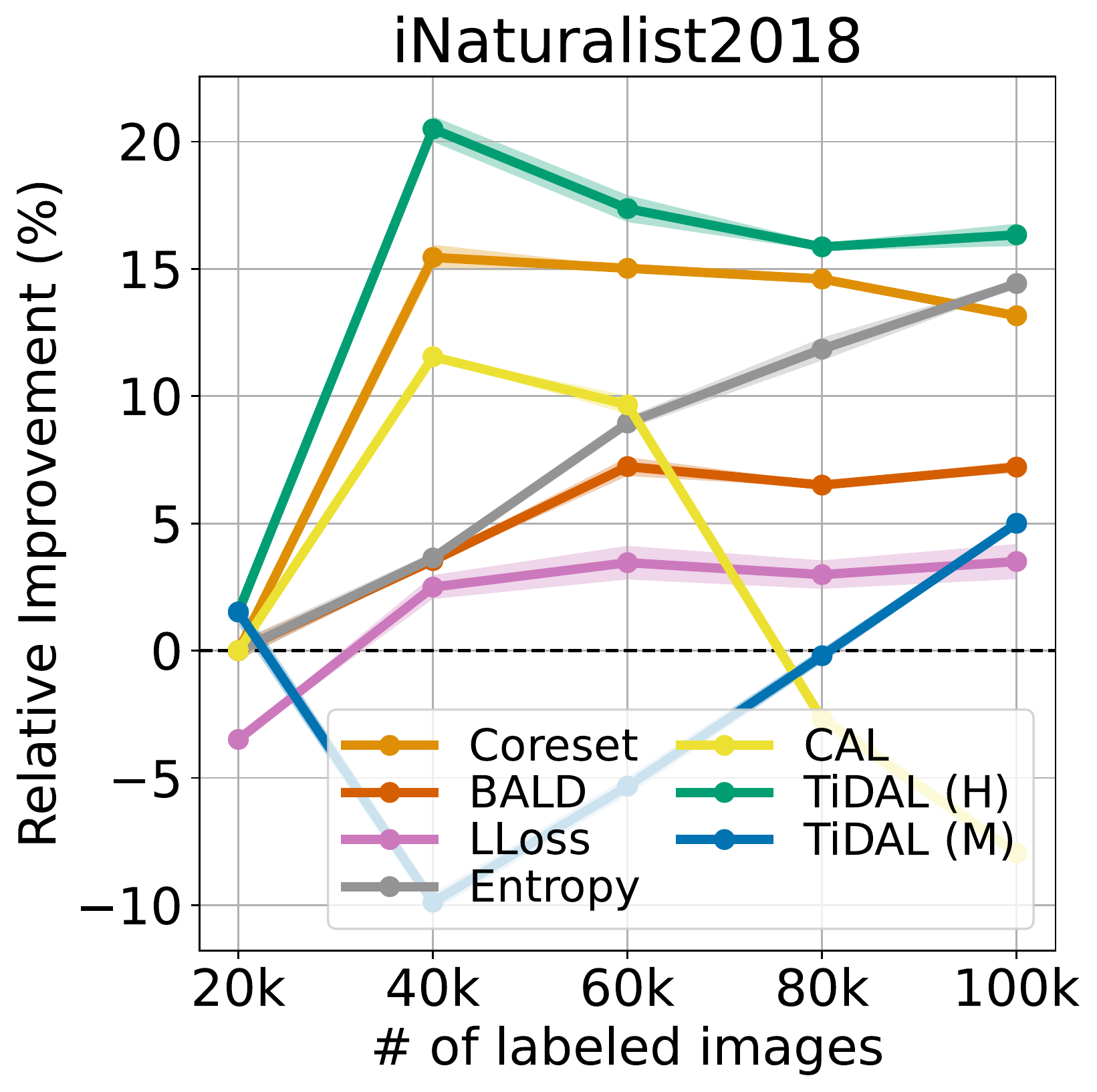}}
\subfloat{\includegraphics[width=0.33\textwidth]{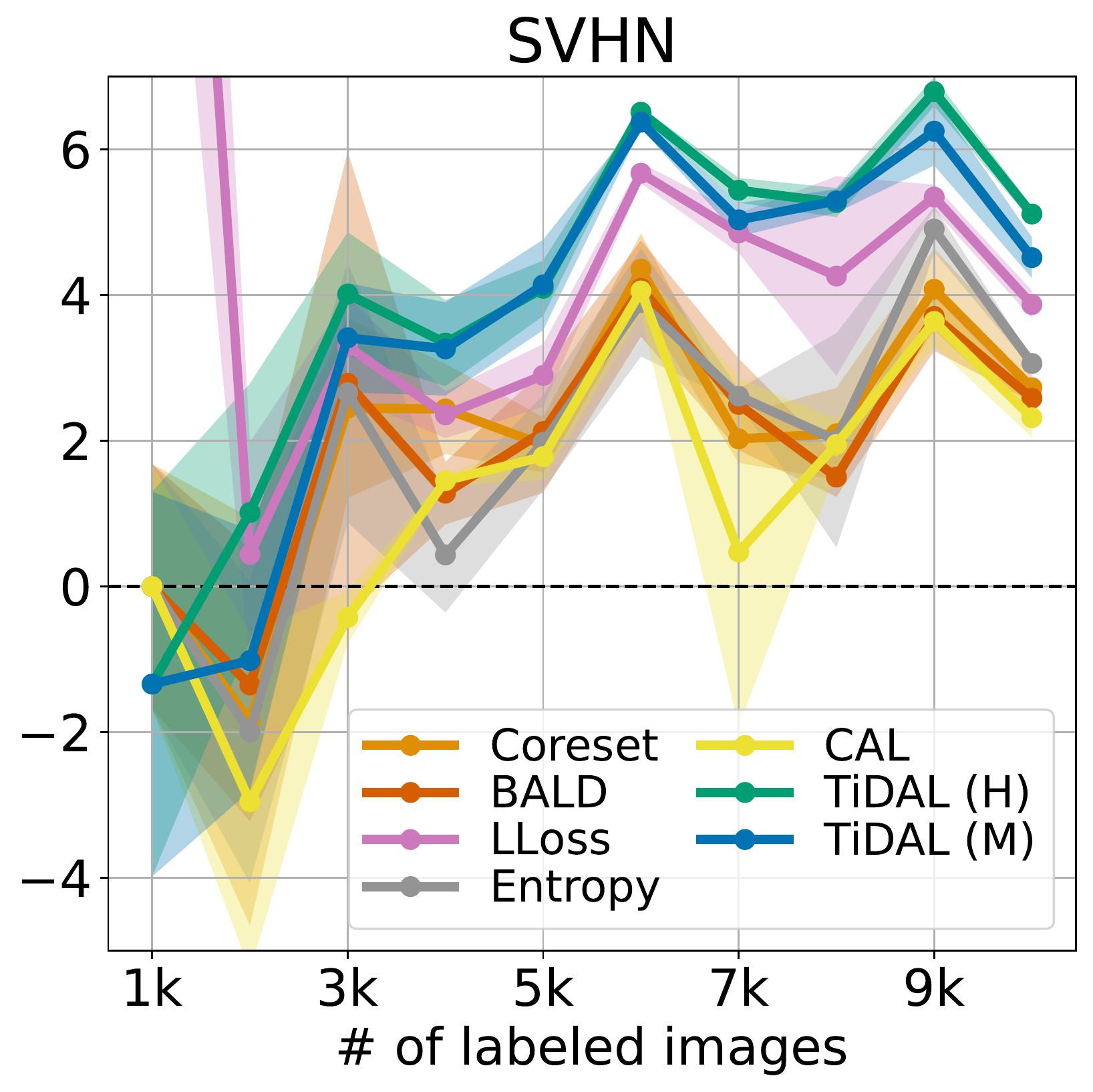}}
\caption{
    Averaged relative accuracy improvement curves and its 95\% confidence interval (shaded) of AL methods over the number of labeled samples on real-world imbalanced datasets: iNaturalist2018 and SVHN.
    For SVHN, LLoss shows a substantial improvement of 20.02\% $\pm$ 6.77\% at the initial phase (1k), but we clip the plot to show the performance afterward more clearly.
}
\label{fig:4_imbalanced_real_sota}
\end{figure*}

%% file: Figures/5_Ablation.tex
\begin{figure*}[t] 
\centering
\begin{minipage}{0.58\textwidth}
\centering
\vspace*{-2.5mm}
\subfloat[CIFAR10]{\includegraphics[width=0.5\textwidth]{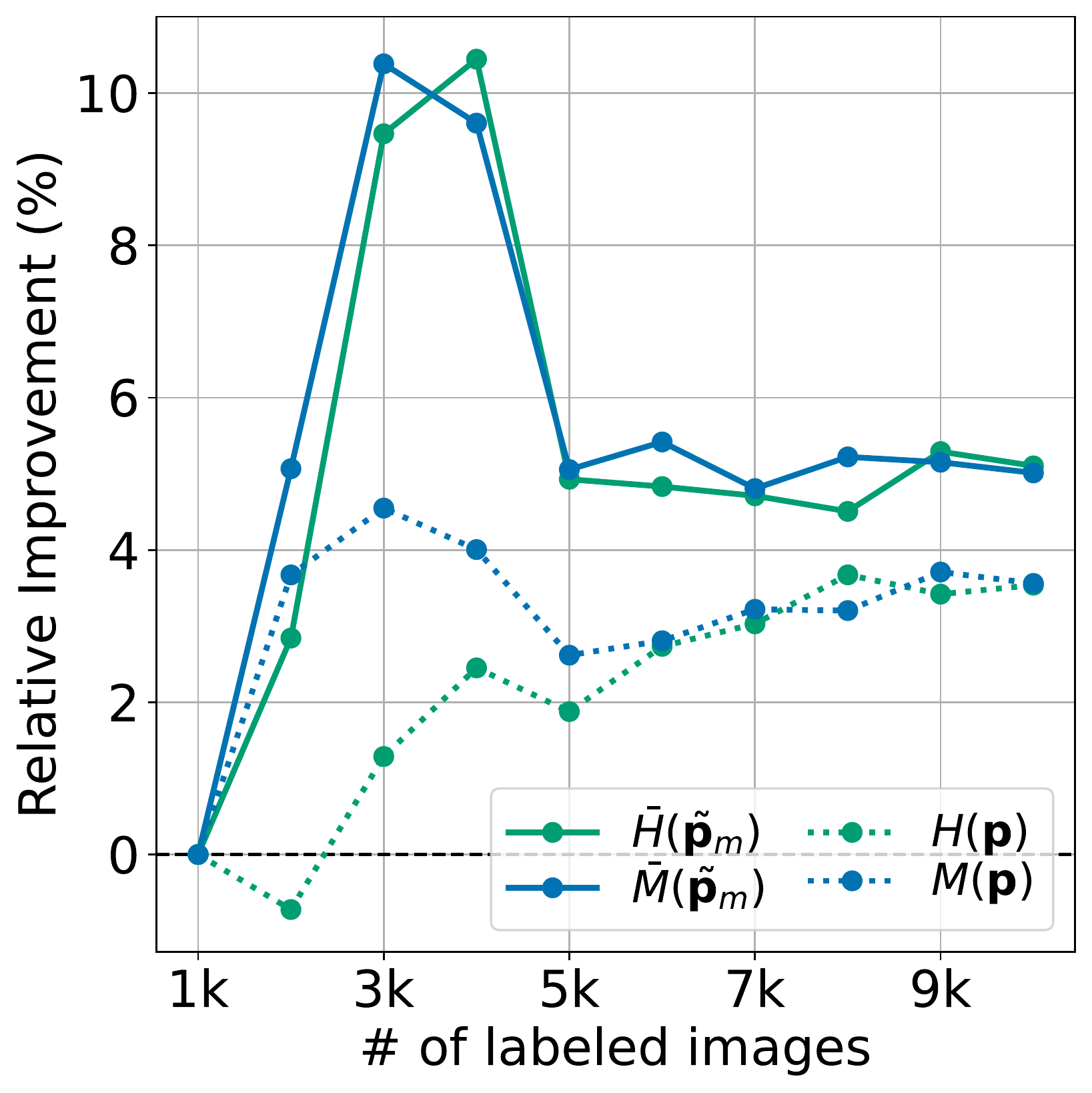}\label{fig:5_1_ablation_cifar10}}
\subfloat[CIFAR100]{\includegraphics[width=0.5\textwidth]{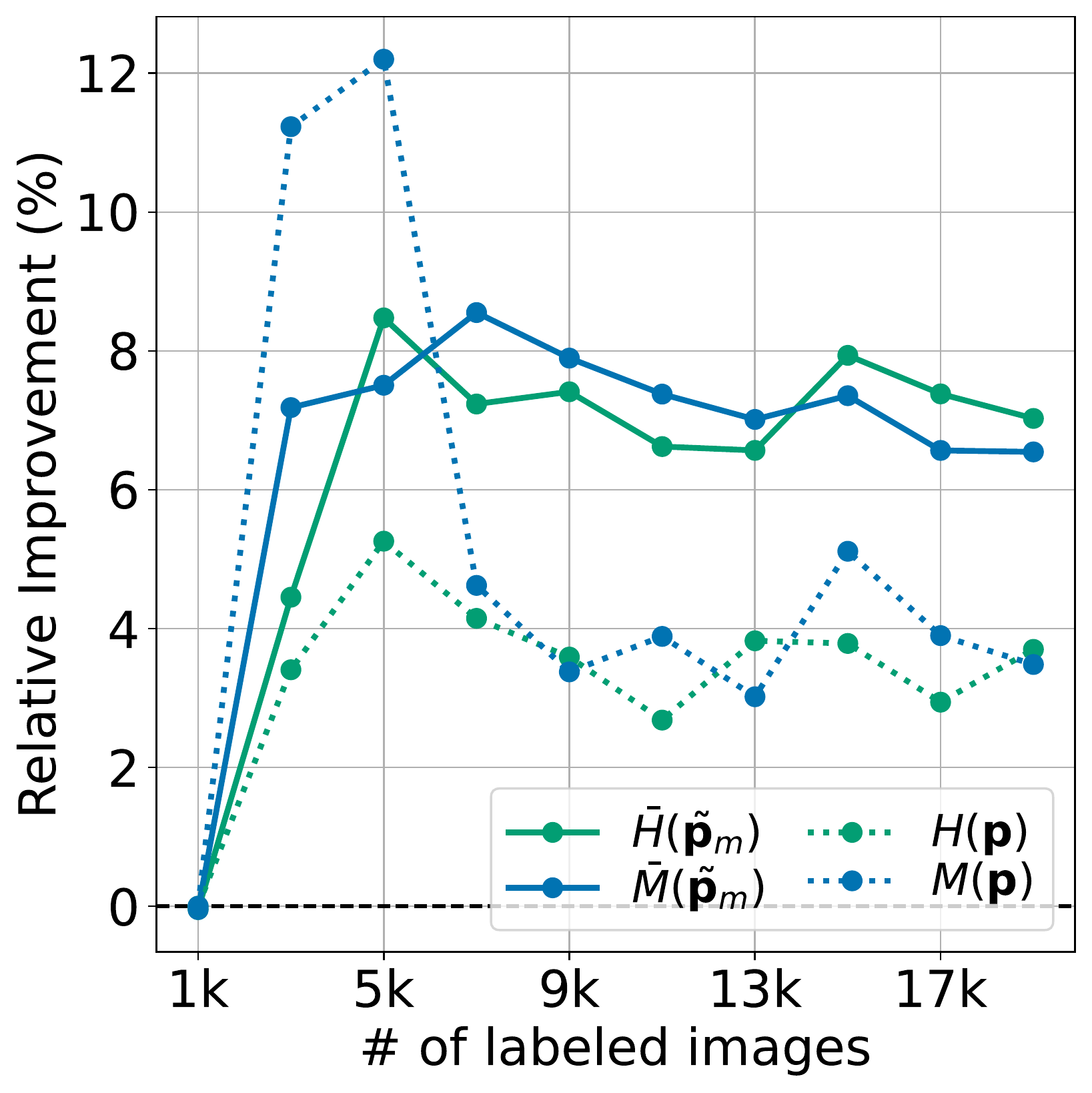}\label{fig:5_2_ablation_cifar100}}
\caption{Ablation test results. $\bar{H}(\Tilde{\vp}_m)$ and $\bar{M}(\Tilde{\vp}_m)$ use the predicted TD $\Tilde{\vp}_m$ of the prediction module $m$. In contrast, $H(\vp)$ and $M(\vp)$ use the predicted probability of the model snapshot $\vp$.
TD shows better performance than the model snapshot, implying that TD is better at quantifying data uncertainty.}
\label{fig:5_ablation}
\end{minipage}
\hspace{0.02\textwidth}
\begin{minipage}{0.38\textwidth}
\centering
\includegraphics[width=0.94\textwidth]{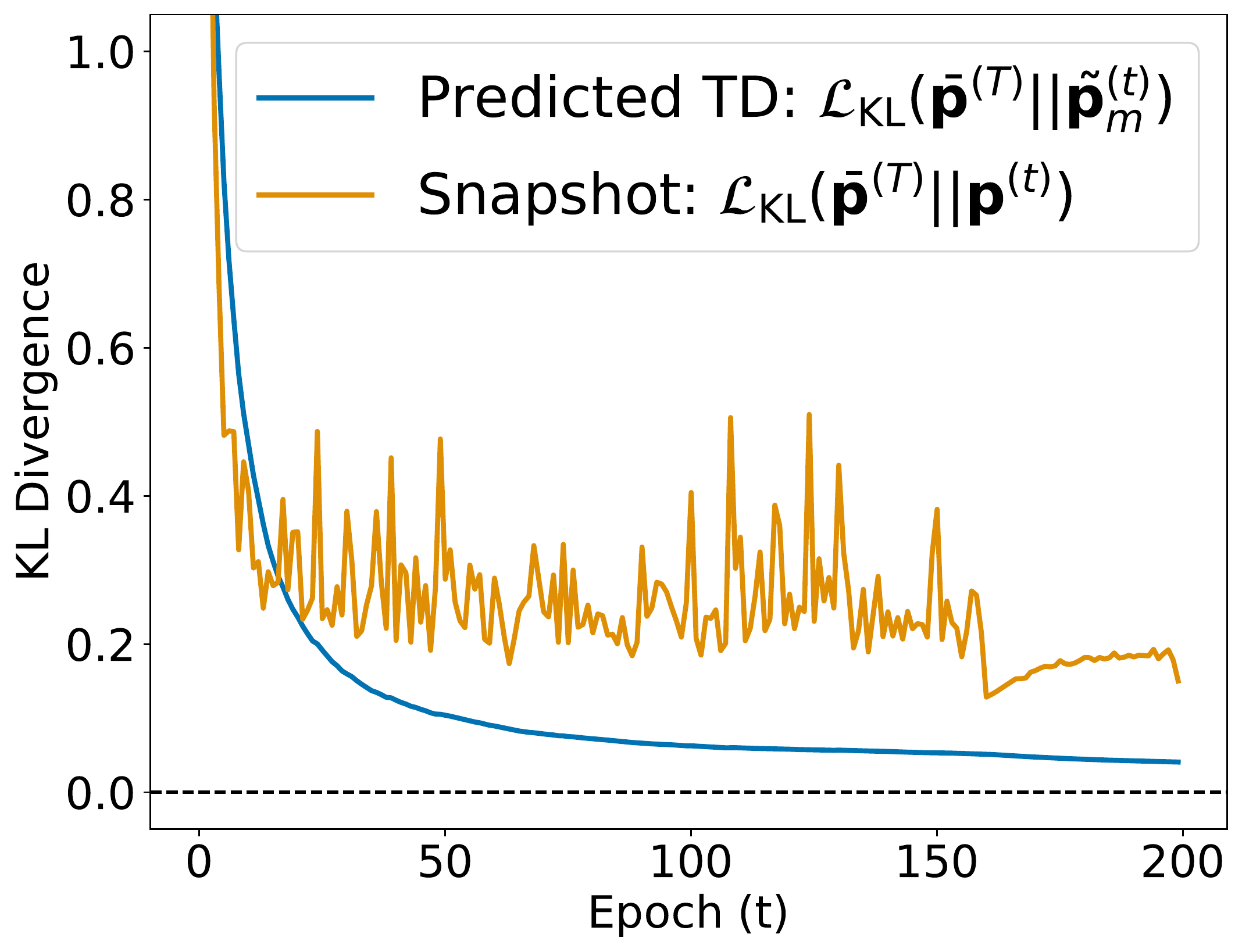}
\caption{KL divergence scores of the actual TD $\bar{\vp}^{(T)}$ with the predicted TD $\Tilde{\vp}^{(t)}_m$ and the predicted probability of the model snapshot $\vp^{(t)}$, respectively, during model optimization. Our predicted TD can accurately approximate the actual TD.}
\label{fig:6_kl_divergence}
\end{minipage}
\end{figure*}

%% file: Sections/4_Related_Work.tex
\section{Related Work}\label{sec:2_related_work}
\subsection{Active Learning}\label{subsec:2_1_active_learning}
AL methods target to construct a dataset with the most useful samples based on the assumption that each sample has different importance in model training \citep{ren2021survey}.
Two mainstream AL approaches exist for efficiently querying the unlabeled data: pool-based methods \citep{lewis1994sequential,yoo2019learning,sinha2019variational} use various ways to extract samples from an unlabeled data pool effectively, and synthesis-based methods \citep{angluin1988queries,zhu2017generative,tran2019bayesian} generate informative samples for the model.
Pool-based methods can be roughly divided based on query strategies: uncertainty-based \citep{gal2017deep,yoo2019learning,sinha2019variational,huang2021semi} and diversity-based \citep{sener2018active,gissin2019discriminative,parvaneh2022active} methods, where some methods use the hybrid of both \citep{ash2019deep,shui2020deep,kim2021task}.
Uncertainty-based methods focus on finding which samples would be the most uncertain for the model, whereas diversity-based methods aim to construct a subset of representative samples of the input distribution.
Our proposed method, TiDAL, lies in uncertainty-based methods.
The significant difference between TiDAL and previous uncertainty-based methods is that TiDAL estimates data uncertainty using TD that contains additional hints generated during model training. 
In contrast, the previous methods leverage only static information (e.g., loss~\citep{yoo2019learning,huang2021semi} and predicted probabilities~\citep{gal2017deep,sinha2019variational,kim2021task}) obtained by a model snapshot at the data selection phase.

\subsection{Training Dynamics}\label{subsec:2_2_training_dynamics}
TD focuses on how deep neural networks are optimized under back-propagation-based stepwise weight updates.
Many studies try to understand how gradient descent can effectively obtain the global minimum by analyzing the loss landscape of neural networks \citep{kawaguchi2016deep,li2018visualizing} or its loss trajectory \citep{arora2018convergence}.
Some also import alternative models that are more mathematically approachable to analyze, such as neural tangent kernels \citep{jacot2018neural}, deep Gaussian processes \citep{lee2018deep}, or stochastic differential equations \citep{zhang2021imitating}.
On the other hand, the phenomenological and practical viewpoint of TD also exists.
\citet{toneva2018empirical} coin the term Forgetting Dynamics to assert that unforgettable samples are often less helpful, and \citet{chang2017active} show that the model could prefer samples that are often wrongly predicted throughout model training.
TD is also commonly used in noisy label literature to find potential noisy labels as they tend to fit later on model training \citep{arazo2019unsupervised,pleiss2020identifying} or locate samples that can be relabeled correctly \citep{song2019selfie}.
Furthermore, \citet{zhou2020curriculum} calculate the Dynamic Instance Hardness score by monitoring losses of each sample or whether the prediction gets flipped so that higher scored samples can be prioritized for curriculum learning, and \citet{jiang2018mentornet} feed the loss history to the auxiliary neural network to mediate the curriculum for training.
\citet{samuli2017temporal} also introduce temporal ensembling for semi-supervised learning, where the model fits towards averaged probability outputs.
\citet{swayamdipta2020dataset,park2022data} devise Data Maps to inspect datasets with two TD measures; confidence and its variability across epochs on the true class prediction.
\citet{zhang2021cartography} further extend the Data Maps for AL, whether the target classifier was consistently correct or not during training.
The proposed method splits the labeled samples by applying a heuristic threshold on the level of consistency to train a binary classifier that is trained to discern uncertain samples.
Even though the work, similar to ours, also utilizes TD, it relies on empirical observations and heuristic choices to separate the certain and uncertain samples.
In this study, we link the concept of TD to AL with both empirical and theoretical results to estimate the uncertainty of unlabeled samples, which is often neglected in previous TD studies.

%% file: Sections/5_Conclusion.tex
\section{Conclusion}\label{sec:5_conclusion}
We propose a novel active learning method, Training Dynamics for Active Learning (TiDAL), by linking the concept of training dynamics to active learning.
We provide motivating observations and theoretical evidence for using training dynamics to estimate the uncertainty of unlabeled data.
Since tracking the training dynamics of large-scale unlabeled data is infeasible, TiDAL utilizes a training dynamics prediction module to efficiently predict the training dynamics of the unlabeled data.
Based on the predicted training dynamics, TiDAL quantifies data uncertainty using the common uncertainty estimators: entropy and margin.
Extensive experiments on multiple benchmark datasets demonstrate the effectiveness of our method, surpassing the existing state-of-the-art active learning methods.
We further analyze that using our training dynamics prediction module is effective and the module successfully predicts the TD of unlabeled data.

%% file: Sections/A_Evidence.tex
\section{Details on the Theoretical Evidence}\label{appendix:a_1_theoretical_guarantee}

\subsection{Proof of Theorem \ref{theorem:1_elastic}}
We adopt the settings of \citep{zhang2021imitating}, with a slight modification of the assumption that sample-level local elasticity affects the training dynamics instead of class-level local elasticity.

Consider the binary classification problem with two classes $k=1, 2$ where class $1$ consists of both certain (easy) and uncertain (hard) samples and class $2$ only consists of samples with the same certainty (easiness).
Let $\mathcal{S}_{1,e}$, $\mathcal{S}_{1,h}$ and $\mathcal{S}_{2}$ denote the easy samples from class $1$, hard samples from class $1$, and samples from class $2$ respectively, which constitutes the partition of the whole set of training samples $\mathcal{S}$: $\mathcal{S} = \mathcal{S}_{1,e} \cup \mathcal{S}_{1,h} \cup \mathcal{S}_{2}$.
Let the corresponding sample sizes be $n_{1, e} = |\mathcal{S}_{1,e}|$, $n_{1, h} = |\mathcal{S}_{1,h}|$, $n_{2} = |\mathcal{S}_{2}|$ and $n = |\mathcal{S}| = n_{1,e} + n_{1,h} + n_{2}$, respectively.

At each iteration $m$, a training candidate sample $J_{m} \in \mathcal{S}$ with class $L_{m}$ is sampled uniformly from the whole training set $\mathcal{S}$ with replacement.
Training using this sample $J_m$ via SGD affects the training dynamics of other samples $s \in \mathcal{S}$ of class $k$ as:
\begin{equation}
    X_{s}^{k}(m) = X_{s}^{k}(m-1) + h E_{s, J_{m}} X_{J_{m}}^{L_{m}}(m-1) + \sqrt{h} \zeta_{s}^{k}(m-1), \label{equation:appendix_a:training_dynamics}
\end{equation}
where $X > 0$ is logit of the true label, $h > 0$ is the step size, $\zeta \sim \mathcal{N}(0, \sigma^{2})$ denotes the noise term arises during training, and $E \in \mathbb{R}^{|\mathcal{S}| \times |\mathcal{S}|}$ refers to the sample-level local elasticity \citep{he2019local} where each entry $E_{s, s'}$ measures the strength of the local elasticity of $s'$ by $s$.
For simplicity, we assume this local elasticity does not depend on the time step $m$.
Furthermore, we consider that the sample-level local elasticity only depends on the set $\mathcal{S}_{1,e}$, $\mathcal{S}_{1,h}$ and $\mathcal{S}_{2}$ in which each samples are in.

Let 
\begin{equation}
    \bar{X}^{1, e}(t) = \frac{1}{n_{1,e}}\sum_{s \in \mathcal{S}_{1,e}} X_{s}^{1}(t), \bar{X}^{1, h}(t) = \frac{1}{n_{1,h}}\sum_{s \in \mathcal{S}_{1,h}} X_{s}^{1}(t), \bar{X}^{2}(t) = \frac{1}{n_{2}} \sum_{s \in \mathcal{S}_{2}} X_{s}^{2}(t) \label{equation:appendix_a:averaged_dynamics}
\end{equation} be the averaged logits for certain samples in class $1$, uncertain samples in class $1$, and class $2$ respectively.

Regarding the strength of local elasticity between ``class'' of samples, for some constants $\alpha_{e}$, $\alpha_{h}$ and $\beta$, we set the value of $E_{s, s'}$ to model sample-level local elasticity for (1) between easy and hard samples in the class $1$ and (2) between classes $1$ and $2$.
We use the values $\alpha_{e} > \alpha_{h} > \beta > 0$ to define the easiness such that the power exerted by sample-level local elasticity between easy samples are stronger for the pair of easy samples than for the pair consists of one or more hard sample.

\begin{itemize}
    \item $E_{s, s'} = \alpha_{e}$ if $(s, s') \in (\mathcal{S}_{1,e} \times \mathcal{S}_{1,e}) \cup (\mathcal{S}_{2} \times \mathcal{S}_{2})$,
    \item $E_{s, s'} = \alpha_{h}$ if $(s, s') \in (\mathcal{S}_{1,e} \times \mathcal{S}_{1,h}) \cup (\mathcal{S}_{1,h} \times \mathcal{S}_{1,e}) \cup (\mathcal{S}_{1,h} \times \mathcal{S}_{1,h})$ (either $s \in \mathcal{S}_{1,h}$ or $s' \in \mathcal{S}_{1,h}$),
    \item $E_{s, s'} = \beta$ otherwise.
\end{itemize}
Intuitively, one can interpret the above assumption as easy samples being clustered with each other \citep{jiang2018trust,papernot2018deep}, hence having a stronger influence on each other due to the local elasticity.
On the contrary, hard samples are often distant from other same-class samples.
Their influence is often limited, as memorizing is easy for the neural nets due to their large capacity \citep{zhang2021understanding}.
Finally, we ignore the influence of other class samples in this proof for simplicity, as we are only considering the logits of the true label.

\setcounter{theorem}{0}
\begin{theorem}\label{theorem:3_elastic_formal}
(Formal) On average, the convergence speed of logit is faster for easy samples than hard samples. Formally:
\begin{equation}
    \frac{d\bar{X}^{1,e}(t)}{dt} > \frac{d\bar{X}^{1,h}(t)}{dt}.
\end{equation}
\end{theorem}

\begin{proof} Fix a target sample $s \in \mathcal{S}$, and execute the dynamics (\ref{equation:appendix_a:training_dynamics}) $r$ times since step $m$. Accumulated change for feature $X$ becomes
\begin{equation}
    X_{s}^{k}(m + r) - X_{s}^{k}(m) = h\sum_{q=1}^{r} E_{k, L_{m+q}} X_{J_{m+q}}^{L_{m+q}}(m+q-1) + \epsilon_{s,k,r,h}, \label{equation:appendix_a:accumulated_dynamics}
\end{equation}
where $\epsilon = \sqrt{h}\sum_{q=1}^{r}\zeta_{s}^{k}(m+q-1) \sim \mathcal{N}(0, \sigma^{2} rh)$ is the accumulated noise terms during $r$ updates. Regarding terms inside the summation, we can divide cases based on which sample $J_{r}$ (with corresponding class $L_{r}$) is actually selected as a training candidate at iteration $\nu (=m+q)$:
\begin{equation}
    E_{k, J_{\nu}}X_{J_{\nu}}^{L_{\nu}} (\nu-1) = \mathbf{1}_{J_{\nu} \in \mathcal{S}_{1, e}}E_{k, J_{\nu}}X_{J_{\nu}}^{1} (\nu-1) 
    + \mathbf{1}_{J_{\nu} \in \mathcal{S}_{1, h}}E_{k, J_{\nu}}X_{J_{\nu}}^{1} (\nu-1) 
    + \mathbf{1}_{J_{\nu} \in \mathcal{S}_{2}}E_{k, J_{\nu}}X_{J_{\nu}}^{2} (\nu-1) ,
\end{equation}
hence the summand from (\ref{equation:appendix_a:accumulated_dynamics}) becomes (omitting time index for $X$ for simplicity)
\begin{equation}
    h\sum_{q=1}^{r} \left( \mathbf{1}_{J_{m+q} \in \mathcal{S}_{1, e}}E_{k, J_{m+q}}X_{J_{m+q}}^{1} 
    + \mathbf{1}_{J_{m+q} \in \mathcal{S}_{1, h}}E_{k, J_{m+q}}X_{J_{m+q}}^{1}
    + \mathbf{1}_{J_{m+q} \in \mathcal{S}_{2}}E_{k, J_{m+q}}X_{J_{m+q}}^{2} \right), \nonumber
\end{equation}
and for sufficiently large $r$ we can approximate the summations as the sample-average dynamics:
\begin{align}
    &h\sum_{q=1}^{r} \left( \mathbf{1}_{J_{m+q} \in \mathcal{S}_{1, e}}E_{k, J_{m+q}}X_{J_{m+q}}^{1} 
    + \mathbf{1}_{J_{m+q} \in \mathcal{S}_{1, h}}E_{k, J_{m+q}}X_{J_{m+q}}^{1} 
    + \mathbf{1}_{J_{m+q} \in \mathcal{S}_{2}}E_{k, J_{m+q}}X_{J_{m+q}}^{2} \right) \nonumber \\
    &\approx h r \left( \mathbb{P}\left(J \in \mathcal{S}_{1,e}\right) \frac{\sum_{s \in \mathcal{S}_{1,e}}E_{k, s}X_{s}^{1} }{n_{1,e}} +  \mathbb{P}\left(J \in \mathcal{S}_{1,h}\right) \frac{\sum_{s \in \mathcal{S}_{1,h}}E_{k, s}X_{s}^{1} }{n_{1,h}} +  \mathbb{P}\left(J \in \mathcal{S}_{2}\right) \frac{\sum_{s \in \mathcal{S}_{2}}E_{k, s}X_{s}^{2} }{n_{2}}\right) \nonumber \\
    &\approx h r \left( \frac{n_{1,e}}{n} \frac{\sum_{s \in \mathcal{S}_{1,e}}E_{k, s}X_{s}^{1} }{n_{1,e}} +  \frac{n_{1,h}}{n} \frac{\sum_{s \in \mathcal{S}_{1,h}}E_{k, s}X_{s}^{1} }{n_{1,h}} +  \frac{n_{2}}{n} \frac{\sum_{s \in \mathcal{S}_{2}}E_{k, s}X_{s}^{2} }{n_{2}}\right) \label{equation:appendix_a:approximated_deterministic_dynamics}
\end{align}

As the components of $E$ only depend on the subset sample relies, we can rewrite accumulated dynamics of logits (\ref{equation:appendix_a:accumulated_dynamics}) for three cases separately, utilizing the notation of averaged logit (\ref{equation:appendix_a:averaged_dynamics}):
\begin{align}
    X_{s}^{1, e}(m+r) - X_{s}^{1, e}(m) &= hr \left( \frac{n_{1,e}}{n} \alpha_{e} \bar{X}^{1,e}(m) + \frac{n_{1,h}}{n}\alpha_{h} \bar{X}^{1,h}(m) + \frac{n_{2}}{n}\beta \bar{X}^{2}(m) \right) + \epsilon_{s, k, r, h} \nonumber \\
    X_{s}^{1, h}(m+r) - X_{s}^{1, h}(m) &= hr \left( \frac{n_{1,e}}{n} \alpha_{h} \bar{X}^{1,e}(m) + \frac{n_{1,h}}{n}\alpha_{h} \bar{X}^{1,h}(m) + \frac{n_{2}}{n}\beta \bar{X}^{2}(m) \right) + \epsilon_{s, k, r, h} \nonumber \\
    X_{s}^{2}(m+r) - X_{s}^{2}(m) &= hr \left( \frac{n_{1,e}}{n} \beta \bar{X}^{1,e}(m) + \frac{n_{1,h}}{n}\beta \bar{X}^{1,h}(m) + \frac{n_{2}}{n}\alpha_{e} \bar{X}^{2}(m) \right) + \epsilon_{s, k, r, h}, \label{equation:appendix_a:sample_discrete_dynamics_per_class}
\end{align}
with a little bit of abbreviated notation for class $1$: $X_{s}^{1, e} = X_{s}^{1}$ for easy sample $s$, and similarly for hard samples. The differential counterpart of the above difference equation is
\begin{align}
    dX_{s}^{1, e}(t) &= \left( \frac{n_{1,e}}{n} \alpha_{e} \bar{X}^{1,e}(t) + \frac{n_{1,h}}{n}\alpha_{h} \bar{X}^{1,h}(t) + \frac{n_{2}}{n}\beta \bar{X}^{2}(t) \right) dt + \sigma dW^{s}(t) \nonumber \\
    dX_{s}^{1, h}(t) &= \left( \frac{n_{1,e}}{n} \alpha_{h} \bar{X}^{1,e}(t) + \frac{n_{1,h}}{n}\alpha_{h} \bar{X}^{1,h}(t) + \frac{n_{2}}{n}\beta \bar{X}^{2}(t) \right) dt + \sigma dW^{s}(t) \nonumber \\
    dX_{s}^{2}(t) &= \left( \frac{n_{1,e}}{n} \beta \bar{X}^{1,e}(t) + \frac{n_{1,h}}{n}\beta \bar{X}^{1,h}(t) + \frac{n_{2}}{n}\alpha_{e} \bar{X}^{2}(t) \right) dt + \sigma dW^{s}(t), \label{equation:appendix_a:sample_sde_per_class}
\end{align}
where $W^{s}(t)$ is standard Wiener process per sample. Averaging each differential equation with respect to each set of samples and ignoring error terms yield a set of simultaneous deterministic differential equations for averaged logits:
\begin{align}
    d\bar{X}^{1, e}(t) &= \left( \frac{n_{1,e}}{n} \alpha_{e} \bar{X}^{1,e}(t) + \frac{n_{1,h}}{n}\alpha_{h} \bar{X}^{1,h}(t) + \frac{n_{2}}{n}\beta \bar{X}^{2}(t) \right) dt \nonumber \\
    d\bar{X}^{1, h}(t) &= \left( \frac{n_{1,e}}{n} \alpha_{h} \bar{X}^{1,e}(t) + \frac{n_{1,h}}{n}\alpha_{h} \bar{X}^{1,h}(t) + \frac{n_{2}}{n}\beta \bar{X}^{2}(t) \right) dt \nonumber \\
    d\bar{X}^{2}(t) &= \left( \frac{n_{1,e}}{n} \beta \bar{X}^{1,e}(t) + \frac{n_{1,h}}{n}\beta \bar{X}^{1,h}(t) + \frac{n_{2}}{n}\alpha_{e} \bar{X}^{2}(t) \right) dt, \label{equation:appendix_a:aggregate_ode_per_class}
\end{align}

To compare the convergence speed of average logit between certain and uncertain samples in the same class $1$, observe that
\begin{equation}
    \frac{d\bar{X}^{1,e}(t)}{dt} - \frac{d\bar{X}^{1,h}(t)}{dt} = \frac{n_{1,e}}{n} (\alpha_{e} - \alpha_{h}) \bar{X}^{1,e}(t) > 0.
\end{equation}
\end{proof}

With additional assumptions on the other class logits being the same, one can also conclude that the estimated probability of the true label will increase steeply during training for the easy samples.
After increasing to some extent, the probability will saturate to one; hence the snapshot model predictions will contain less useful information than monitoring its training dynamics.
However, future work on extending the above theorem is needed.
Starting from the basic idea above, that sample proximity and its amount influence the training dynamics, one can further relax the above assumptions, such as concentrating on the individuality of each sample or considering the changing elasticities during training.
We hope our work ignites the theoretical research on uncertainty from the viewpoint of training dynamics.

\subsection{Proof of Theorem \ref{theorem:2_entropy_td}}
We aim to show the effectiveness of the proposed estimators, entropy (Equation \ref{eq:entropy}) and margin (Equation \ref{eq:margin}), especially in the case where the probabilities converge.
After training, it is commonly observed that the probabilities of the true label of all the samples tend to converge to one, whereas the speed of the convergence differs (Theorem~\ref{theorem:1_elastic}).
Hence, we show that the estimators can effectively discern the differences during training.

For each time step $t$ during training, we have a sequence of predicted probabilities $p^{(t)}(y = c|x)$ corresponds to $t$, for each target class $c = 1, 2, \cdots, C$.
In our paper, we regard the area under the predicted probability $\bar{p}^{(T)}(y=c|x)$ of the sample $x$ as the training dynamics (Equation~\ref{eq:td_vector_def}), which is indeed a well-known metric of area under the curve, except that it is normalized properly to have value between $0$ and $1$.
For convenience, let
\begin{equation*}
    \vs(x) 
    = \left[ \begin{matrix} s_{1}(x) \\ s_{2}(x) \\ \vdots \\ s_{C}(x) \end{matrix}\right]    
    = \left[\begin{matrix} \bar{p}^{(T)}(y=1|x) \\ \bar{p}^{(T)}(y=2|x) \\ \vdots \\ \bar{p}^{(T)}(y=C|x)  \end{matrix}\right]
\end{equation*}
be the vector consisting the area under the prediction curve for each class up to final epoch $T$. By construction, the components in $\vs(x)$ are nonnegative and sum to $1$.

\begin{theorem} (Formal) Assume that all target classes have the same area under the prediction curve except for the true class $y$. 
Suppose two training samples $(x_1,y_1),(x_2, y_2) \in \gD$ satisfies
\begin{itemize}
    \item[a.] $p^{(T)}(y_1|x_1)$=$p^{(T)}(y_2|x_2)$ (same predicted probability at the end of training)
    \item[b.] $\frac{1}{2}<s_{y_{1}}(x_1)<s_{y_{2}}(x_2)$ (but different TD, in terms of the area under the curve)
\end{itemize}

Then, the following inequalities hold:
\begin{enumerate}
    \item $H(\vs(x_1)) > H(\vs(x_2))$;
    \item $M(\vs(x_1)) < M(\vs(x_2))$.
\end{enumerate}

\end{theorem}

\begin{proof} 

By the assumption, for all target class $c$ except the true class $y$, the area under the prediction curve is given by
\begin{equation}
    s_{c}(x) = \frac{1 - s_{y}(x)}{C-1},
\end{equation}

and the corresponding entropy can be calculated as
\begin{equation}
\begin{split}
    H(\vs(x)) &= \sum_{c = 1}^{C} \left( -s_{c}(x)\text{log}(s_{c}(x))\right) \\
    &= - s_y(x)\text{log}s_y(x) - (C-1)\cdot\left(\frac{1-s_y(x)}{C-1}\right)\text{log}\left(\frac{1-s_y(x)}{C-1}\right) \\
    & = -\left\{s_y(x)\text{log}s_y(x) + (1-s_y(x))\text{log}(1-s_y(x))\right\} + (1-s_y(x))\text{log}(C-1)\\
    & = H_2(s_y(x)) + (1-s_y(x))\text{log}(C-1).
\end{split}
\end{equation}
where $H_2(p) = -p\text{log}p - (1-p)\text{log}(1-p)$ stands for the binary entropy function. Since $H_{2}(p)$ is a decreasing function for $p > \frac{1}{2}$,
\begin{equation*}
    H(\vs(x_1)) - H(\vs(x_2)) = \left\{H_2(s_{y_1}(x_1)) - H_2(s_{y_2}(x_2))\right\}+ \left\{s_{y_2}(x_2) - s_{y_1}(x_1)\right\}\text{log}(C-1) > 0,
\end{equation*}
which proves the first inequality stated.

The first assumption also gives the simplified formulation for the margin
\begin{equation}
    M(\vs(x)) = s_y(x) - \frac{1-s_y(x)}{C-1} = \frac{C}{C-1}s_y(x) - \frac{1}{C-1},
\end{equation}
in whicn the second inequality directly follows:
\begin{equation}
    M(\vs(x_1)) - M(\vs(x_2)) = \frac{C}{C-1}\left(s_{y_1}(x_1) - s_{y_2}(x_2)\right) < 0.
\end{equation}

\end{proof}

While the final predicted probabilities $p^{(T)}(y|x)$ of the training samples tend to converge to $1$ for the true class $y$, otherwise $0$, their TD (in this case $\vs(x) = \bar{\vp}^{(T)}$) may be different depending on the easiness of the samples.
Thus, the degree of the easiness of the samples (i.e. uncertainty) could be captured from TD $\bar{\vp}$, whereas the predictions $\vp$ from a model snapshot cannot.

%% file: Sections/B_Pilot_Details.tex
\section{Details on the Motivating Observation}\label{appendix:b_pilot_details}
\input{Figures/9_Pilot_Study_Details.tex}
$\S$\ref{subsubsec:motivation_observation} empirically show that using TD is effective in separating uncertain samples from certain samples.
Before diving into the experimental details, we want to emphasize that it is difficult to control the level of data difficulty (or uncertainty).
First and foremost, human perception of data difficulty will be highly subjective and potentially different from its model counterpart.
This limitation hinders the quantitative analysis, and thus some previous works had to rely on qualitative substitutes or analyze mislabeled samples which are impossible to control its difficulty \citep{pleiss2020identifying,northcutt2021confident,toneva2018empirical}.
Also, even if we could obtain sample-wise difficulty, it is often nontrivial to analyze the overall trend during training due to sheer data size.

To avoid the two challenges above, we borrow the settings from studies on long-tail visual recognition \citep{liu2019large,bengar2022class}.
\citet{cao2019learning} show that generalization error is bounded by the inverse square root of the dataset size.
Further, many long-tail literature \citep{liu2019large,zhou2020bbn,hong2021disentangling} have also empirically shown that it is hard for the deep neural network-based model to train with fewer samples, showing lower accuracy.
Hence, we consider the major and minor classes as certain and uncertain classes, as the binned classification error is often used as the definition of confidence \citep{guo2017calibration}.

We train ResNet-18 \citep{he2016deep} on the CIFAR10 dataset~\citep{krizhevsky2014cifar,cao2019learning} with an imbalance ratio of $10$ for $30$ epochs using the Adam optimizer \citep{kingma2015adam}.
Figure \ref{fig:pilot_label_distrb} shows the label distribution of the training dataset.
Similar to \citet{bengar2022class}, we choose classes $0, 1, 2, 3$ and $4$ as the major class and the rest as the minor class, randomly removing 90\% of the training samples for the minor class.
We reduce the inter-class differences of CIFAR10 by merging five classes into one, and demonstrate both the overall distribution and samplewise scores in Figure \ref{fig:2_pilot_study}.
We conclude that TD successfully captures data uncertainties, where its characteristics are more helpful in separating uncertain samples from certain samples than the information obtained from a model snapshot.
Also, we empirically reaffirm that the major classes being more advantageous than minor classes in terms of accuracy during model training (Figure~\ref{fig:pilot_trainacc}, \ref{fig:pilot_testacc}).

\section{Details on the TD Prediction Module}\label{appendix:module_detail}
One can offer numerous alternatives on the design of the TD prediction module $m$, but we adopt the architecture of the loss prediction module~\citep{yoo2019learning} except for the last layer.
By adopting the architecture used in the previous study, it is intended to show that the performance improvement of TiDAL does not come from adopting an advanced prediction module architecture, but from using TD.
The TD prediction module takes several hidden feature maps extracted between the mid-level blocks of the target classifier $f$ as inputs.
Through a global average pooling layer and a fully-connected layer, each feature map is reduced to a fixed dimensional feature vector.
All the reduced feature vectors are concatenated to take multi-level knowledge of the target classifier into consideration for TD prediction.
Using a single Softmax layer, the TD prediction module outputs a $C$-dimensional prediction $\Tilde{y}^{(t)} \in [0, 1]^C$, which are used as the predicted TD.

For a better understanding of the architecture of our TD prediction module $m$, please refer to \citet{yoo2019learning}.

%% file: Figures/9_Pilot_Study_Details.tex
\begin{figure}[h] 
\centering
\begin{subfigure}{0.33\textwidth}
  \centering
  \resizebox{\textwidth}{!}{\includegraphics[width=\textwidth]{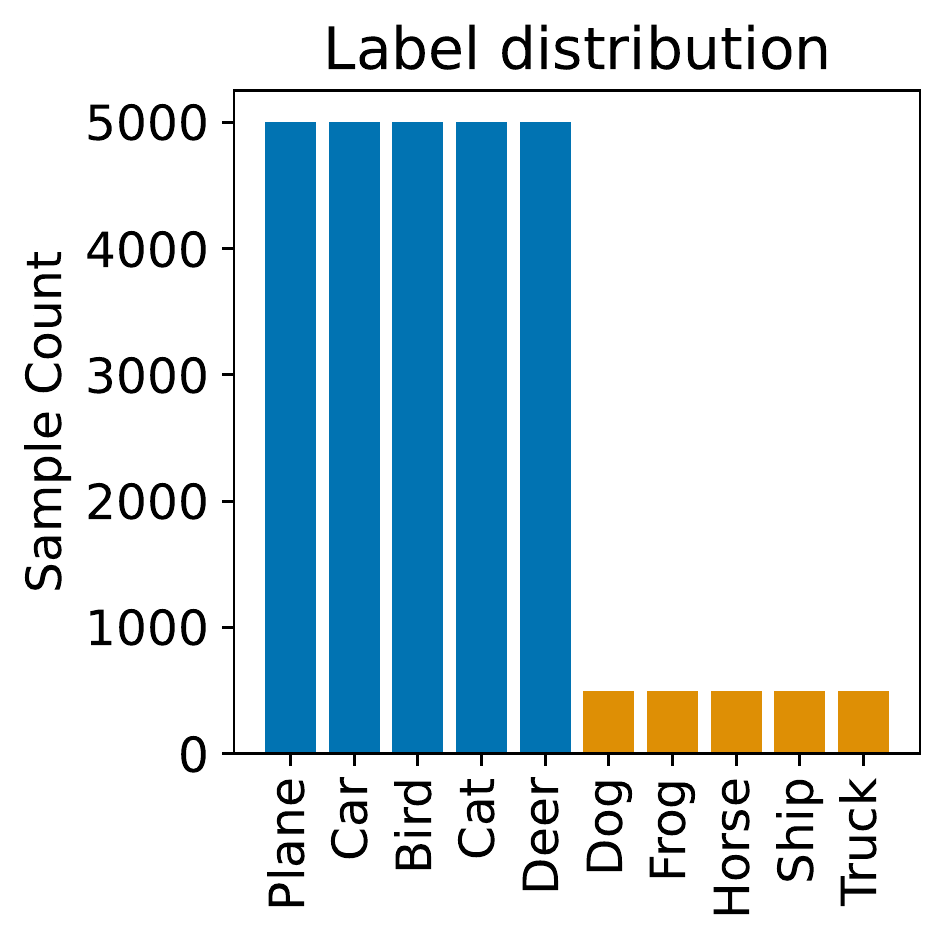}}
  \caption{Training set label distribution.}
  \label{fig:pilot_label_distrb}
\end{subfigure}%
\begin{subfigure}{.33\textwidth}
  \centering
  \resizebox{\textwidth}{!}{\includegraphics[width=\textwidth]{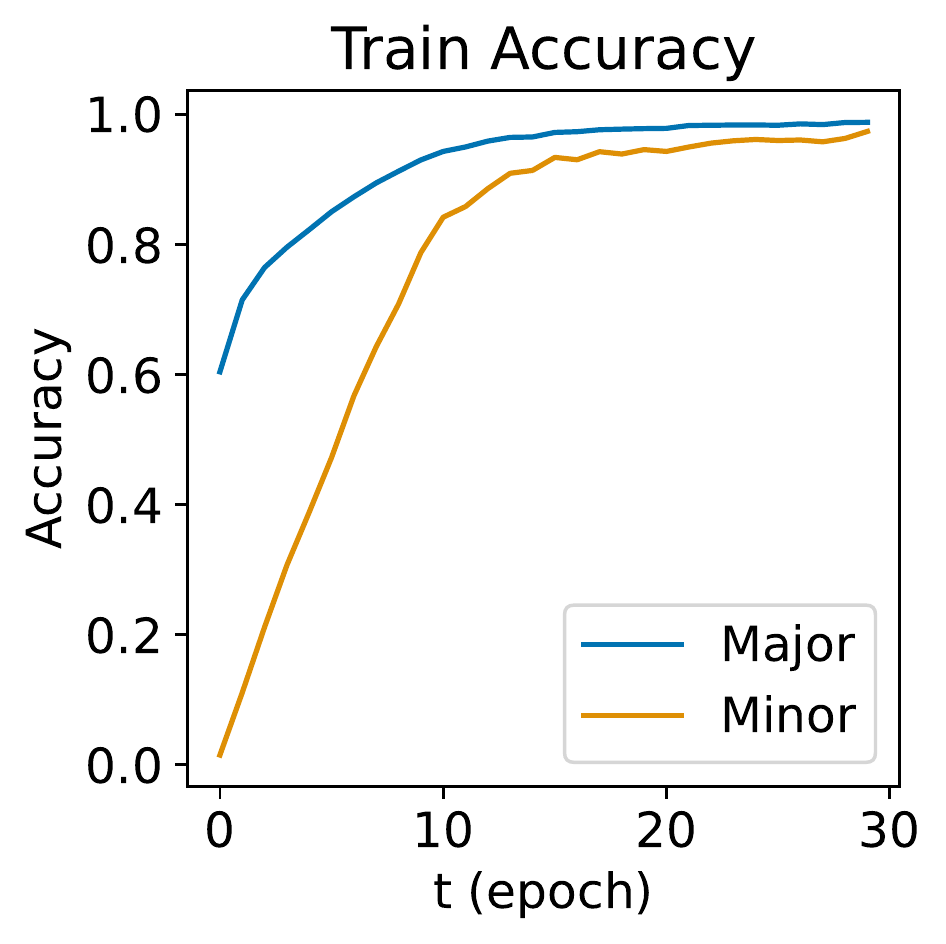}}
  \caption{Training accuracy per epoch.}
  \label{fig:pilot_trainacc}
\end{subfigure}
\begin{subfigure}{.33\textwidth}
  \centering
  \resizebox{\textwidth}{!}{\includegraphics[width=\textwidth]{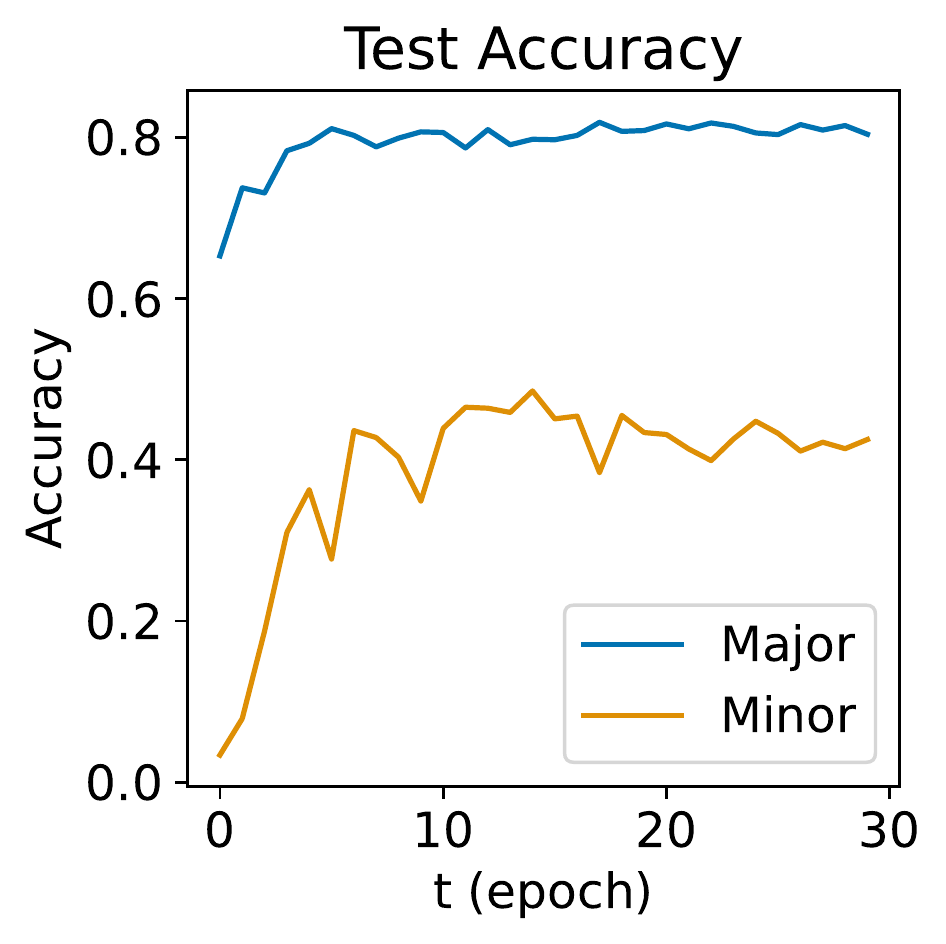}}
  \caption{Test accuracy per epoch.}
  \label{fig:pilot_testacc}
\end{subfigure}
\caption{Training label distribution and accuracy curves for the motivating experiment in $\S$\ref{subsubsec:motivation_observation}.}
\end{figure}

%% file: Sections/C_Further_Analysis.tex
\section{Additional Experiments}\label{appendix:b_expeirments_details}

We conduct additional experiments to further demonstrate the effectiveness of our method, TiDAL. We provide the dataset statistics in Table \ref{tab:dataset_statistics}.

\input{Tables/dataset_statistics}

\subsection{Additional Results on Imbalanced Datasets}
\input{Figures/10_Additional_Imbalanced_Synth_SOTA}

Figure \ref{fig:10_additional_imbalanced_synth_sota} shows the experimental results on the imbalance ratio 100. 
Except for CIFAR10, our methods show superiority over other state-of-the-art methods.

\subsection{Additional Results on Absolute Accuracy}
\input{Figures/13_absolute_sota}

Figure \ref{fig:13_appendix_absolute_real_sota} and \ref{fig:14_appendix_absolute_imbalanced_synth_sota} provides the absolute accuracy plots for the completeness of the evaluation for real and synthetic data, respectively. We can observe the superiority of our method further on many of the settings.

\subsection{Additional Baselines}\label{appendix:vaal}
\input{Figures/11_VAAL_SOTA}
Figure~\ref{fig:11_vaal_sota} compares our TiDAL with VAAL \cite{sinha2019variational} and TA-VAAL \cite{kim2021task}.
Except for the case of CIFAR10 with the imbalance ratio of $100$, both TiDAL strategies excel in performance.
Note that both VAAL and TA-VAAL use a semi-supervised approach to train the selection module and further leverage the unlabeled data for training.

\subsection{Variants of Training Dynamics-Aware Margin}\label{appendix:c_2_variants_margins}
\input{Figures/12_Margin_Variants}
We introduced two TD-aware strategies: entropy $\bar{H}$ and margin $\bar{M}$, in $\S$\ref{subsec:2_2_definition_of_training_dynamics}.
We further demonstrate various uncertainty estimation strategies as follows:
\begin{align}
    &\bar{M}_0(\Tilde{\vp}_m) = \Tilde{p}_m(\Tilde{y}|x) - \max_{c \neq \Tilde{y}} \Tilde{p}_m(c|x),\\
    &\bar{P}(\Tilde{\vp}_m) = \Tilde{p}_m(\hat{y}|x),\\
    &\bar{P}_0(\Tilde{\vp}_m) = \Tilde{p}_m(\Tilde{y}|x),
\end{align}
where $\Tilde{y} = \text{argmax}_c \ \Tilde{\vp}_m(c|x)$ is the class of the maximum module output.

$\hat{M}_0$ is the naive variant of the margin $\hat{M}$ where it does not utilize the predicted label $\hat{y}$ of the target classifier $f$.
It calculates the margin between the biggest and the second biggest outputs of the module $m$.
$\bar{P}$ uses the module output on the predicted label $\hat{y}$ from the target classifier $f$ and $\bar{P}_0$ is the naive variant of $\bar{P}$ that uses the maximum output of the module $m$.

Figure~\ref{fig:12_margin_sota} shows the average accuracy of three runs for the entropy $\bar{H}$ and margin $\bar{M}$, where we show the accuracy of a single run for other strategies.
We can observe that the naive variant of the margin $\bar{M}_0$ generally underperforms compared to the margin $\bar{M}$ except CIFAR100 with the imbalance ratio of $100$.
There seems to be no clear dominance between $\bar{P}$ and its naive variant $\bar{P}_0$.
However, both $\bar{P}$ and $\bar{P}_0$ perform moderately well on both CIFAR100 and FashionMNIST despite its simplicity.
Future studies may concentrate on broader query strategies based on various training dynamics and its module predictions.

%% file: Tables/dataset_statistics.tex
\begin{table}[ht]
\centering
\caption{Statistics of the dataset used for experiments.}
{
\begin{tabular}{l c c c}
\toprule
Dataset & \# of classes & \# of samples & Imbalance ratio \\
\midrule
CIFAR10 & 10 & 50k & \{1, 10, 100\} \\
CIFAR100 & 100 & 50k & \{1, 10, 100\} \\
FashionMNIST & 10 & 60k & \{1, 10, 100\} \\
SVHN & 10 & 73k & 2.98 \\
iNaturalist2018 & 8k & 437k & 500 \\
\bottomrule
\end{tabular}
}
\label{tab:dataset_statistics}
\end{table}

%% file: Figures/10_Additional_Imbalanced_Synth_SOTA.tex
\begin{figure}[t] 
\centering
\subfloat{\includegraphics[width=0.33\columnwidth]{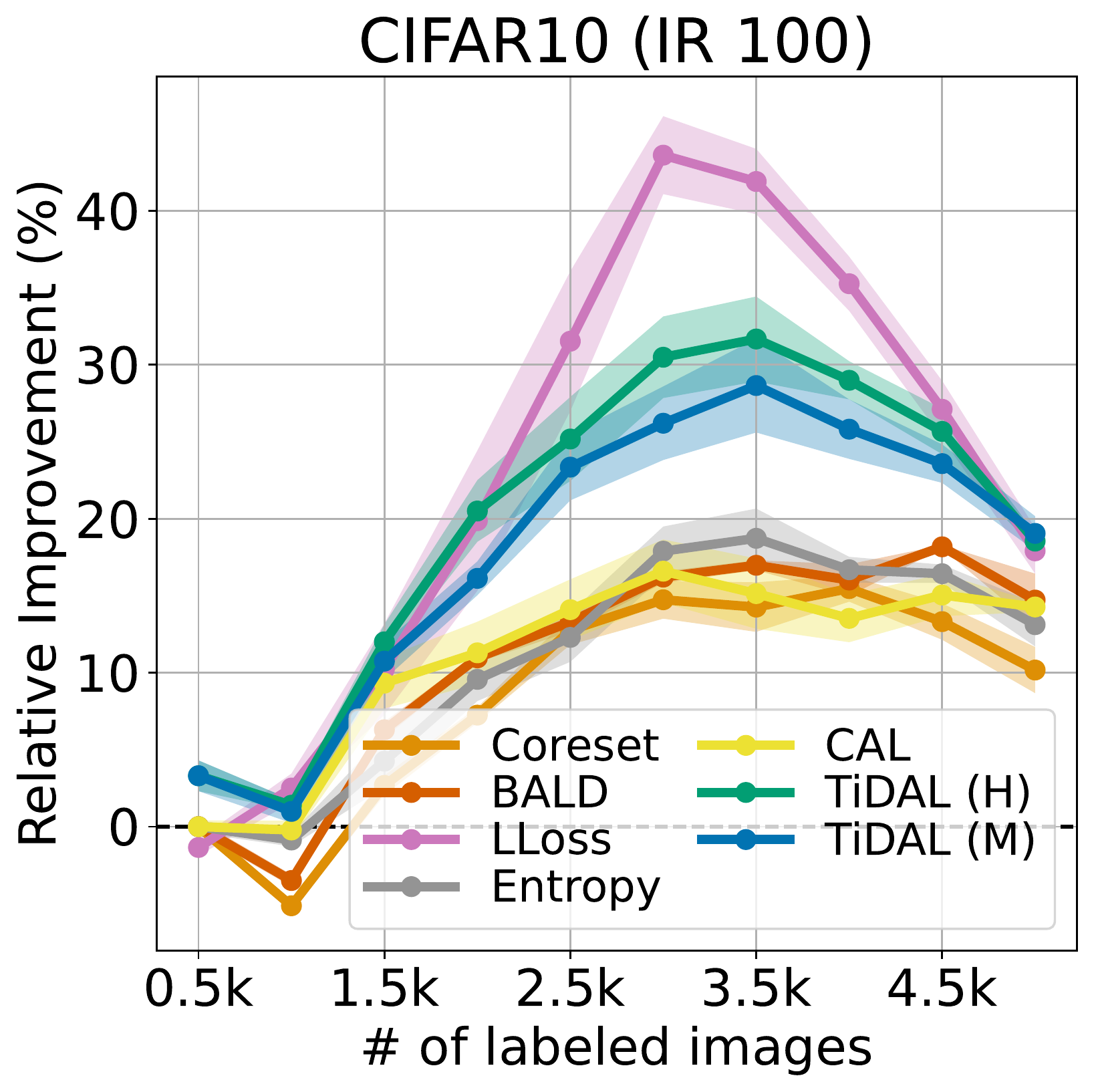}}
\subfloat{\includegraphics[width=0.33\columnwidth]{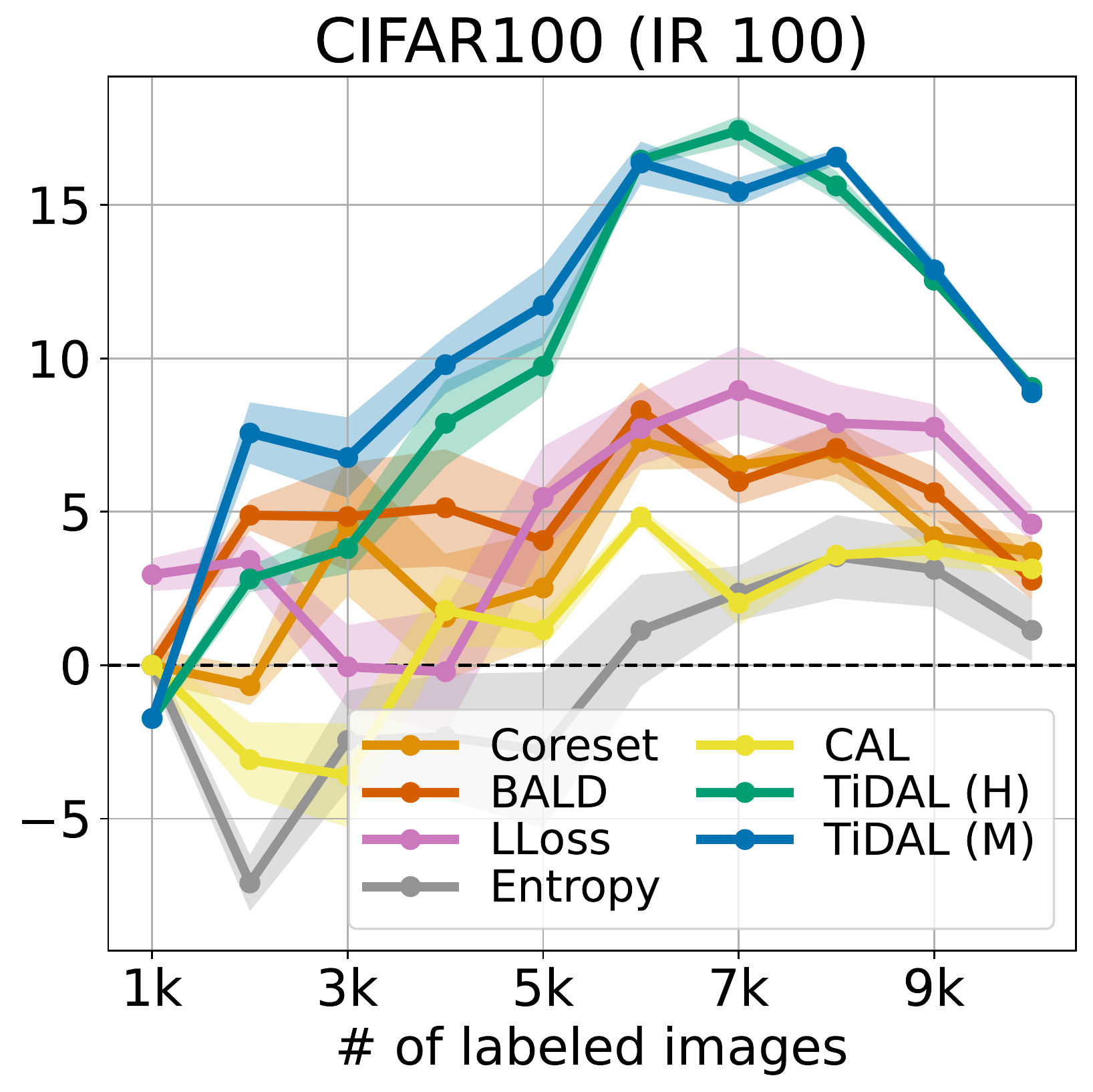}}
\subfloat{\includegraphics[width=0.33\columnwidth]{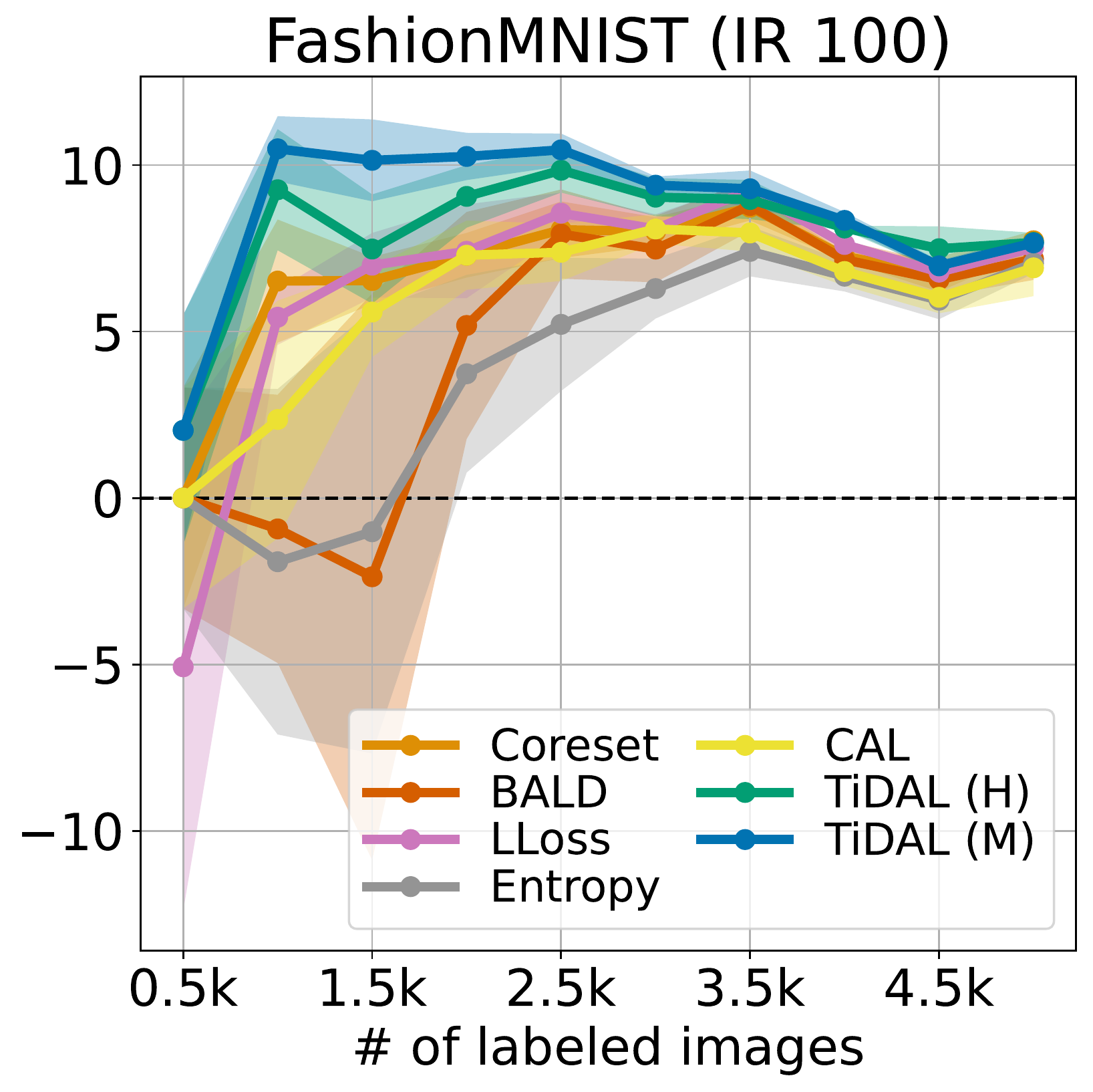}}
\caption{
    Averaged relative accuracy improvement curves and their 95\% confidence interval (shaded) of AL methods over the number of labeled samples on synthetically imbalanced datasets.
    We use the imbalance ratio (IR) of 100 on CIFAR10, CIFAR100, and FashionMNIST.
}
\label{fig:10_additional_imbalanced_synth_sota}
\end{figure}

%% file: Figures/13_absolute_sota.tex
\begin{figure}[t] 
\centering
\subfloat{\includegraphics[width=0.33\columnwidth]{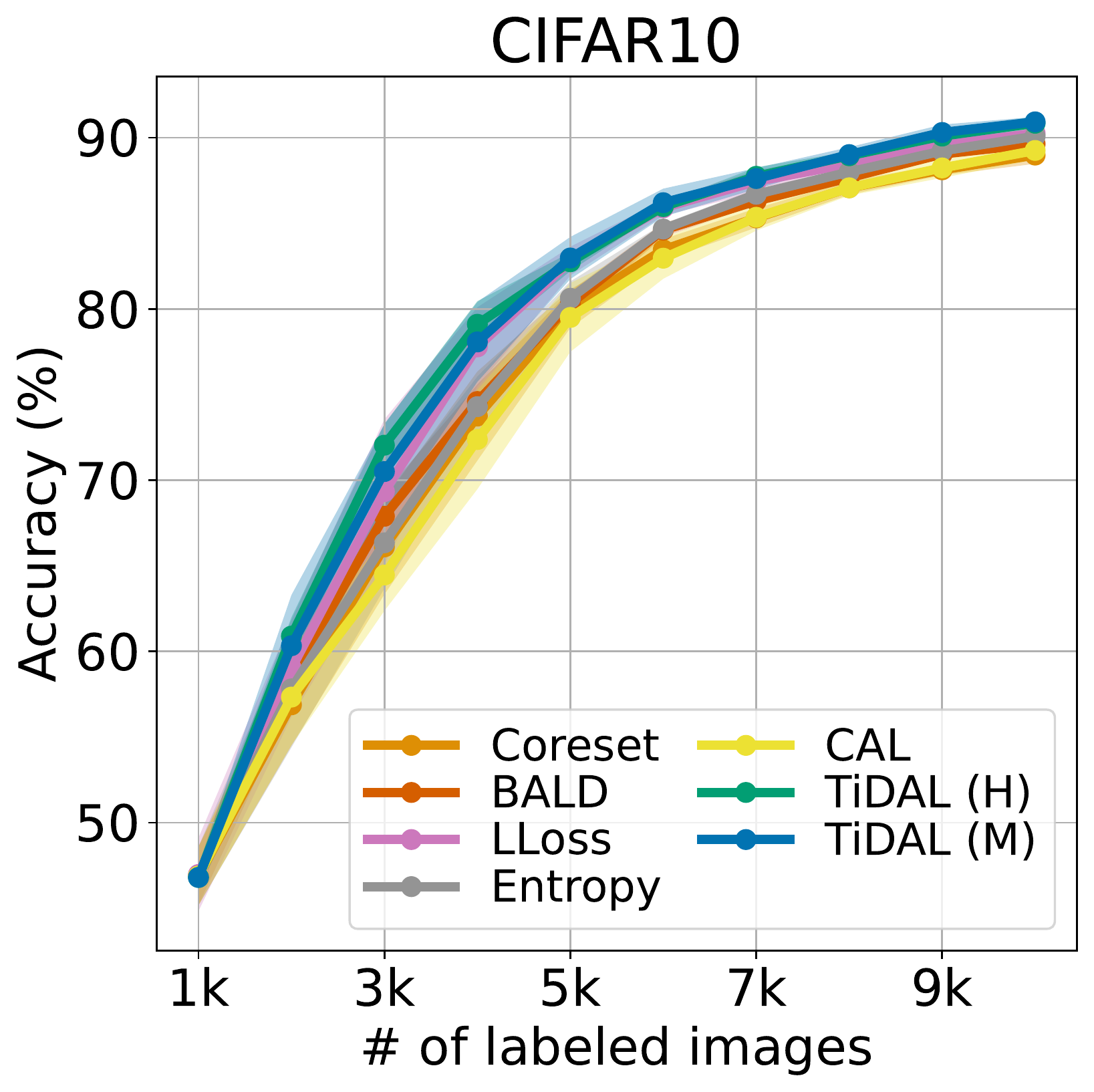}}
\subfloat{\includegraphics[width=0.33\columnwidth]{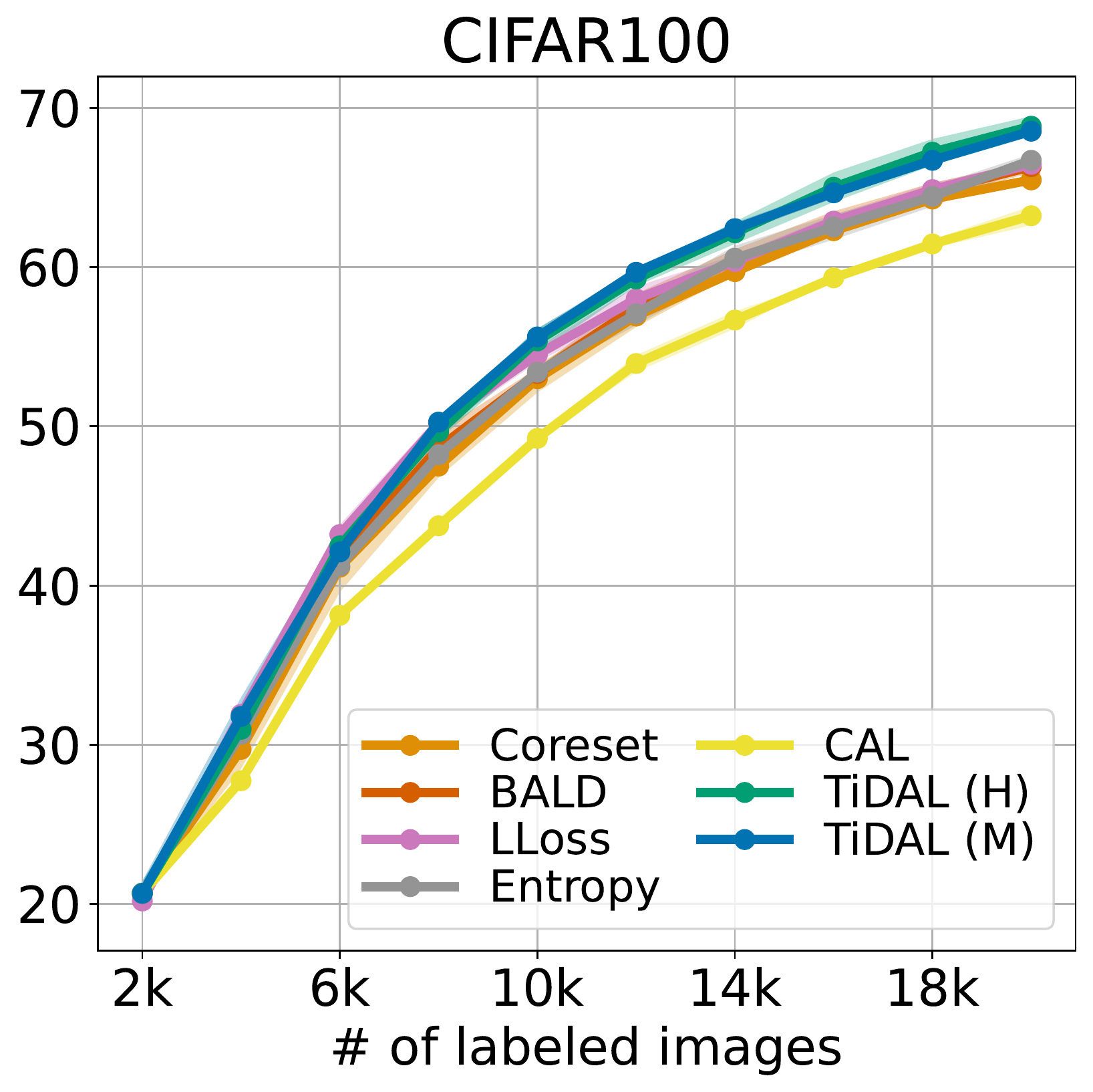}}
\subfloat{\includegraphics[width=0.33\columnwidth]{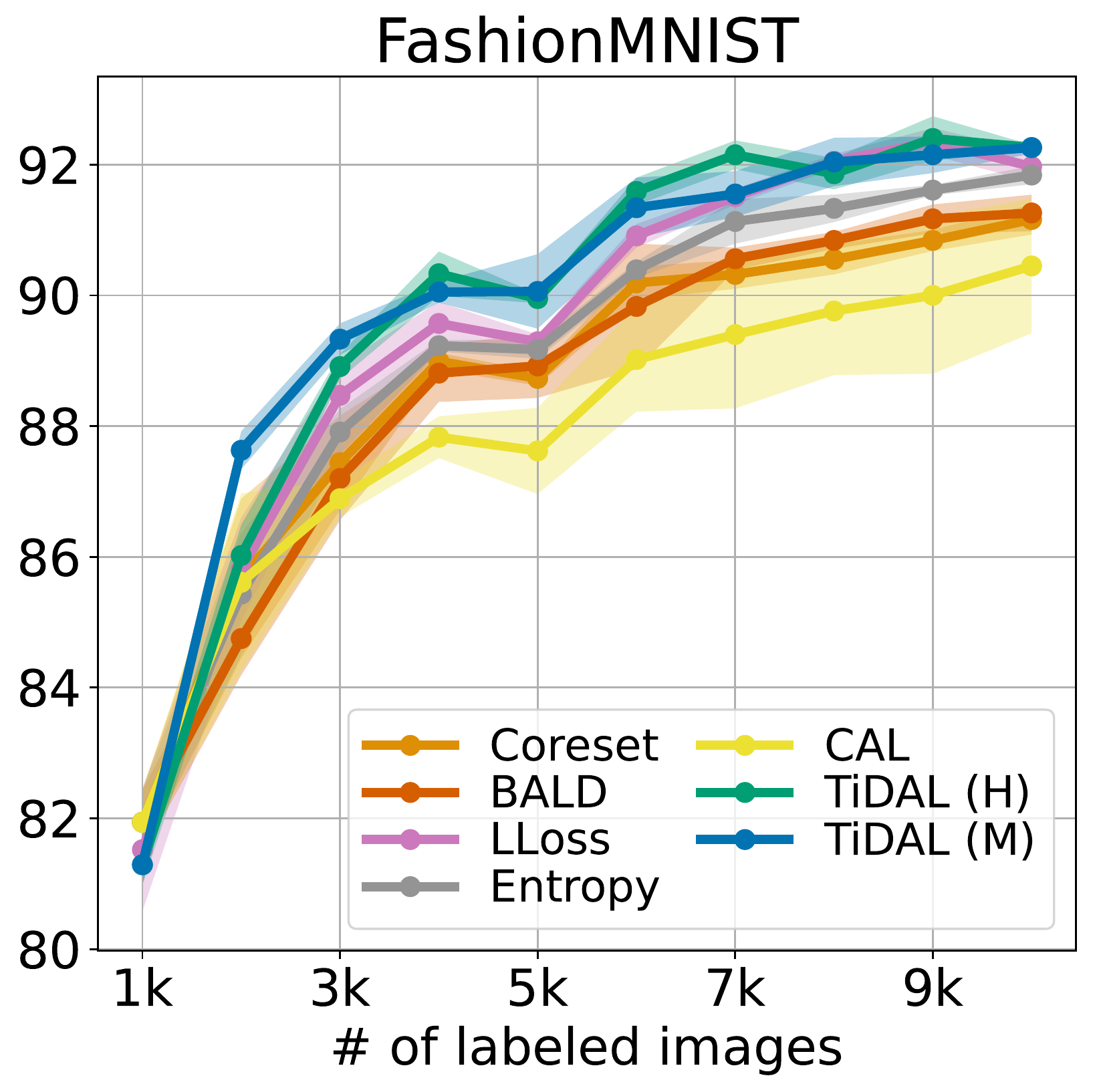}}\\
\subfloat{\includegraphics[width=0.33\columnwidth]{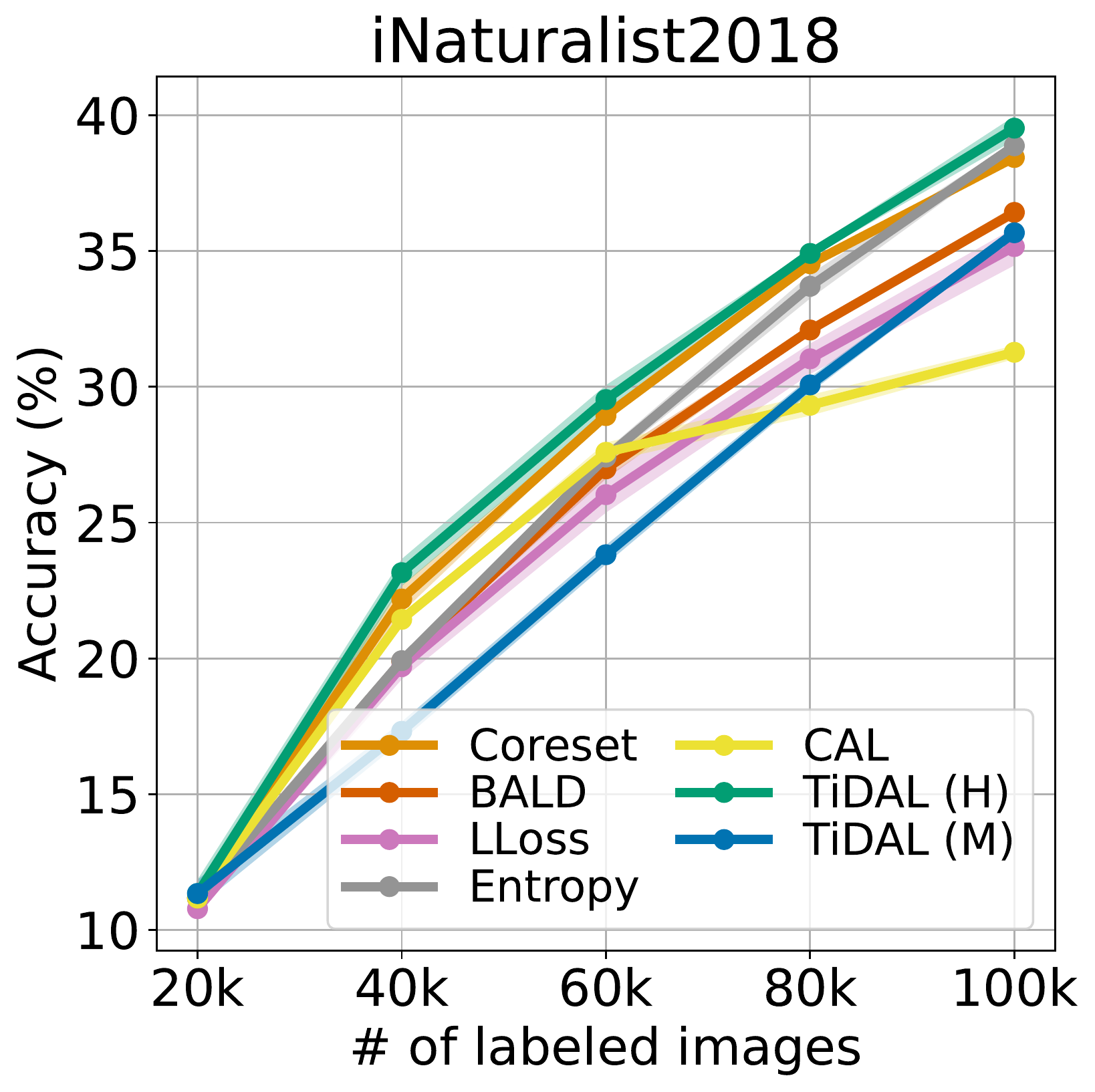}}
\subfloat{\includegraphics[width=0.33\columnwidth]{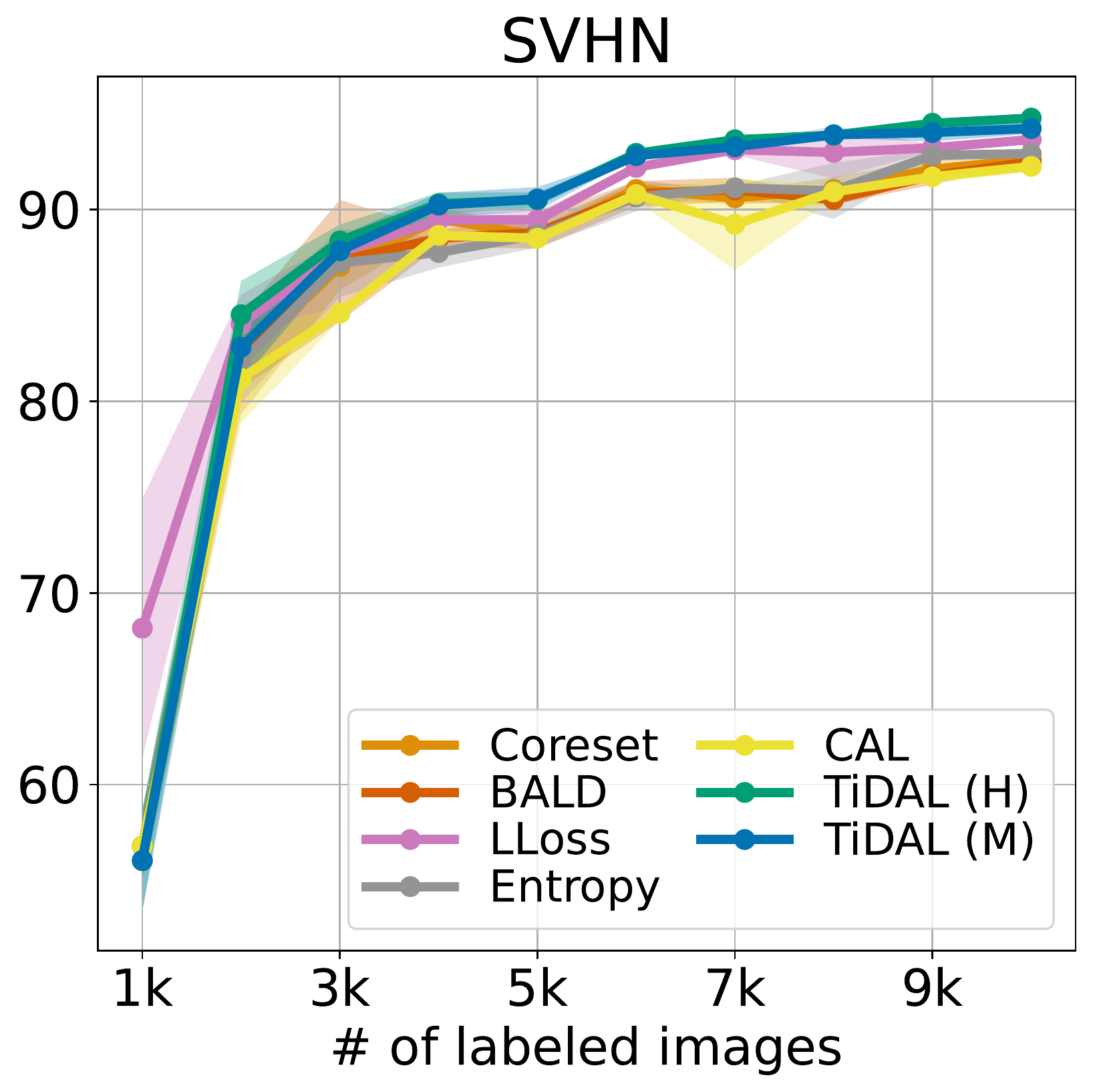}}
\caption{
    Averaged absolute accuracy improvement curves and its 95\% confidence interval (shaded) of AL methods over the number of labeled samples on balanced and imbalanced datasets.
}
\label{fig:13_appendix_absolute_real_sota}
\end{figure}

\begin{figure}[t] 
\centering
\subfloat{\includegraphics[width=0.33\columnwidth]{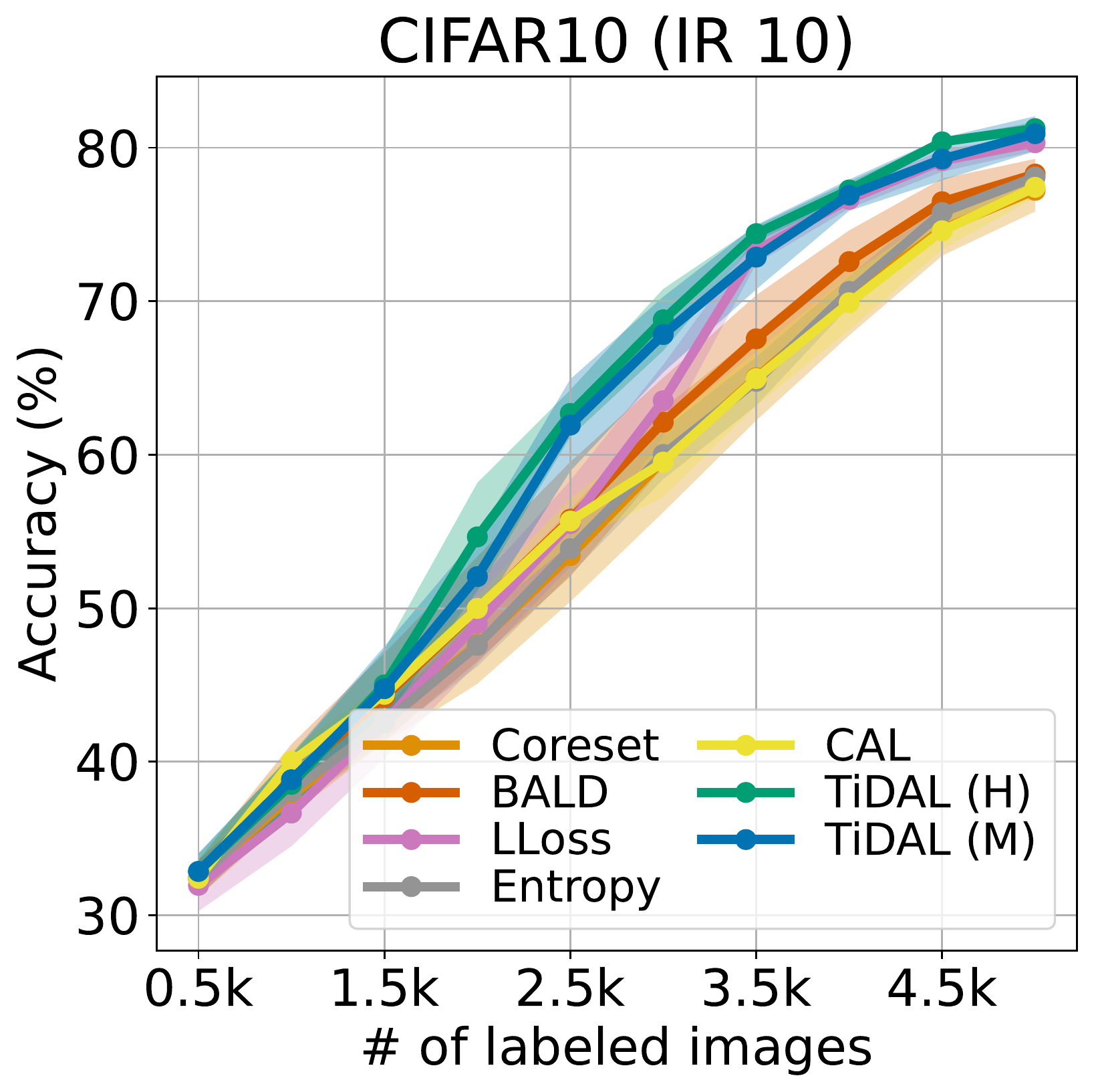}}
\subfloat{\includegraphics[width=0.33\columnwidth]{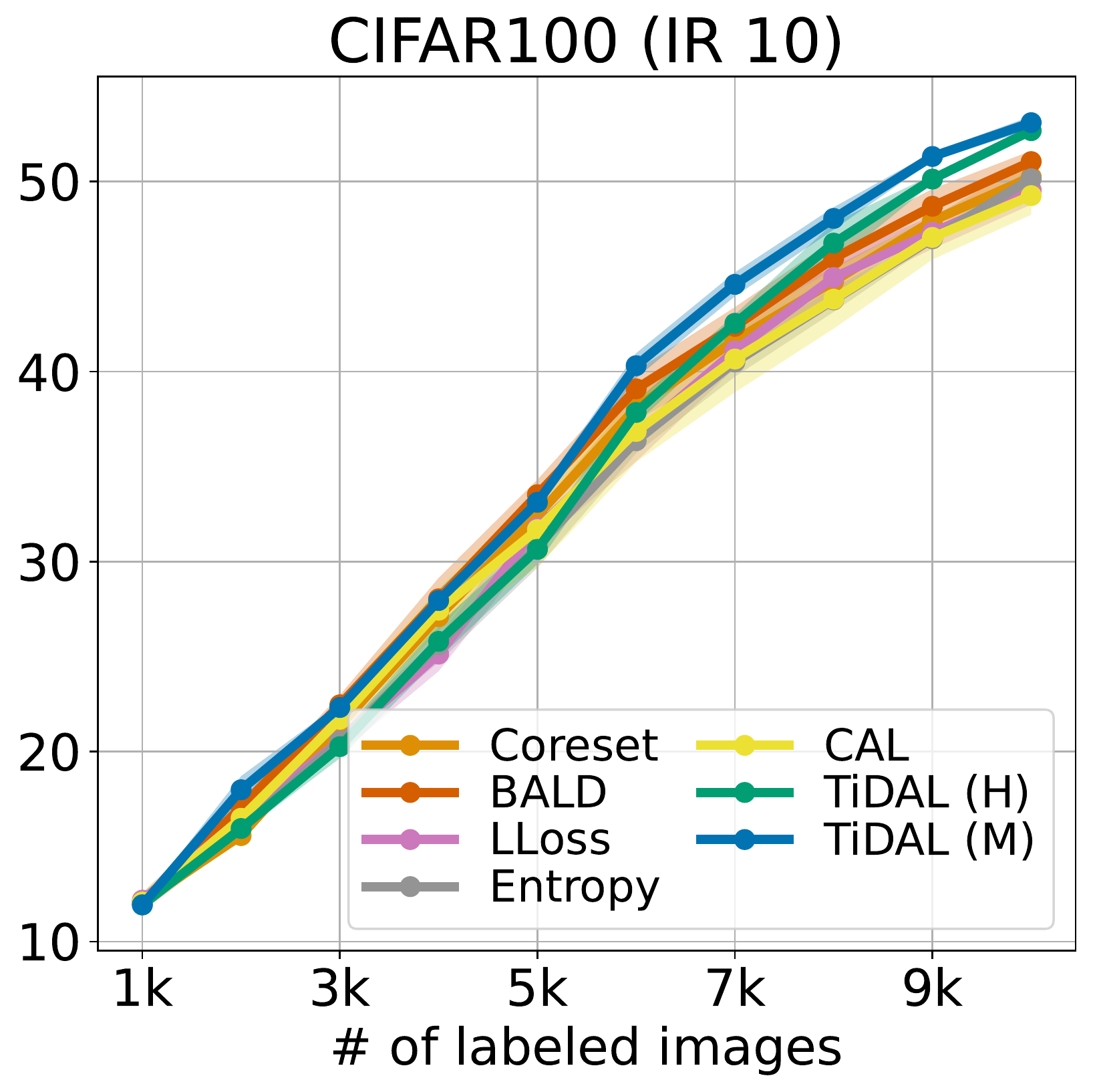}}
\subfloat{\includegraphics[width=0.33\columnwidth]{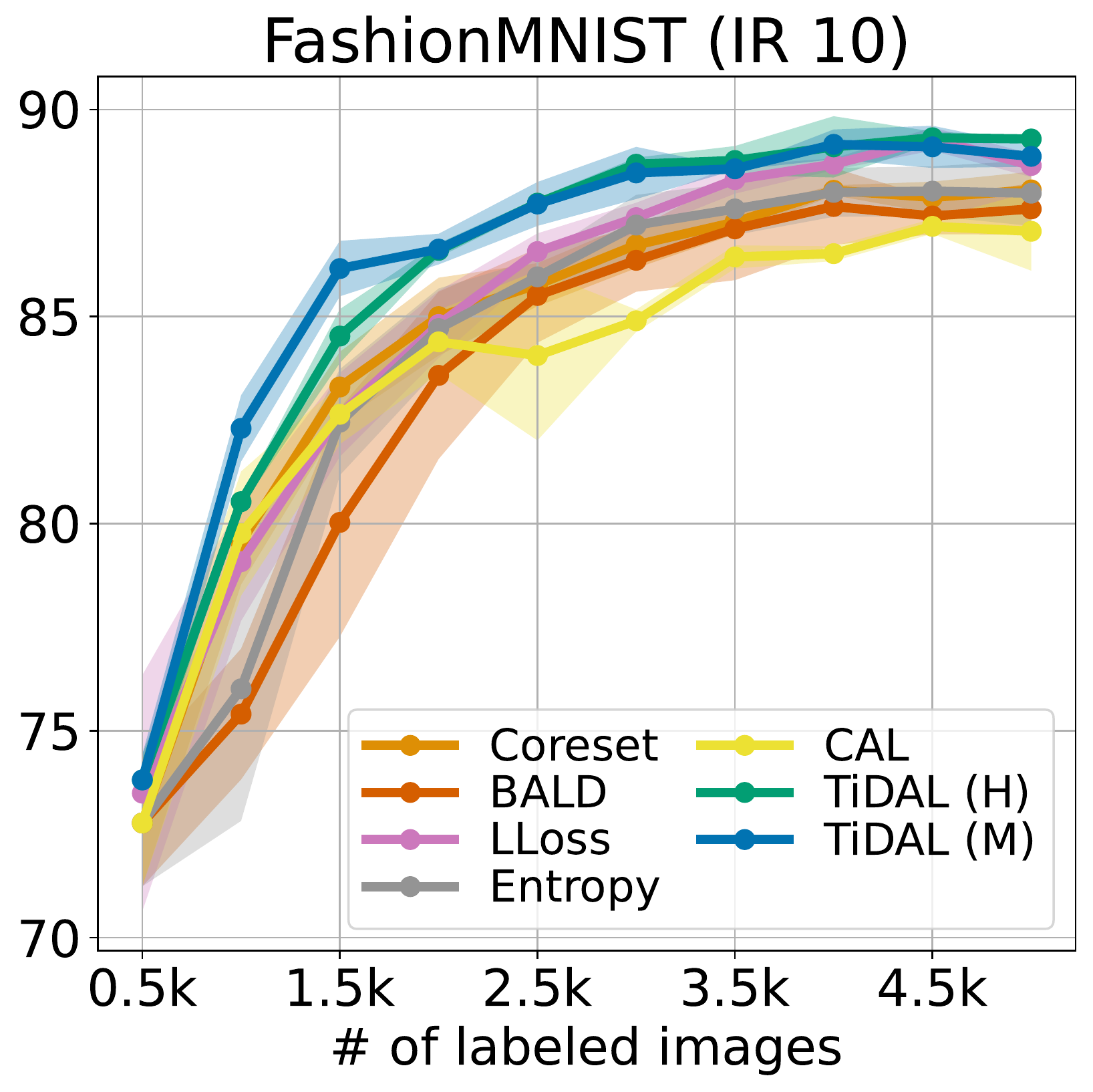}}\\
\subfloat{\includegraphics[width=0.33\columnwidth]{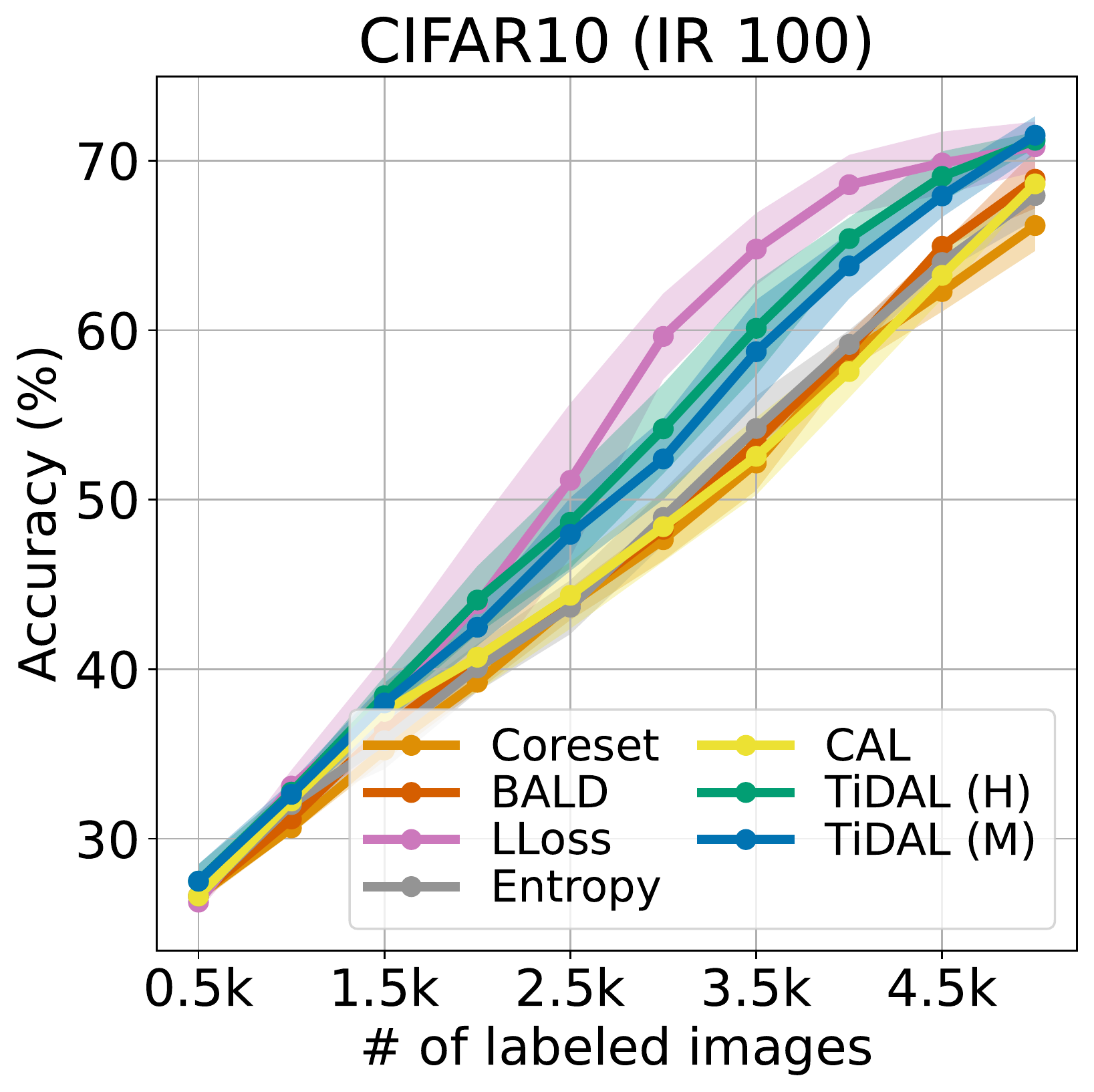}}
\subfloat{\includegraphics[width=0.33\columnwidth]{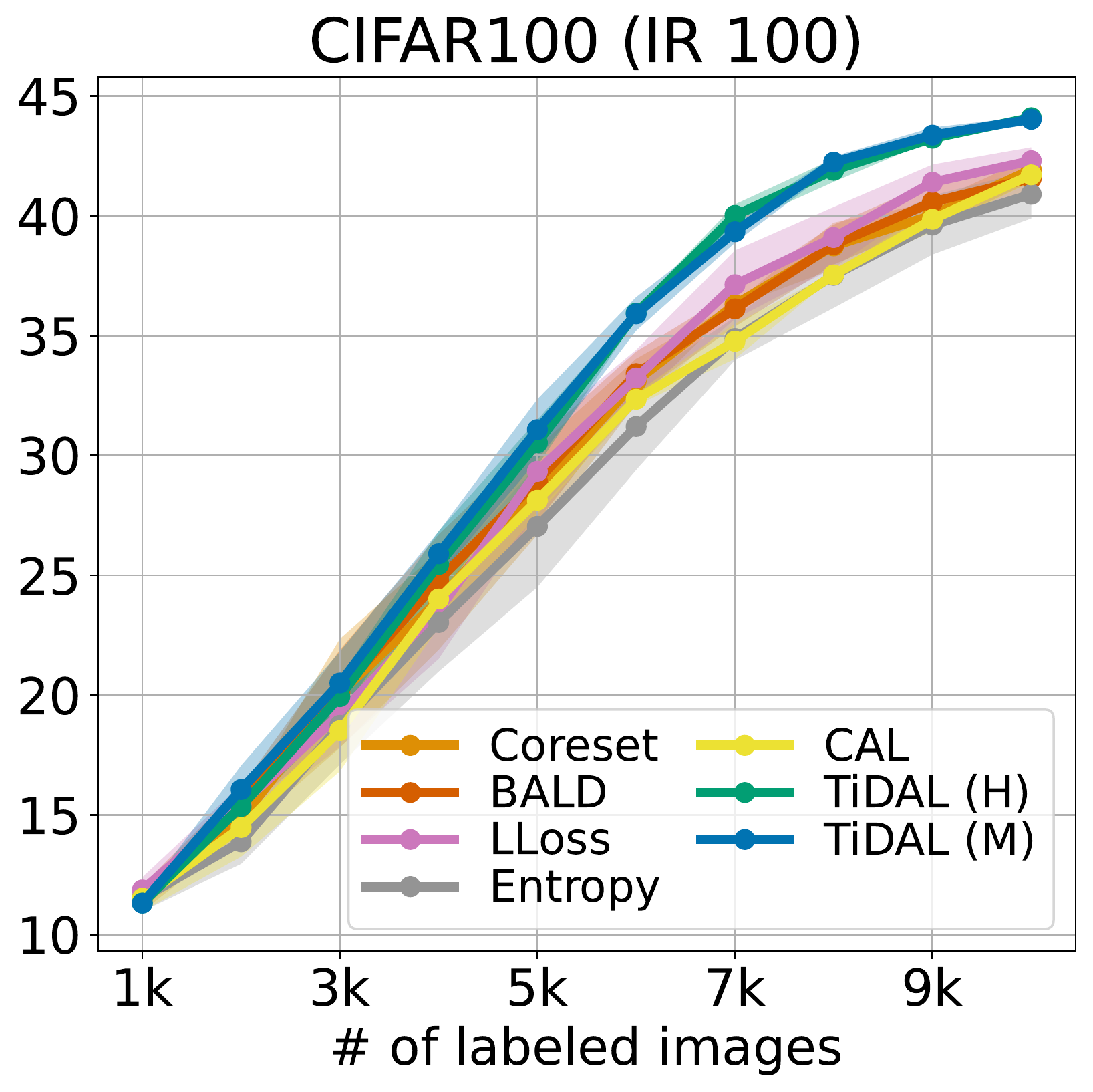}}
\subfloat{\includegraphics[width=0.33\columnwidth]{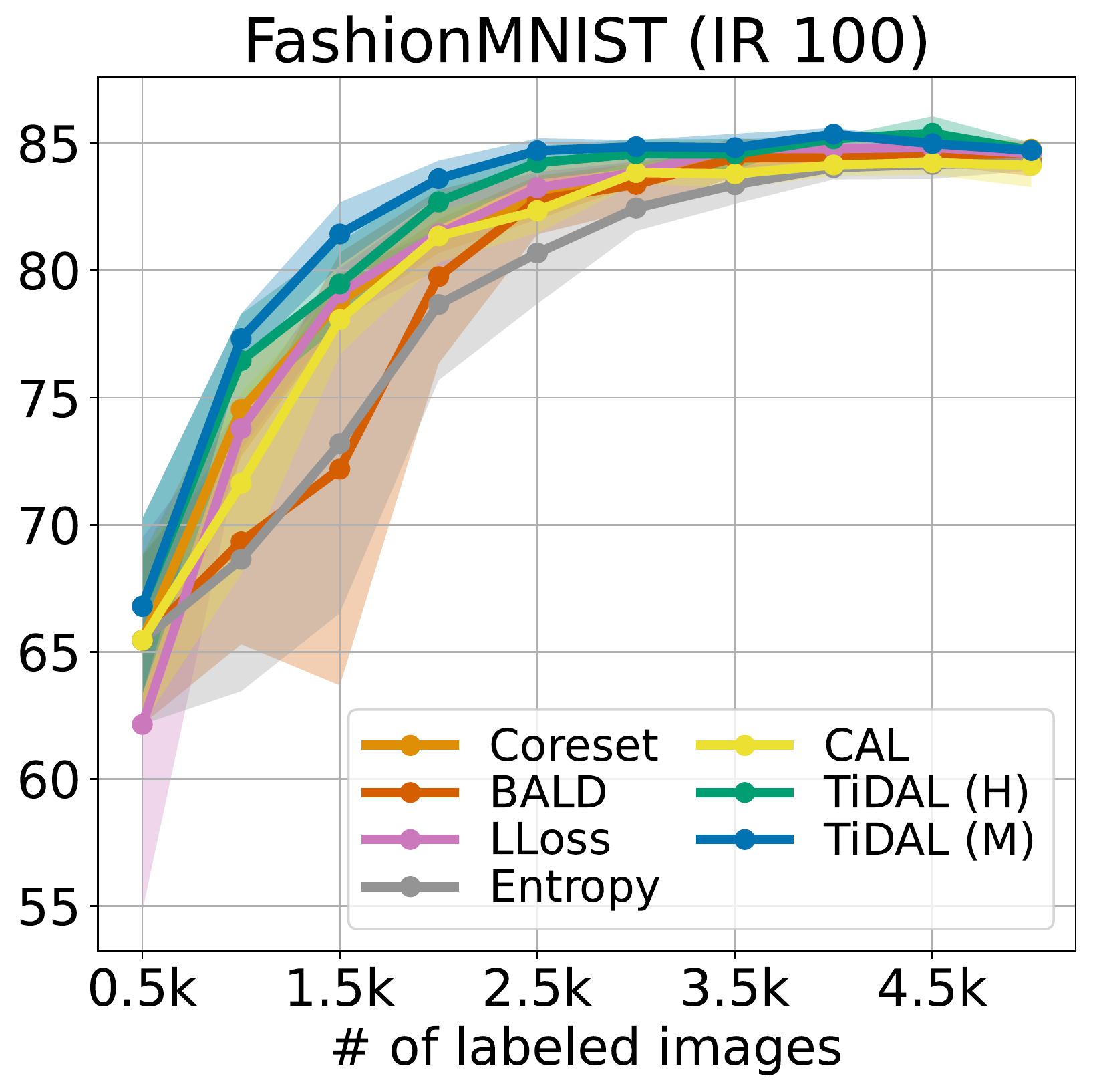}}
\caption{
    Averaged absolute accuracy improvement curves and their 95\% confidence interval (shaded) of AL methods over the number of labeled samples on synthetically imbalanced datasets.
    We use the imbalance ratio (IR) of 10 and 100 on CIFAR10, CIFAR100, and FashionMNIST.
}
\label{fig:14_appendix_absolute_imbalanced_synth_sota}
\end{figure}

%% file: Figures/11_VAAL_SOTA.tex
\begin{figure}[t] 
\centering
\subfloat{\includegraphics[width=0.33\columnwidth]{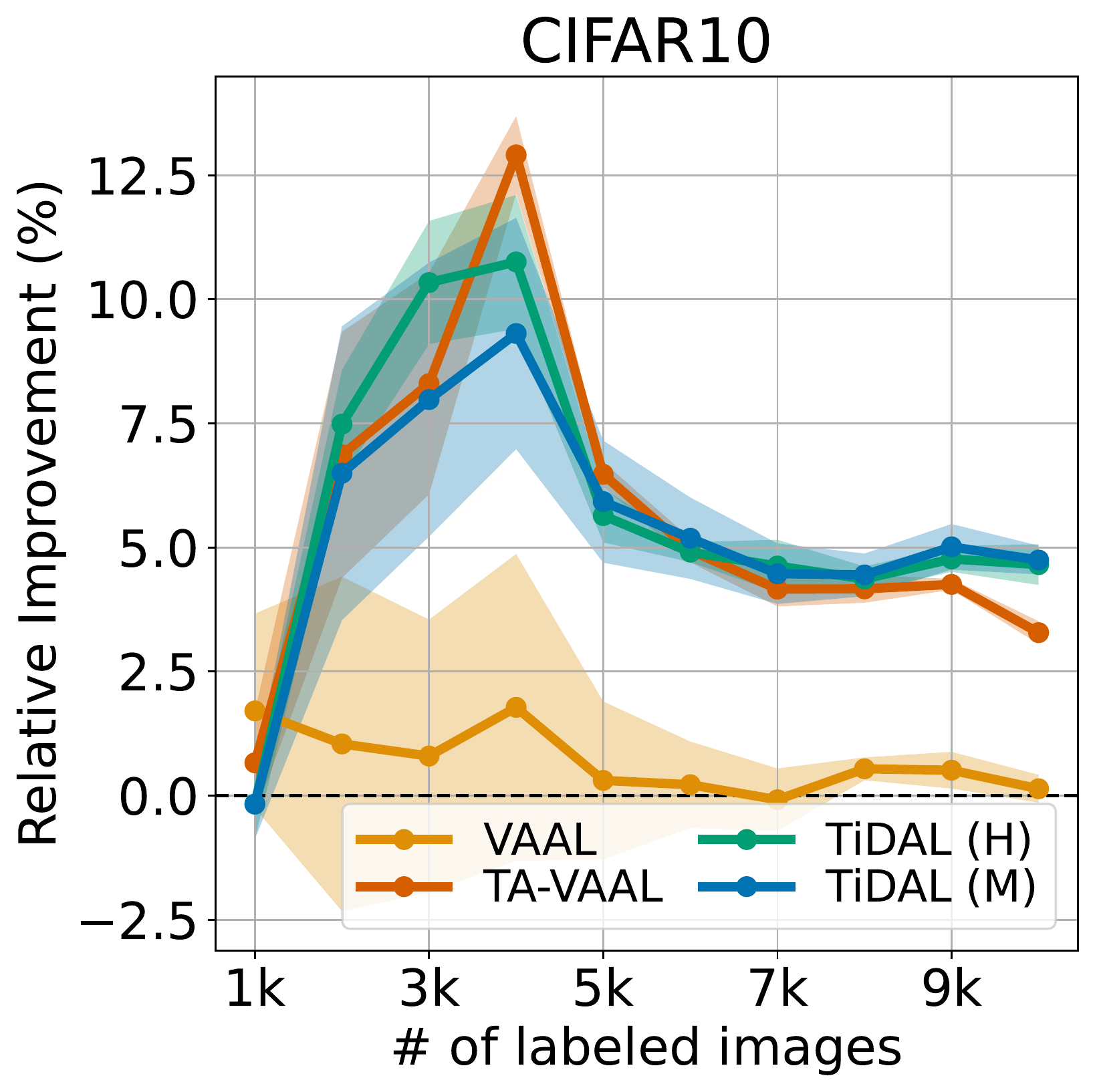}}
\subfloat{\includegraphics[width=0.33\columnwidth]{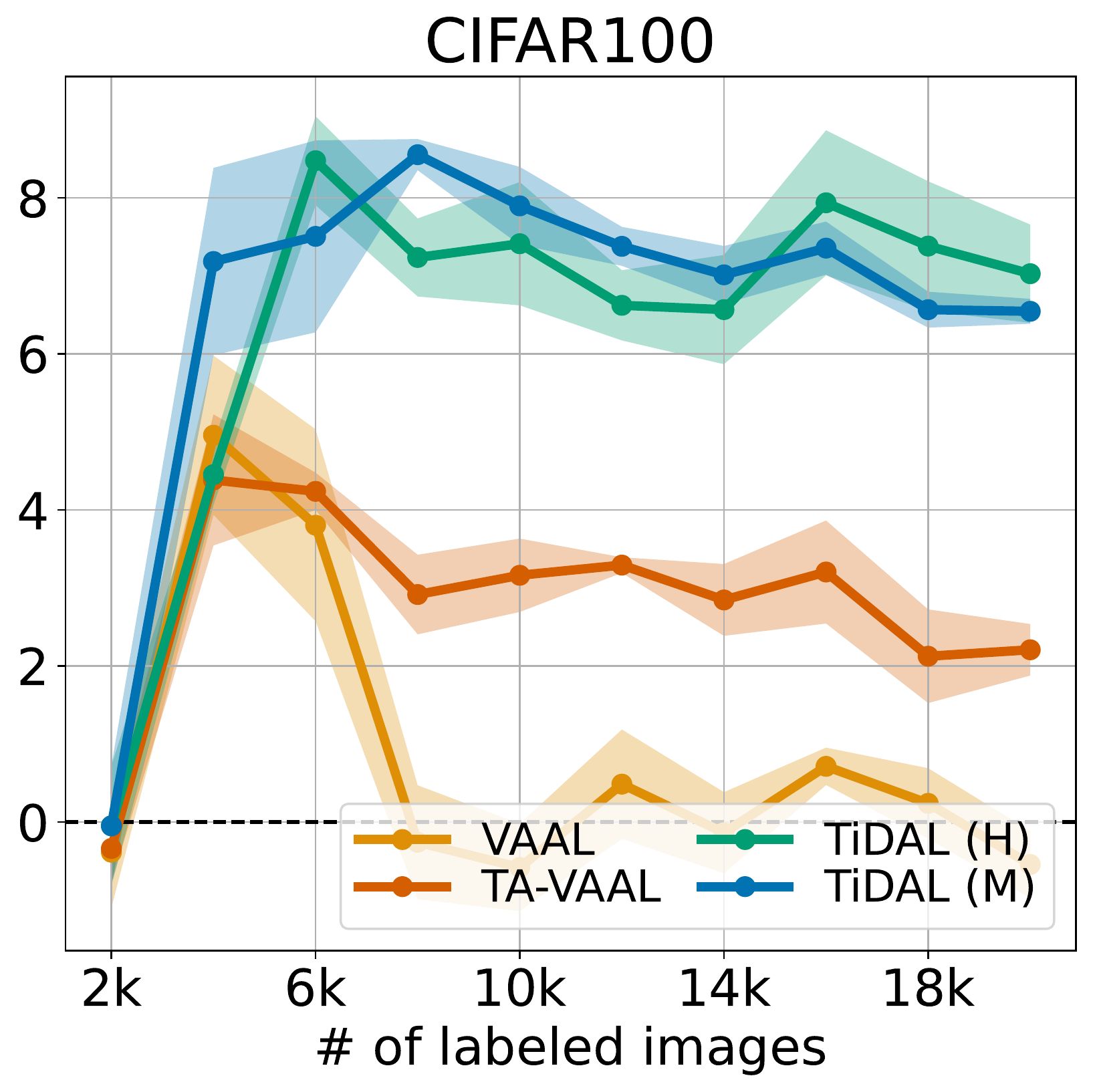}}
\subfloat{\includegraphics[width=0.33\columnwidth]{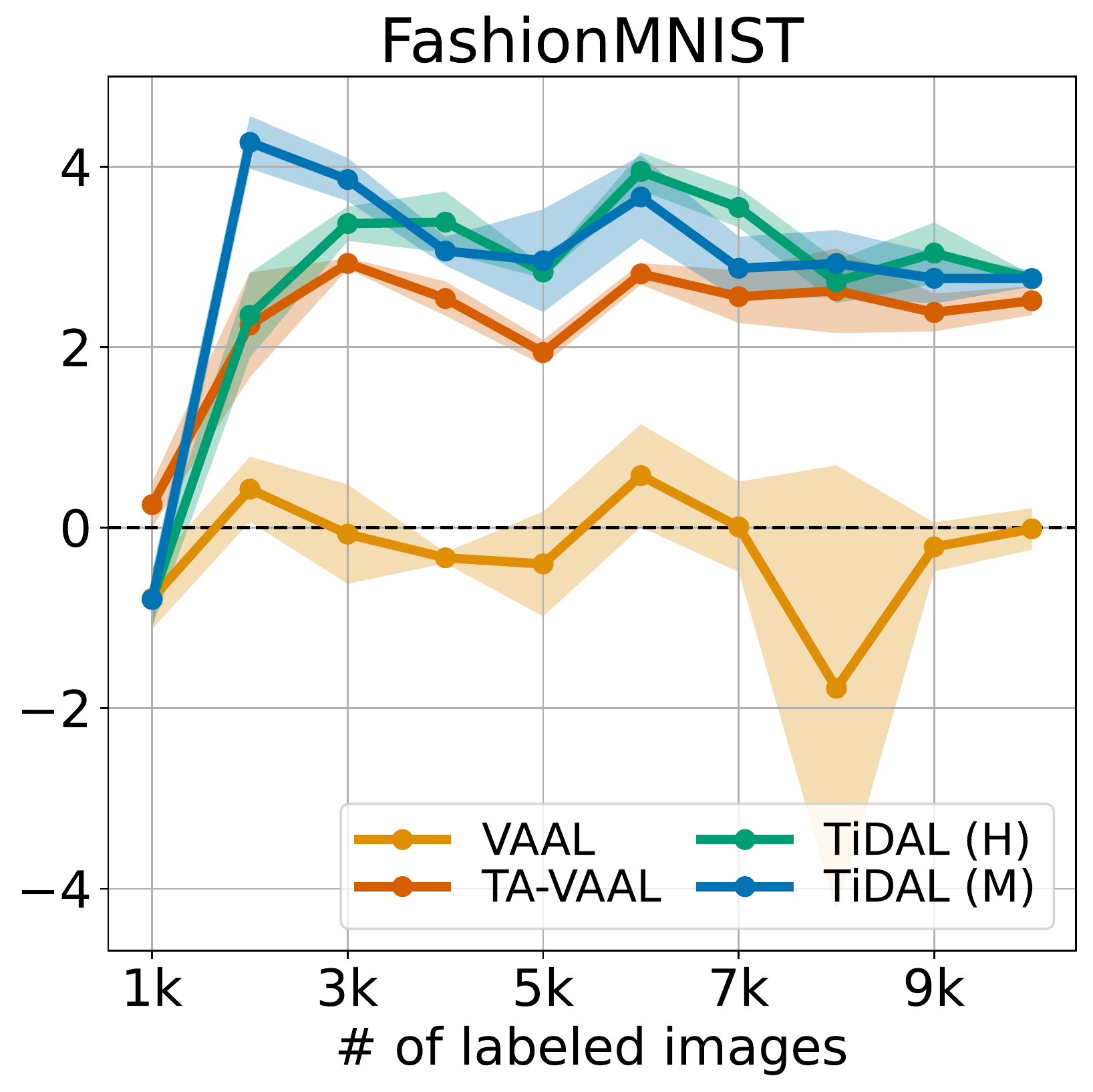}}\\
\subfloat{\includegraphics[width=0.33\columnwidth]{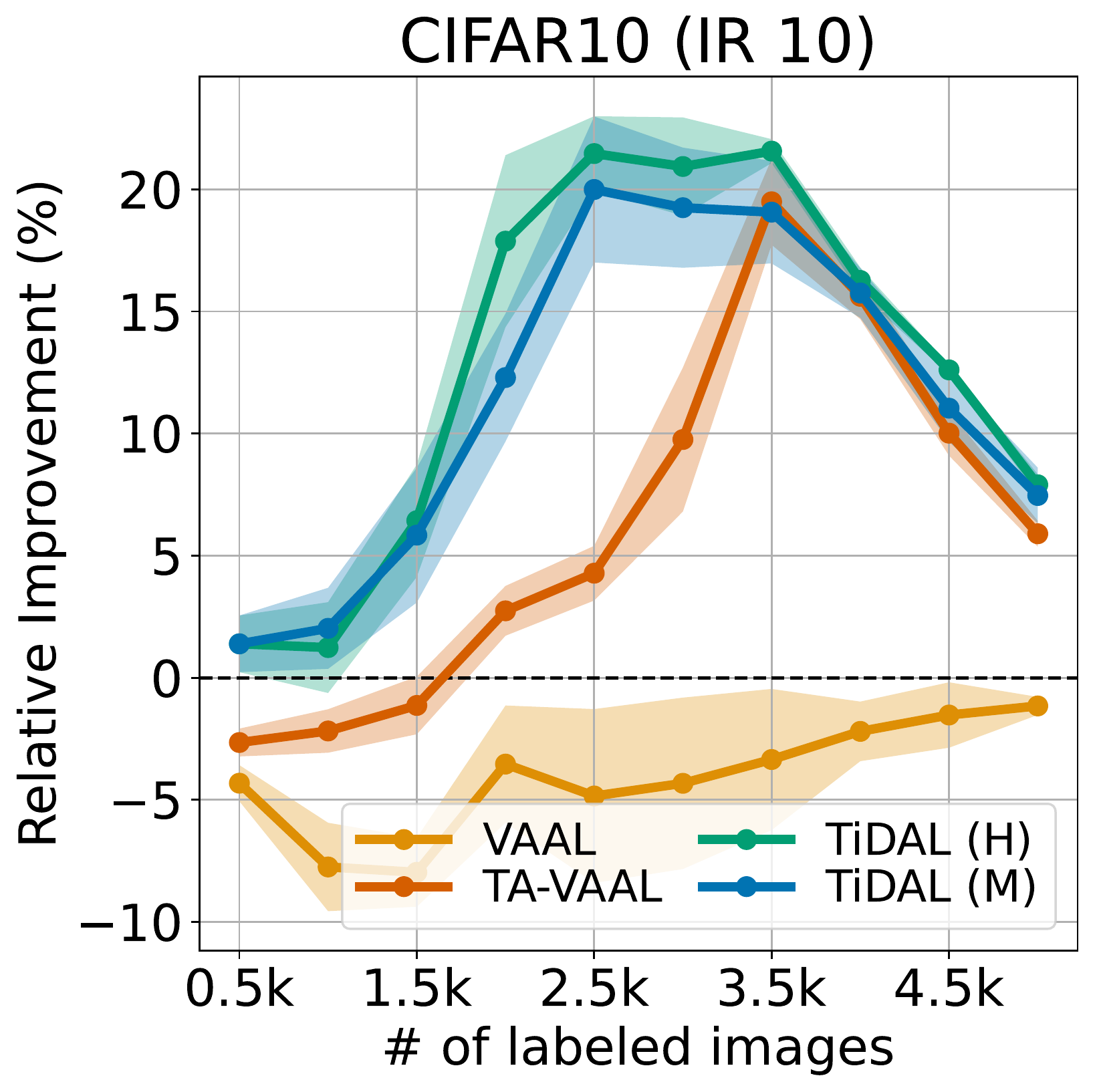}}
\subfloat{\includegraphics[width=0.33\columnwidth]{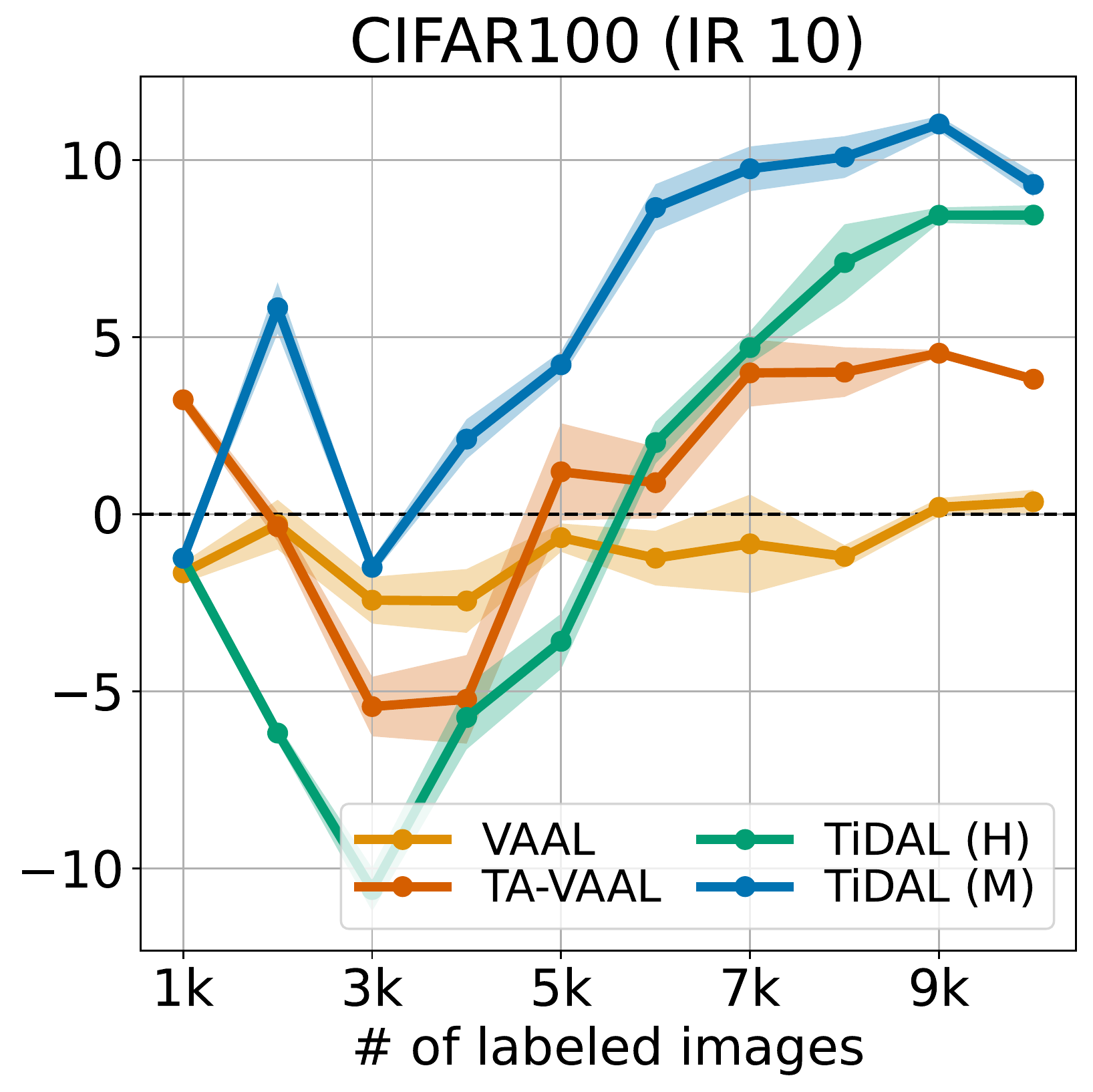}}
\subfloat{\includegraphics[width=0.33\columnwidth]{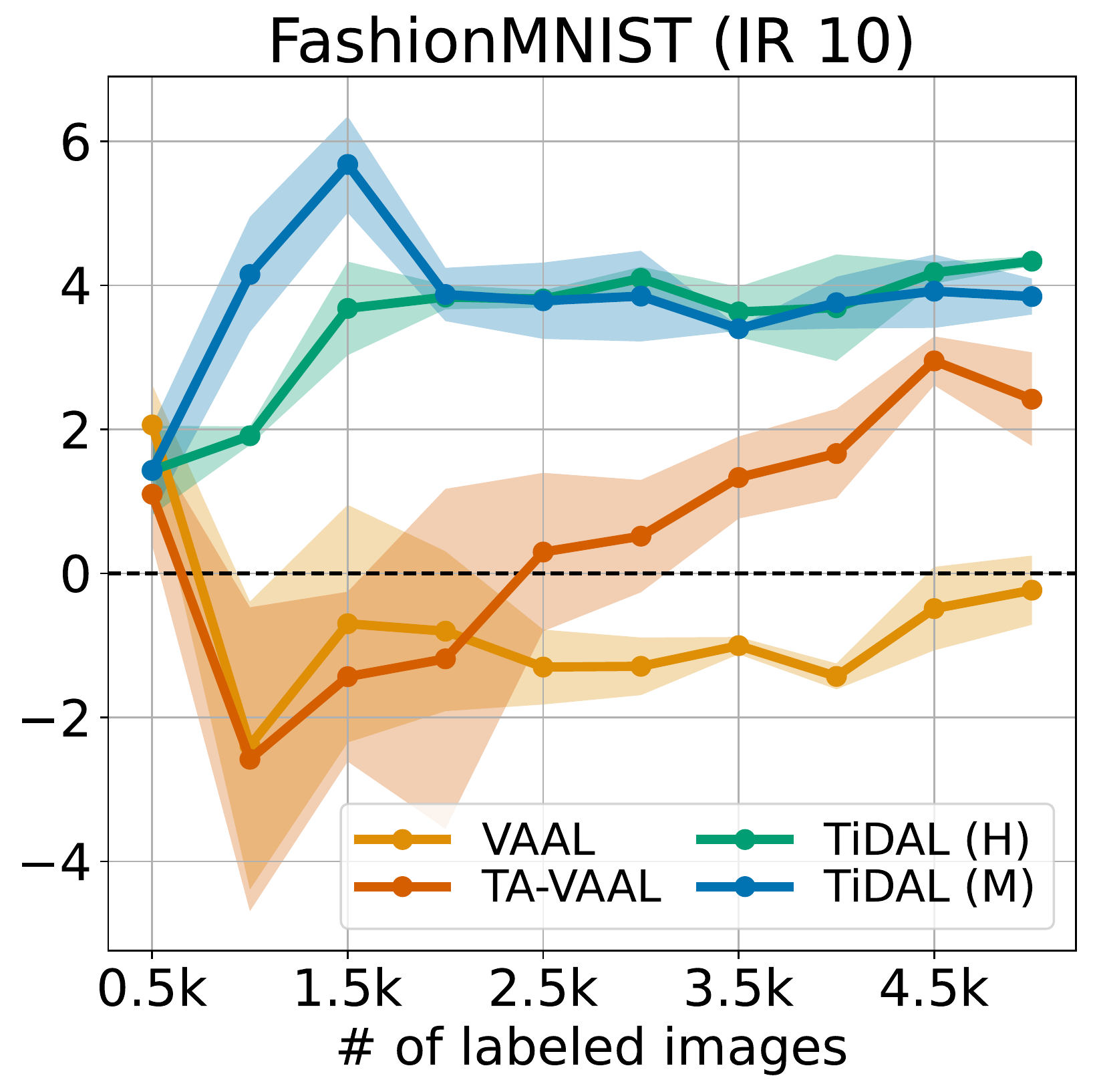}}\\
\subfloat{\includegraphics[width=0.33\columnwidth]{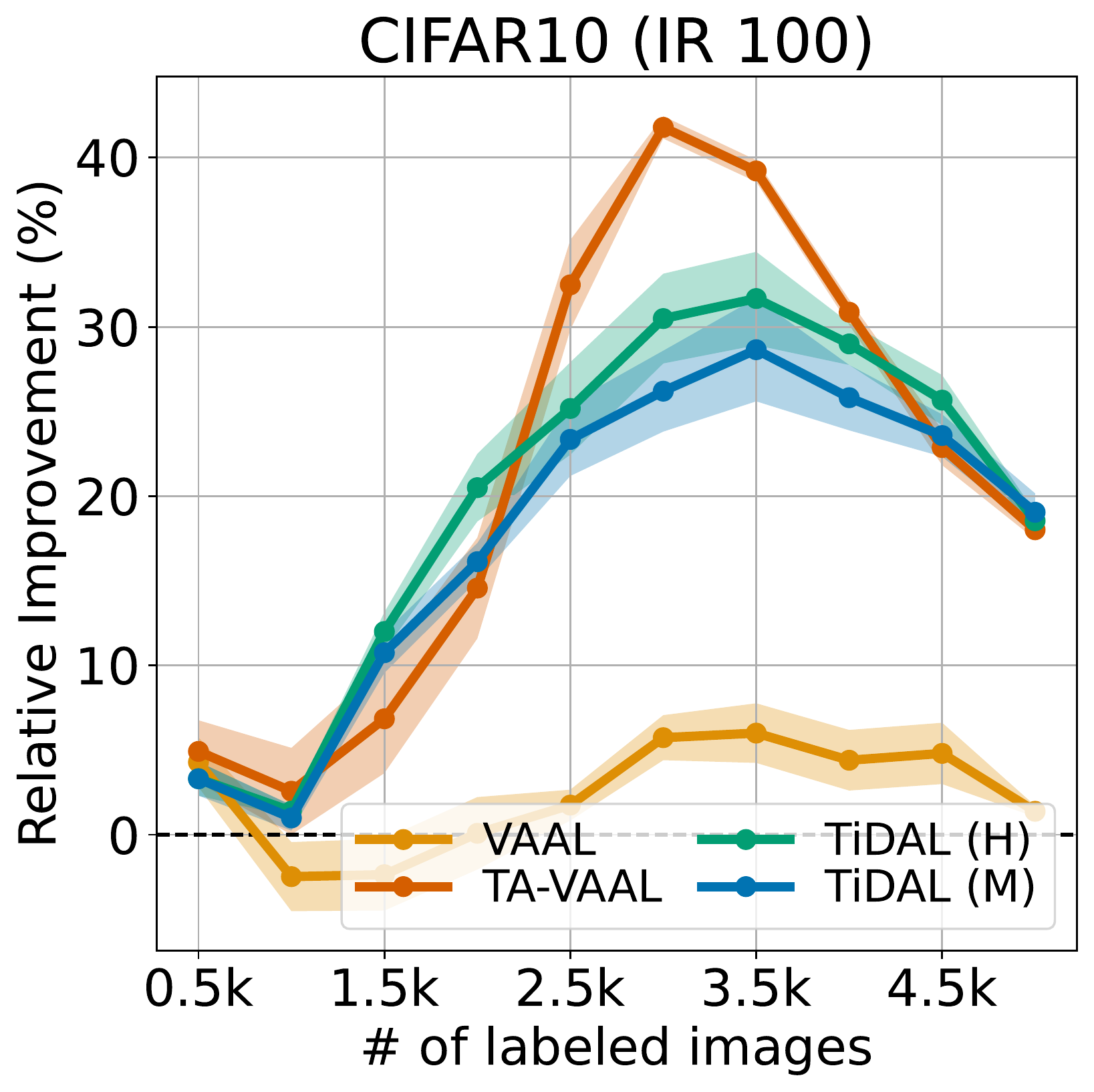}}
\subfloat{\includegraphics[width=0.33\columnwidth]{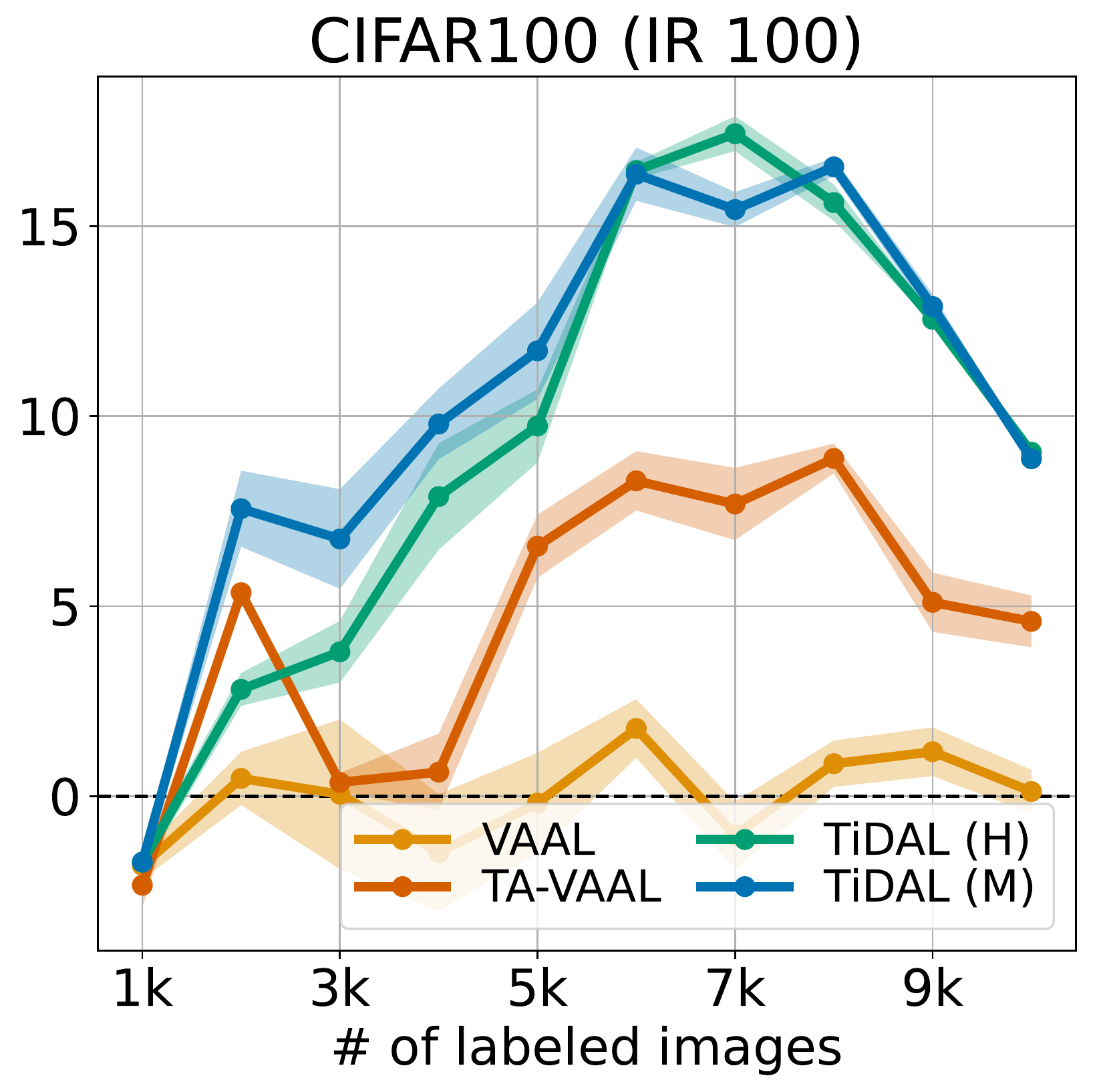}}
\subfloat{\includegraphics[width=0.33\columnwidth]{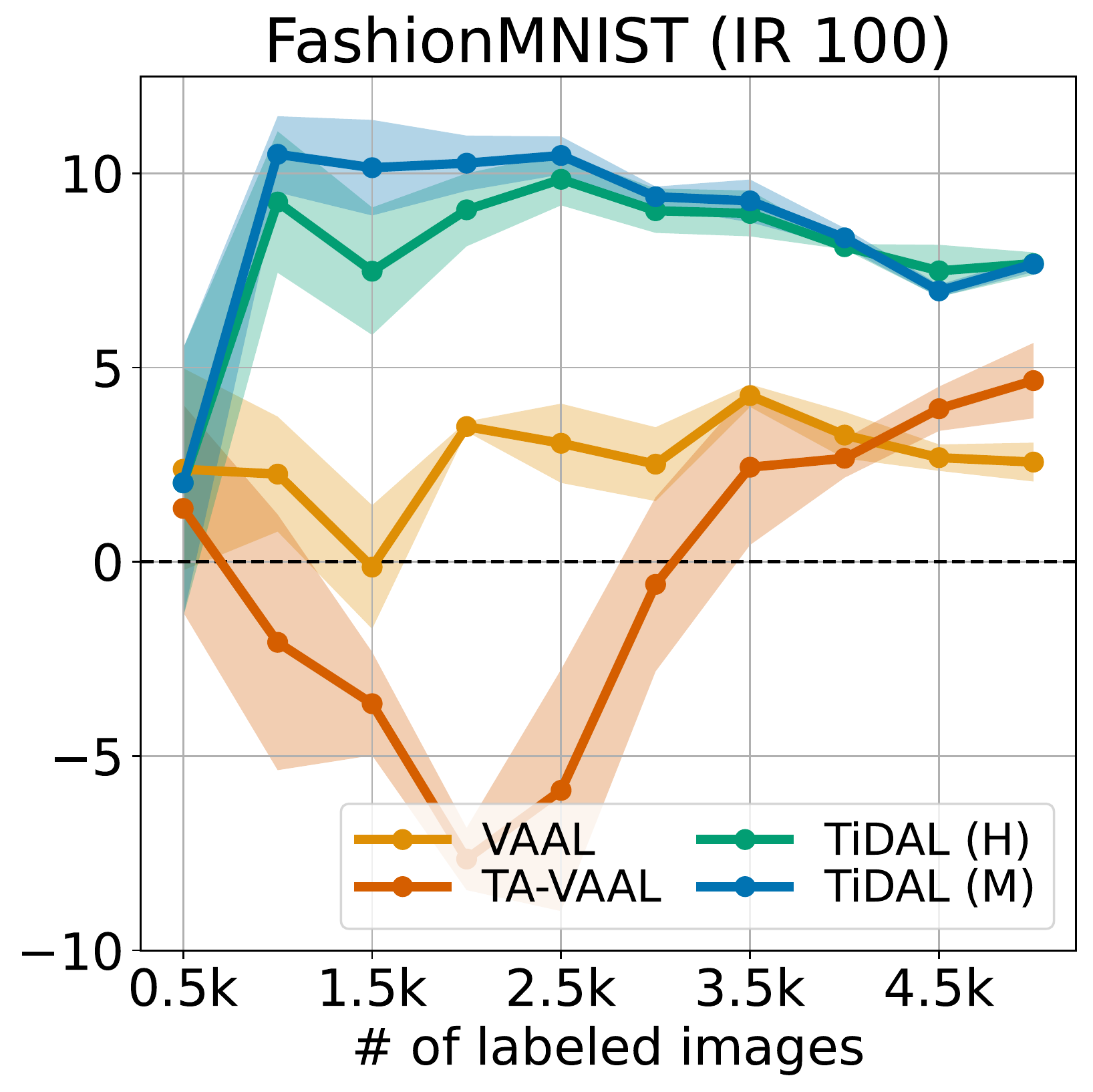}}
\caption{
    Averaged relative accuracy improvement curves and their 95\% confidence interval (shaded) of AL methods over the number of labeled samples on balanced and synthetically imbalanced datasets.
    We use the imbalance ratio (IR) of 10 and 100 on CIFAR10, CIFAR100, and FashionMNIST to synthetically imbalance the dataset.
}
\label{fig:11_vaal_sota}
\end{figure}

%% file: Figures/12_Margin_Variants.tex
\begin{figure}[t] 
\centering
\subfloat{\includegraphics[width=0.33\columnwidth]{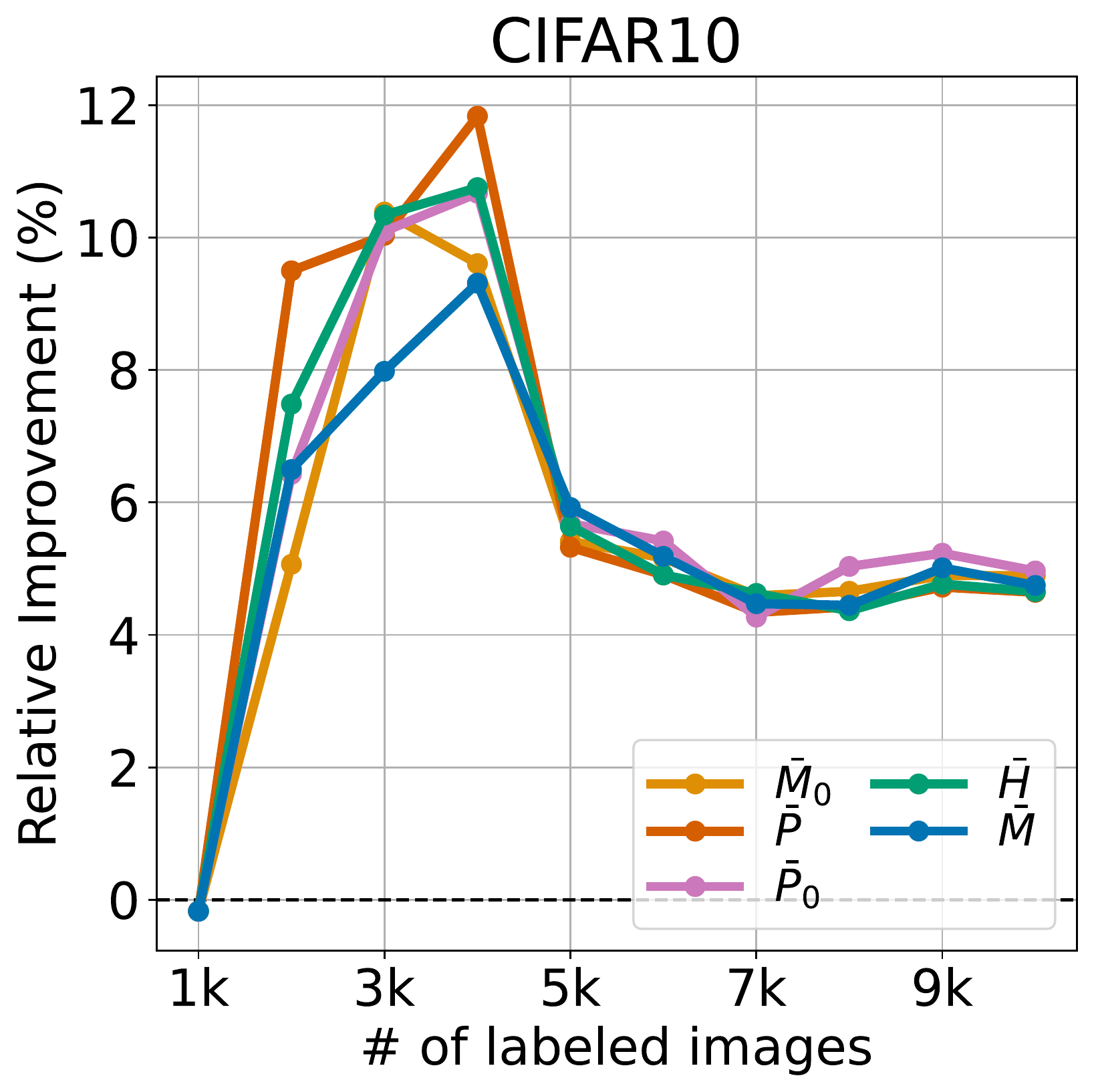}}
\subfloat{\includegraphics[width=0.33\columnwidth]{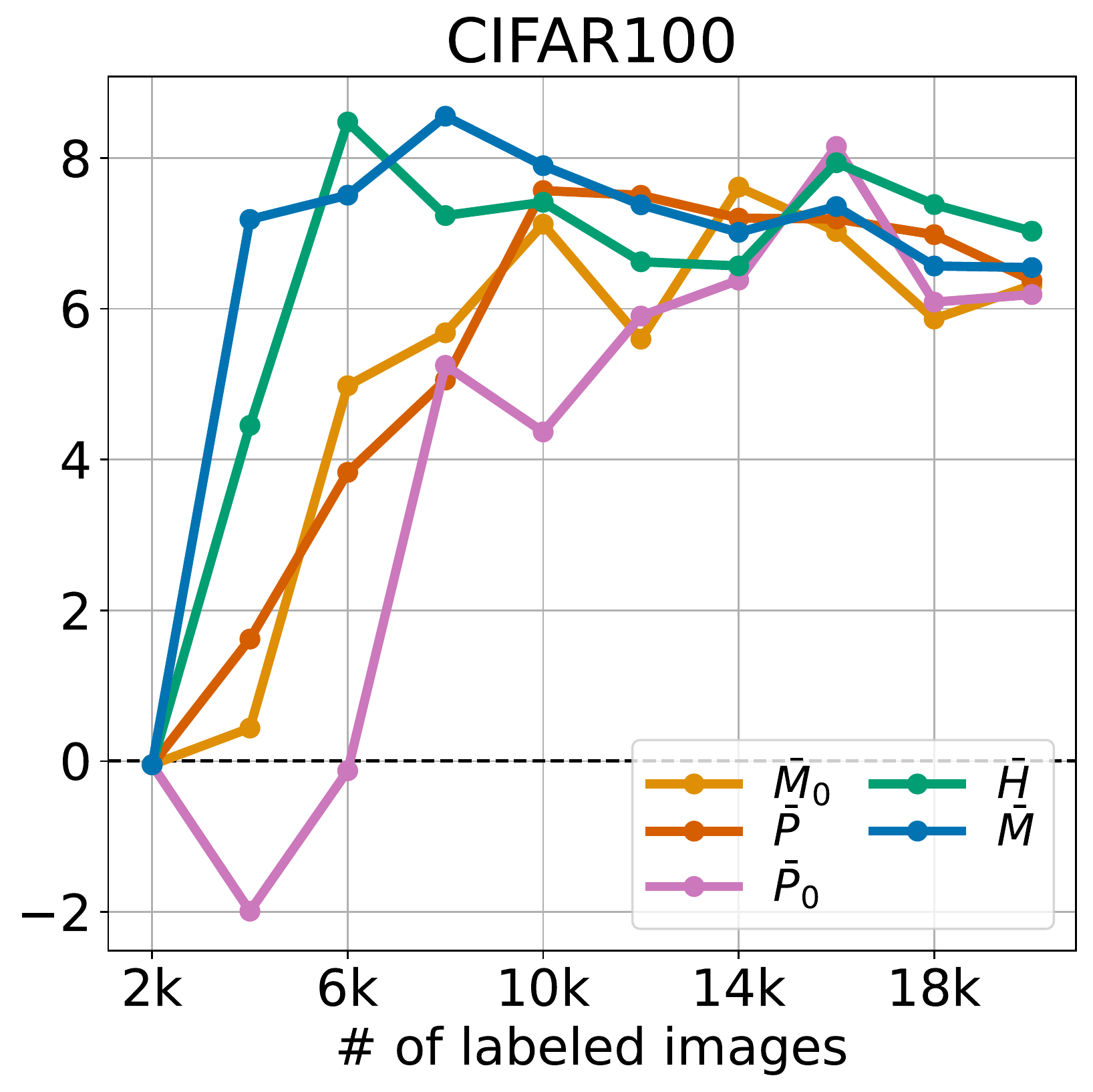}}
\subfloat{\includegraphics[width=0.33\columnwidth]{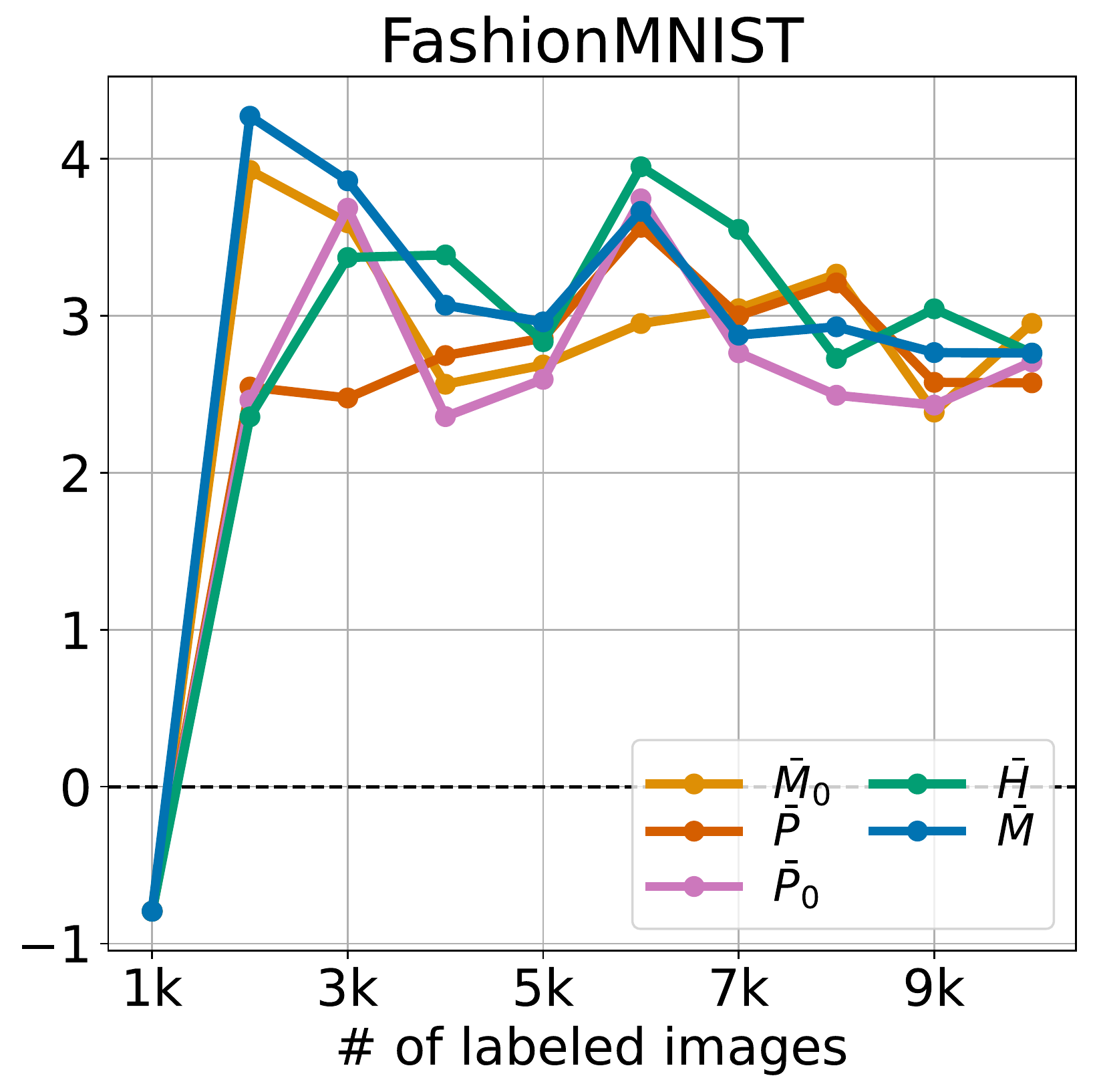}}\\
\subfloat{\includegraphics[width=0.33\columnwidth]{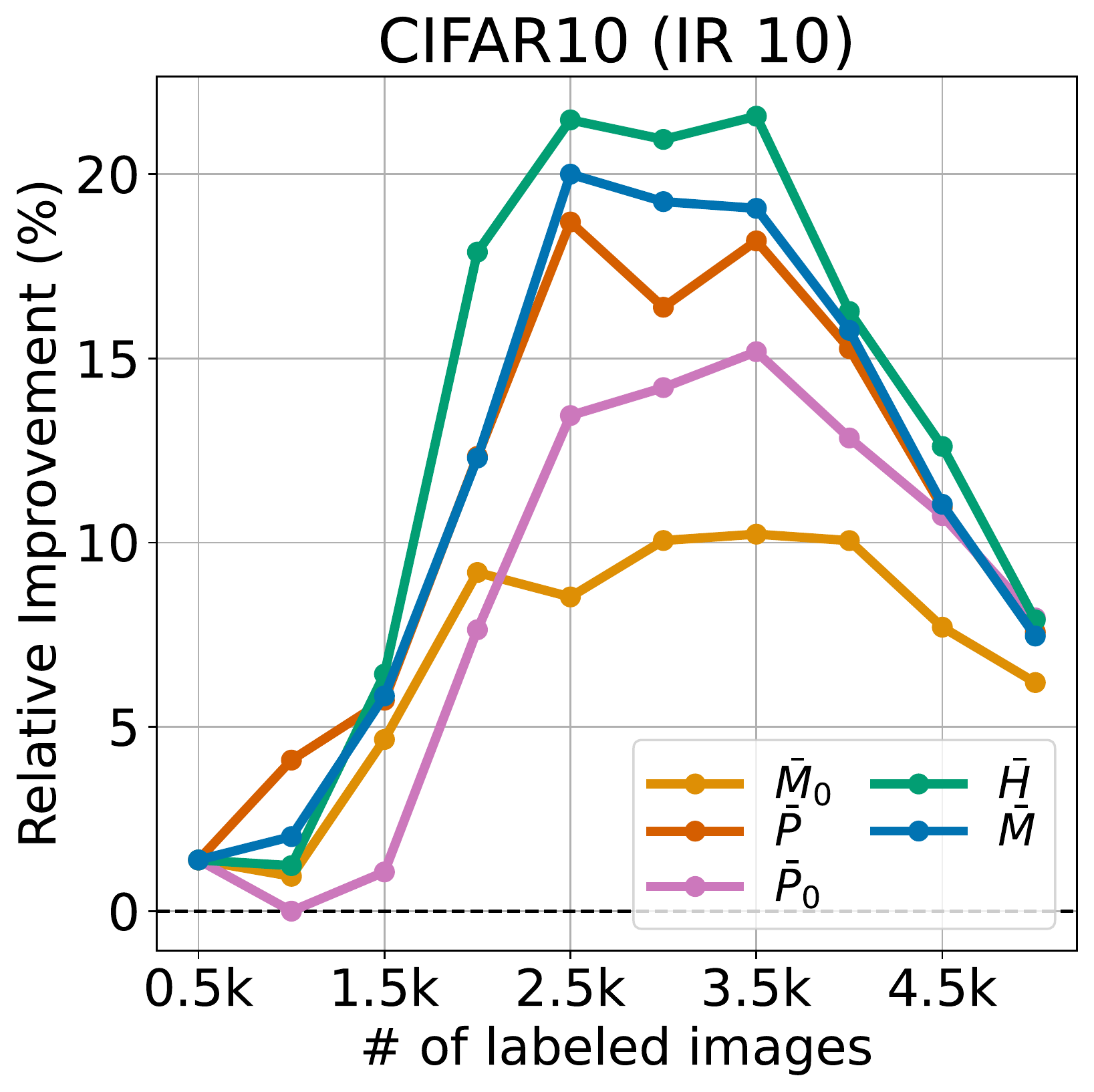}}
\subfloat{\includegraphics[width=0.33\columnwidth]{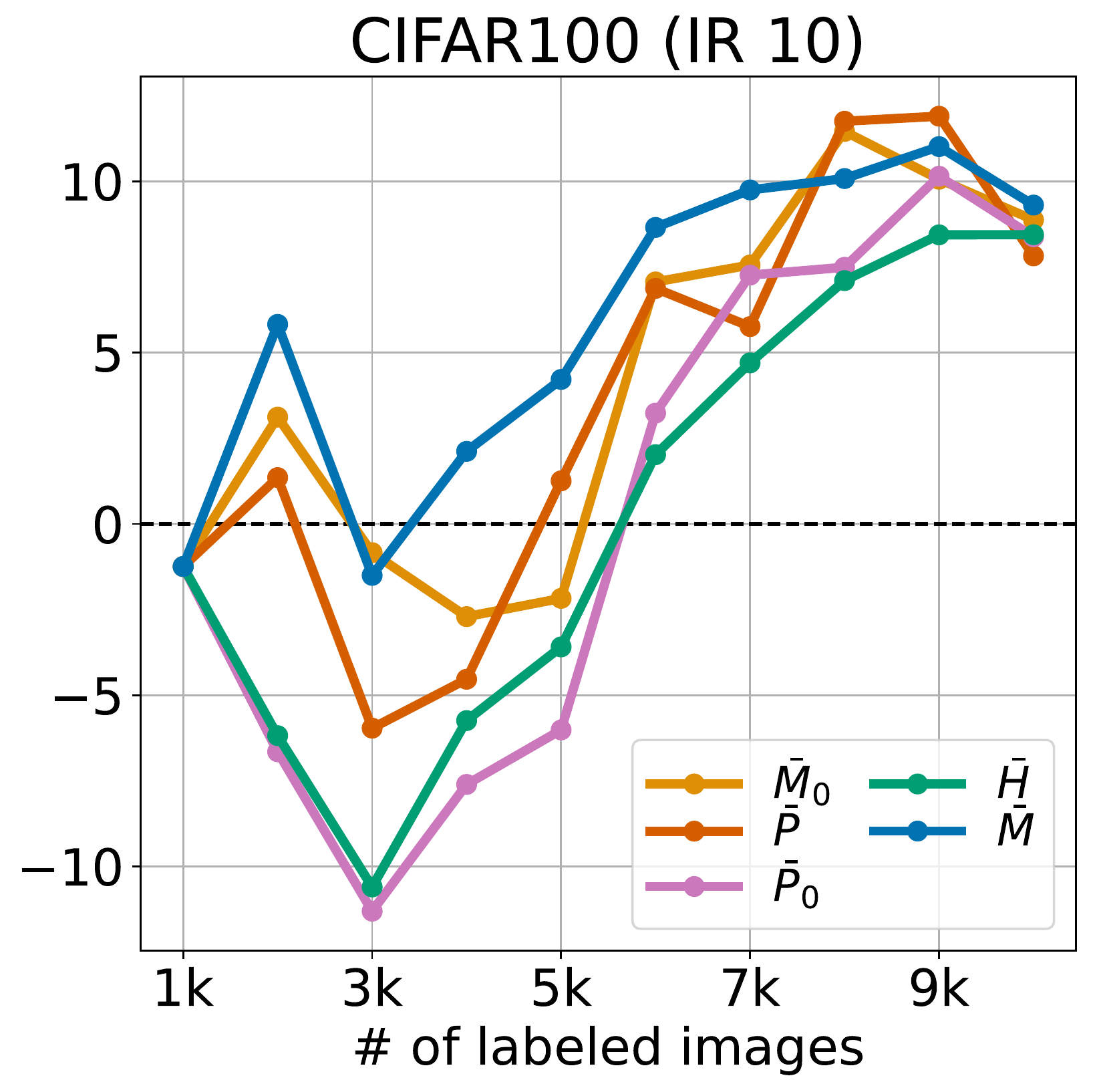}}
\subfloat{\includegraphics[width=0.33\columnwidth]{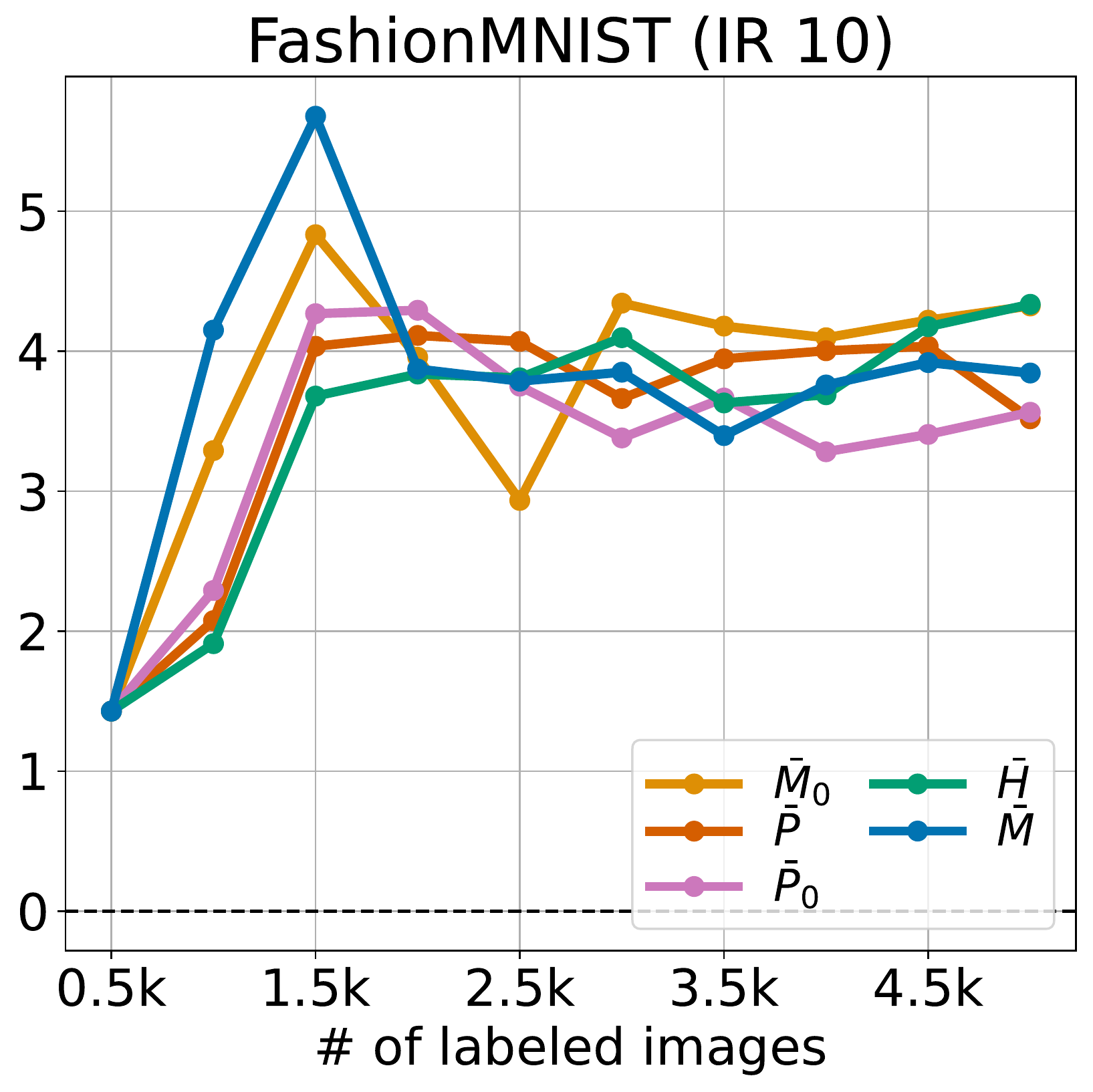}}\\
\subfloat{\includegraphics[width=0.33\columnwidth]{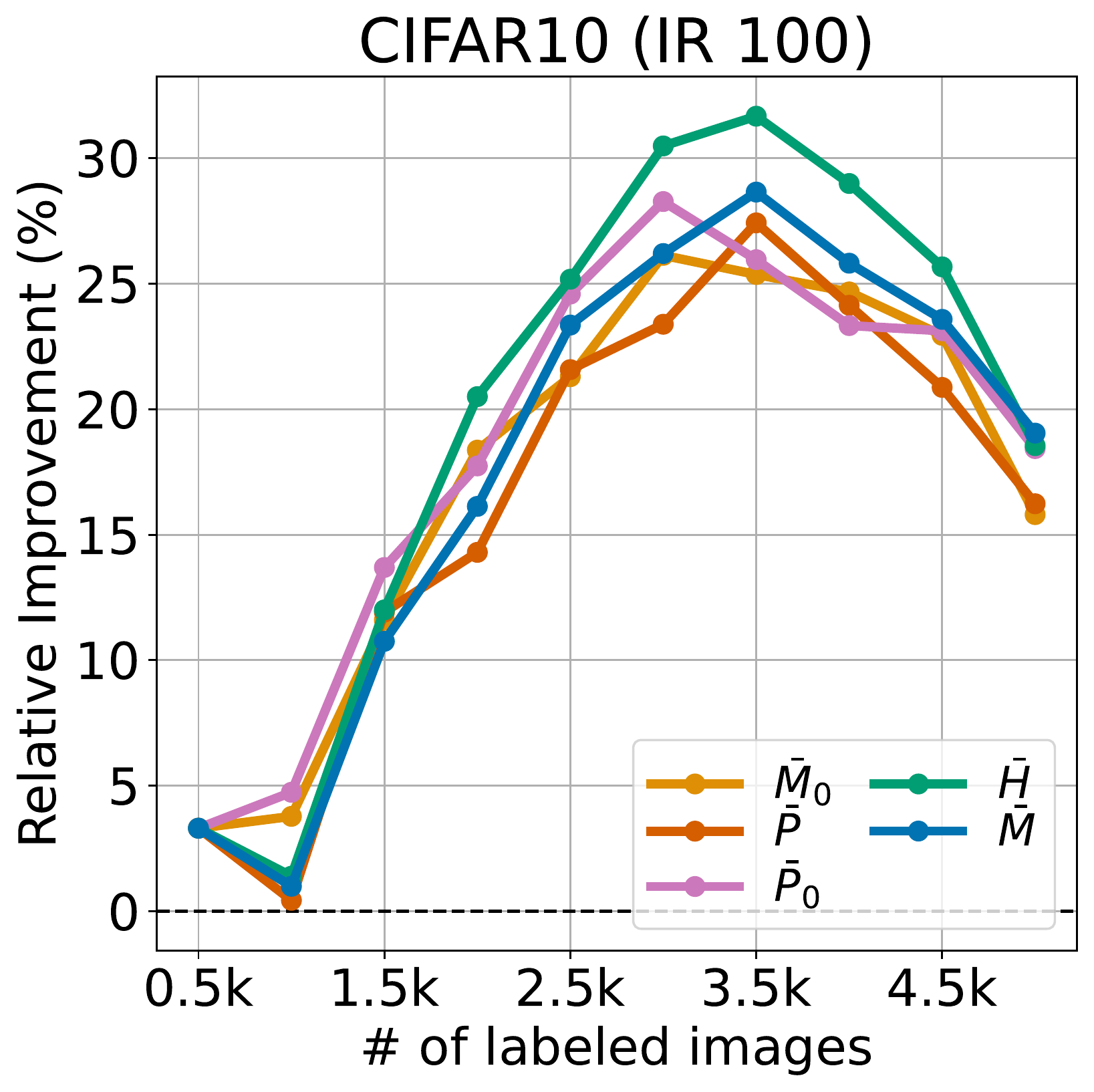}}
\subfloat{\includegraphics[width=0.33\columnwidth]{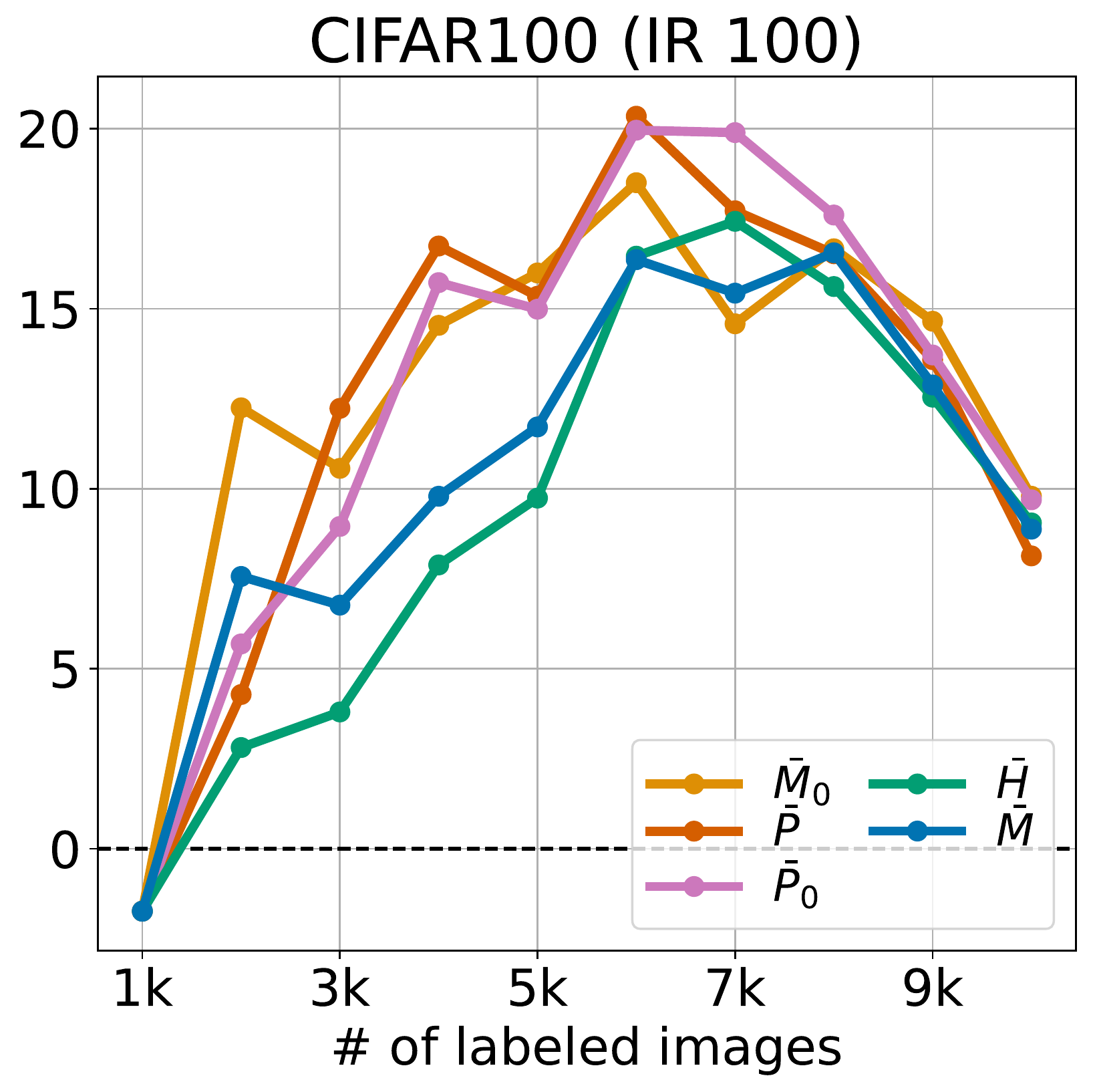}}
\subfloat{\includegraphics[width=0.33\columnwidth]{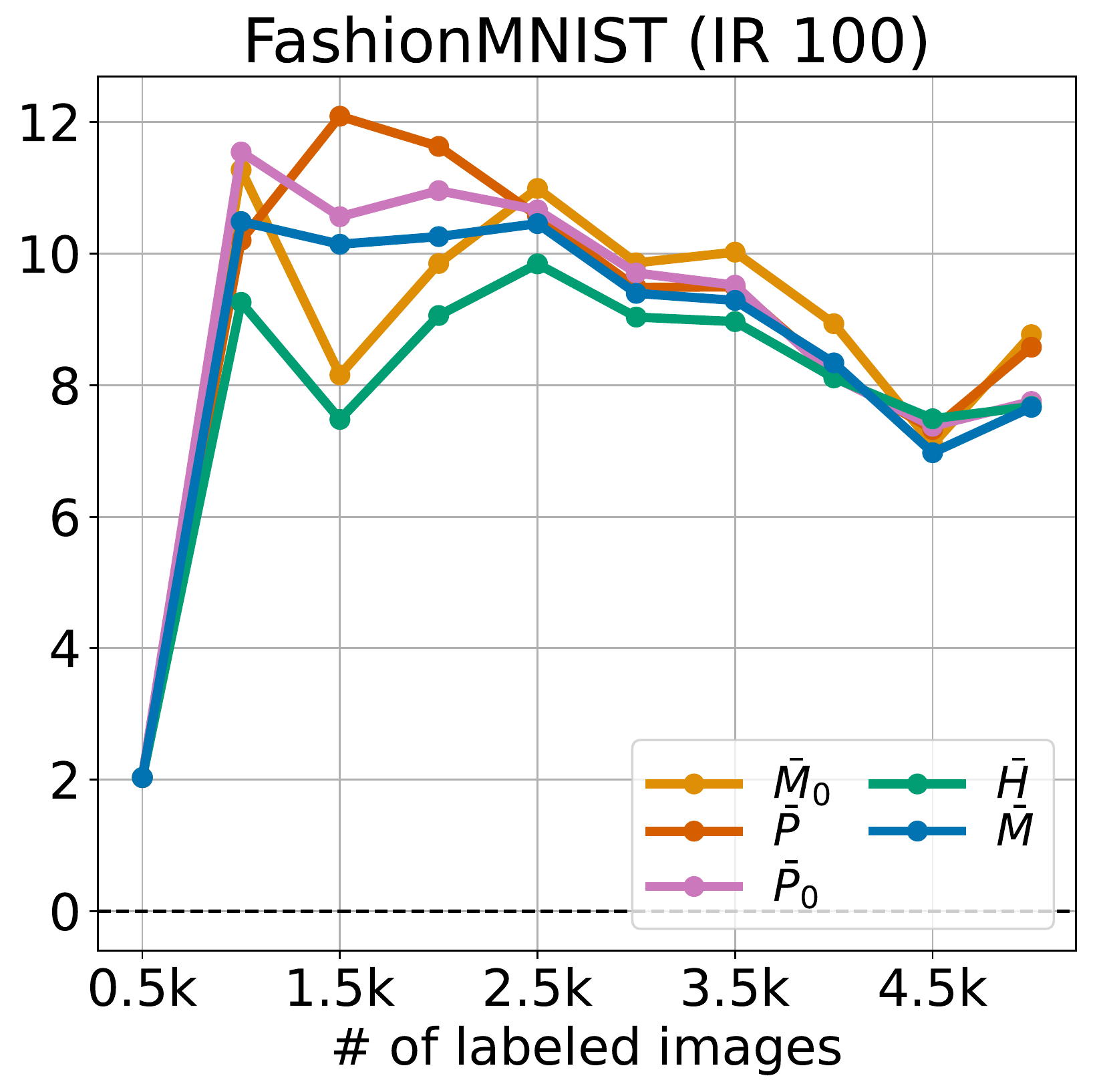}}
\caption{
    Averaged relative accuracy improvement curves of different uncertainty estimation strategies over the number of labeled samples on balanced and synthetically imbalanced datasets.
    We use the imbalance ratio (IR) of 10 and 100 on CIFAR10, CIFAR100, and FashionMNIST to synthetically imbalance the dataset.
}
\label{fig:12_margin_sota}
\end{figure}

%% file: egpaper_for_review.bbl
\begin{thebibliography}{10}\itemsep=-1pt

\bibitem{angluin1988queries}
Dana Angluin.
\newblock Queries and concept learning.
\newblock {\em Machine learning}, 2(4):319--342, 1988.

\bibitem{arazo2019unsupervised}
Eric Arazo, Diego Ortego, Paul Albert, Noel O’Connor, and Kevin McGuinness.
\newblock Unsupervised label noise modeling and loss correction.
\newblock In {\em International conference on machine learning}, pages
  312--321. PMLR, 2019.

\bibitem{arora2018convergence}
Sanjeev Arora, Nadav Cohen, Noah Golowich, and Wei Hu.
\newblock A convergence analysis of gradient descent for deep linear neural
  networks.
\newblock In {\em International Conference on Learning Representations}, 2018.

\bibitem{ash2019deep}
Jordan~T Ash, Chicheng Zhang, Akshay Krishnamurthy, John Langford, and Alekh
  Agarwal.
\newblock Deep batch active learning by diverse, uncertain gradient lower
  bounds.
\newblock {\em arXiv preprint arXiv:1906.03671}, 2019.

\bibitem{atlas1990training}
Les~E Atlas, David~A Cohn, and Richard~E Ladner.
\newblock Training connectionist networks with queries and selective sampling.
\newblock In {\em Advances in neural information processing systems}, pages
  566--573. Citeseer, 1990.

\bibitem{beluch2018power}
William~H Beluch, Tim Genewein, Andreas N{\"u}rnberger, and Jan~M K{\"o}hler.
\newblock The power of ensembles for active learning in image classification.
\newblock In {\em Proceedings of the IEEE conference on computer vision and
  pattern recognition}, pages 9368--9377, 2018.

\bibitem{bengar2022class}
Javad~Zolfaghari Bengar, Joost van~de Weijer, Laura~Lopez Fuentes, and Bogdan
  Raducanu.
\newblock Class-balanced active learning for image classification.
\newblock In {\em Proceedings of the IEEE/CVF Winter Conference on Applications
  of Computer Vision}, pages 1536--1545, 2022.

\bibitem{cao2019learning}
Kaidi Cao, Colin Wei, Adrien Gaidon, Nikos Arechiga, and Tengyu Ma.
\newblock Learning imbalanced datasets with label-distribution-aware margin
  loss.
\newblock {\em Advances in Neural Information Processing Systems},
  32:1567--1578, 2019.

\bibitem{chang2017active}
Haw-Shiuan Chang, Erik Learned-Miller, and Andrew McCallum.
\newblock Active bias: Training more accurate neural networks by emphasizing
  high variance samples.
\newblock {\em Advances in Neural Information Processing Systems}, 30, 2017.

\bibitem{cohn1994improving}
David Cohn, Les Atlas, and Richard Ladner.
\newblock Improving generalization with active learning.
\newblock {\em Machine learning}, 15(2):201--221, 1994.

\bibitem{corbiere2019addressing}
Charles Corbi{\`e}re, Nicolas Thome, Avner Bar-Hen, Matthieu Cord, and Patrick
  P{\'e}rez.
\newblock Addressing failure prediction by learning model confidence.
\newblock {\em Advances in Neural Information Processing Systems}, 32, 2019.

\bibitem{deng2009imagenet}
Jia Deng, Wei Dong, Richard Socher, Li-Jia Li, Kai Li, and Li Fei-Fei.
\newblock Imagenet: A large-scale hierarchical image database.
\newblock In {\em 2009 IEEE conference on computer vision and pattern
  recognition}, pages 248--255. Ieee, 2009.

\bibitem{gal2017deep}
Yarin Gal, Riashat Islam, and Zoubin Ghahramani.
\newblock Deep bayesian active learning with image data.
\newblock In {\em International Conference on Machine Learning}, pages
  1183--1192. PMLR, 2017.

\bibitem{gissin2019discriminative}
Daniel Gissin and Shai Shalev-Shwartz.
\newblock Discriminative active learning.
\newblock {\em arXiv preprint arXiv:1907.06347}, 2019.

\bibitem{gong2022meta}
Jia Gong, Zhipeng Fan, Qiuhong Ke, Hossein Rahmani, and Jun Liu.
\newblock Meta agent teaming active learning for pose estimation.
\newblock In {\em Proceedings of the IEEE/CVF Conference on Computer Vision and
  Pattern Recognition}, pages 11079--11089, 2022.

\bibitem{guo2017calibration}
Chuan Guo, Geoff Pleiss, Yu Sun, and Kilian~Q Weinberger.
\newblock On calibration of modern neural networks.
\newblock In {\em International conference on machine learning}, pages
  1321--1330. PMLR, 2017.

\bibitem{he2019local}
Hangfeng He and Weijie Su.
\newblock The local elasticity of neural networks.
\newblock In {\em International Conference on Learning Representations}, 2019.

\bibitem{he2016deep}
Kaiming He, Xiangyu Zhang, Shaoqing Ren, and Jian Sun.
\newblock Deep residual learning for image recognition.
\newblock In {\em Proceedings of the IEEE conference on computer vision and
  pattern recognition}, pages 770--778, 2016.

\bibitem{hong2021disentangling}
Youngkyu Hong, Seungju Han, Kwanghee Choi, Seokjun Seo, Beomsu Kim, and Buru
  Chang.
\newblock Disentangling label distribution for long-tailed visual recognition.
\newblock In {\em Proceedings of the IEEE/CVF Conference on Computer Vision and
  Pattern Recognition}, pages 6626--6636, 2021.

\bibitem{huang2021semi}
Siyu Huang, Tianyang Wang, Haoyi Xiong, Jun Huan, and Dejing Dou.
\newblock Semi-supervised active learning with temporal output discrepancy.
\newblock In {\em Proceedings of the IEEE/CVF International Conference on
  Computer Vision}, pages 3447--3456, 2021.

\bibitem{jacot2018neural}
Arthur Jacot, Franck Gabriel, and Cl{\'e}ment Hongler.
\newblock Neural tangent kernel: Convergence and generalization in neural
  networks.
\newblock {\em Advances in neural information processing systems}, 31, 2018.

\bibitem{jiang2018trust}
Heinrich Jiang, Been Kim, Melody Guan, and Maya Gupta.
\newblock To trust or not to trust a classifier.
\newblock {\em Advances in neural information processing systems}, 31, 2018.

\bibitem{jiang2018mentornet}
Lu Jiang, Zhengyuan Zhou, Thomas Leung, Li-Jia Li, and Li Fei-Fei.
\newblock Mentornet: Learning data-driven curriculum for very deep neural
  networks on corrupted labels.
\newblock In {\em International Conference on Machine Learning}, pages
  2304--2313. PMLR, 2018.

\bibitem{kawaguchi2016deep}
Kenji Kawaguchi.
\newblock Deep learning without poor local minima.
\newblock {\em Advances in neural information processing systems}, 29, 2016.

\bibitem{kim2021task}
Kwanyoung Kim, Dongwon Park, Kwang~In Kim, and Se~Young Chun.
\newblock Task-aware variational adversarial active learning.
\newblock In {\em Proceedings of the IEEE/CVF Conference on Computer Vision and
  Pattern Recognition}, pages 8166--8175, 2021.

\bibitem{kingma2015adam}
Diederik~P Kingma and Jimmy Ba.
\newblock Adam: A method for stochastic optimization.
\newblock In {\em International Conference on Learning Representations (ICLR)},
  2015.

\bibitem{krizhevsky2009learning}
Alex Krizhevsky and Geoffrey Hinton.
\newblock Learning multiple layers of features from tiny images.
\newblock Technical report, University of Toronto, 2009.

\bibitem{krizhevsky2014cifar}
Alex Krizhevsky, Vinod Nair, and Geoffrey Hinton.
\newblock The cifar-10 dataset.
\newblock {\em online: http://www.cs.toronto.edu/kriz/cifar.html}, 55(5), 2014.

\bibitem{samuli2017temporal}
Samuli Laine and Timo Aila.
\newblock Temporal ensembling for semi-supervised learning.
\newblock In {\em International Conference on Learning Representations (ICLR)}.
  OpenReview.net, 2017.

\bibitem{lee2018deep}
Jaehoon Lee, Yasaman Bahri, Roman Novak, Samuel~S Schoenholz, Jeffrey
  Pennington, and Jascha Sohl-Dickstein.
\newblock Deep neural networks as gaussian processes.
\newblock In {\em International Conference on Learning Representations}, 2018.

\bibitem{lewis1994sequential}
David~D Lewis and William~A Gale.
\newblock A sequential algorithm for training text classifiers.
\newblock In {\em Proceedings of the 17th annual international ACM SIGIR
  conference on Research and development in information retrieval}, pages
  3--12, 1994.

\bibitem{li2018visualizing}
Hao Li, Zheng Xu, Gavin Taylor, Christoph Studer, and Tom Goldstein.
\newblock Visualizing the loss landscape of neural nets.
\newblock {\em Advances in neural information processing systems}, 31, 2018.

\bibitem{liu2019large}
Ziwei Liu, Zhongqi Miao, Xiaohang Zhan, Jiayun Wang, Boqing Gong, and Stella~X
  Yu.
\newblock Large-scale long-tailed recognition in an open world.
\newblock In {\em Proceedings of the IEEE/CVF Conference on Computer Vision and
  Pattern Recognition}, pages 2537--2546, 2019.

\bibitem{netzer2011reading}
Yuval Netzer, Tao Wang, Adam Coates, Alessandro Bissacco, Bo Wu, and Andrew~Y.
  Ng.
\newblock Reading digits in natural images with unsupervised feature learning.
\newblock In {\em NIPS Workshop on Deep Learning and Unsupervised Feature
  Learning 2011}, 2011.

\bibitem{northcutt2021confident}
Curtis Northcutt, Lu Jiang, and Isaac Chuang.
\newblock Confident learning: Estimating uncertainty in dataset labels.
\newblock {\em Journal of Artificial Intelligence Research}, 70:1373--1411,
  2021.

\bibitem{papernot2018deep}
Nicolas Papernot and Patrick McDaniel.
\newblock Deep k-nearest neighbors: Towards confident, interpretable and robust
  deep learning.
\newblock {\em arXiv preprint arXiv:1803.04765}, 2018.

\bibitem{park2022data}
Seo~Yeon Park and Cornelia Caragea.
\newblock A data cartography based mixup for pre-trained language models.
\newblock {\em arXiv preprint arXiv:2205.03403}, 2022.

\bibitem{parvaneh2022active}
Amin Parvaneh, Ehsan Abbasnejad, Damien Teney, Gholamreza~Reza Haffari, Anton
  van~den Hengel, and Javen~Qinfeng Shi.
\newblock Active learning by feature mixing.
\newblock In {\em Proceedings of the IEEE/CVF Conference on Computer Vision and
  Pattern Recognition}, pages 12237--12246, 2022.

\bibitem{pleiss2020identifying}
Geoff Pleiss, Tianyi Zhang, Ethan Elenberg, and Kilian~Q Weinberger.
\newblock Identifying mislabeled data using the area under the margin ranking.
\newblock {\em Advances in Neural Information Processing Systems},
  33:17044--17056, 2020.

\bibitem{ren2021survey}
Pengzhen Ren, Yun Xiao, Xiaojun Chang, Po-Yao Huang, Zhihui Li, Brij~B Gupta,
  Xiaojiang Chen, and Xin Wang.
\newblock A survey of deep active learning.
\newblock {\em ACM Computing Surveys (CSUR)}, 54(9):1--40, 2021.

\bibitem{roth2006margin}
Dan Roth and Kevin Small.
\newblock Margin-based active learning for structured output spaces.
\newblock In {\em European Conference on Machine Learning}, pages 413--424.
  Springer, 2006.

\bibitem{sener2018active}
Ozan Sener and Silvio Savarese.
\newblock Active learning for convolutional neural networks: A core-set
  approach.
\newblock In {\em International Conference on Learning Representations}, 2018.

\bibitem{shannon1948mathematical}
Claude~Elwood Shannon.
\newblock A mathematical theory of communication.
\newblock {\em The Bell system technical journal}, 27(3):379--423, 1948.

\bibitem{shui2020deep}
Changjian Shui, Fan Zhou, Christian Gagn{\'e}, and Boyu Wang.
\newblock Deep active learning: Unified and principled method for query and
  training.
\newblock In {\em International Conference on Artificial Intelligence and
  Statistics}, pages 1308--1318. PMLR, 2020.

\bibitem{sinha2019variational}
Samarth Sinha, Sayna Ebrahimi, and Trevor Darrell.
\newblock Variational adversarial active learning.
\newblock In {\em Proceedings of the IEEE/CVF International Conference on
  Computer Vision}, pages 5972--5981, 2019.

\bibitem{song2019selfie}
Hwanjun Song, Minseok Kim, and Jae-Gil Lee.
\newblock Selfie: Refurbishing unclean samples for robust deep learning.
\newblock In {\em International Conference on Machine Learning}, pages
  5907--5915. PMLR, 2019.

\bibitem{swayamdipta2020dataset}
Swabha Swayamdipta, Roy Schwartz, Nicholas Lourie, Yizhong Wang, Hannaneh
  Hajishirzi, Noah~A Smith, and Yejin Choi.
\newblock Dataset cartography: Mapping and diagnosing datasets with training
  dynamics.
\newblock In {\em Proceedings of the 2020 Conference on Empirical Methods in
  Natural Language Processing (EMNLP)}, pages 9275--9293, 2020.

\bibitem{toneva2018empirical}
Mariya Toneva, Alessandro Sordoni, Remi~Tachet des Combes, Adam Trischler,
  Yoshua Bengio, and Geoffrey~J Gordon.
\newblock An empirical study of example forgetting during deep neural network
  learning.
\newblock In {\em International Conference on Learning Representations}, 2018.

\bibitem{tran2019bayesian}
Toan Tran, Thanh-Toan Do, Ian Reid, and Gustavo Carneiro.
\newblock Bayesian generative active deep learning.
\newblock In {\em International Conference on Machine Learning}, pages
  6295--6304. PMLR, 2019.

\bibitem{van2018inaturalist}
Grant Van~Horn, Oisin Mac~Aodha, Yang Song, Yin Cui, Chen Sun, Alex Shepard,
  Hartwig Adam, Pietro Perona, and Serge Belongie.
\newblock The inaturalist species classification and detection dataset.
\newblock In {\em Proceedings of the IEEE conference on computer vision and
  pattern recognition}, pages 8769--8778, 2018.

\bibitem{xiao2017fashion}
Han Xiao, Kashif Rasul, and Roland Vollgraf.
\newblock Fashion-mnist: a novel image dataset for benchmarking machine
  learning algorithms.
\newblock {\em arXiv preprint arXiv:1708.07747}, 2017.

\bibitem{yoo2019learning}
Donggeun Yoo and In~So Kweon.
\newblock Learning loss for active learning.
\newblock In {\em Proceedings of the IEEE/CVF Conference on Computer Vision and
  Pattern Recognition}, pages 93--102, 2019.

\bibitem{zhang2021understanding}
Chiyuan Zhang, Samy Bengio, Moritz Hardt, Benjamin Recht, and Oriol Vinyals.
\newblock Understanding deep learning (still) requires rethinking
  generalization.
\newblock {\em Communications of the ACM}, 64(3):107--115, 2021.

\bibitem{zhang2021imitating}
Jiayao Zhang, Hua Wang, and Weijie Su.
\newblock Imitating deep learning dynamics via locally elastic stochastic
  differential equations.
\newblock {\em Advances in Neural Information Processing Systems}, 34, 2021.

\bibitem{zhang2021cartography}
Mike Zhang and Barbara Plank.
\newblock Cartography active learning.
\newblock In {\em Findings of the Association for Computational Linguistics:
  EMNLP 2021}, pages 395--406, 2021.

\bibitem{zhou2020bbn}
Boyan Zhou, Quan Cui, Xiu-Shen Wei, and Zhao-Min Chen.
\newblock Bbn: Bilateral-branch network with cumulative learning for
  long-tailed visual recognition.
\newblock In {\em Proceedings of the IEEE/CVF Conference on Computer Vision and
  Pattern Recognition}, pages 9719--9728, 2020.

\bibitem{zhou2020curriculum}
Tianyi Zhou, Shengjie Wang, and Jeffrey Bilmes.
\newblock Curriculum learning by dynamic instance hardness.
\newblock {\em Advances in Neural Information Processing Systems},
  33:8602--8613, 2020.

\bibitem{zhu2017generative}
Jia-Jie Zhu and Jos{\'e} Bento.
\newblock Generative adversarial active learning.
\newblock {\em arXiv preprint arXiv:1702.07956}, 2017.

\end{thebibliography}
